\documentclass[journal]{IEEEtran}
\usepackage{amsmath,amsfonts}
\usepackage{algorithmic}
\usepackage{algorithm}
\usepackage{array}
\usepackage[caption=false,font=normalsize,labelfont=sf,textfont=sf]{subfig}
\usepackage{textcomp}
\usepackage{stfloats}
\usepackage{url}
\usepackage{verbatim}
\usepackage{graphicx}
\usepackage{cite}
\usepackage{multirow}
\usepackage{subcaption}
\usepackage{calc}
\usepackage[table]{xcolor}
\usepackage{makecell}
\usepackage{booktabs}
\usepackage{bbding}
\usepackage[colorlinks,linkcolor=red,citecolor=blue,urlcolor=blue]{hyperref}

\newcommand{\best}[1]{\cellcolor[HTML]{FE996B}#1}
\newcommand{\second}[1]{\cellcolor[HTML]{FFCE93}#1}
\newcommand{\third}[1]{\cellcolor[HTML]{FFF2CC}#1}

\begin{document}

\title{Realizing Immersive Volumetric Video: A Multimodal Framework for 6-DoF VR Engagement}

\author{Zhengxian Yang, Shengqi Wang, Shi Pan, Hongshuai Li, Haoxiang Wang, Lin Li, Guanjun Li~\IEEEmembership{Member,~IEEE}, \\Zhengqi Wen,~\IEEEmembership{Member,~IEEE} Borong Lin, Jianhua Tao,~\IEEEmembership{Senior Member,~IEEE},~and~Tao Yu,~\IEEEmembership{Member,~IEEE}%
\thanks{
This work was supported by the National Natural Science Foundation of China under Grant 62171255, the Tsinghua University-Fuzhou Joint Institute for Data Technology under Grant JIDT2024003, the Tsinghua University-Migu Xinkong Culture Technology (Xiamen) Co. Ltd. Joint Research Center for Intelligent Light Field and Interaction Technology, and the Beijing Municipal Science and Technology Commission ``Generic Technology Platform for Light Fields'' Project under Grant Z23111000290000.
\textit{(Zhengxian Yang, Shengqi Wang and Shi Pan contributed equally to this work.)}
\textit{(Corresponding author: Tao Yu.)}}%
\thanks{Zhengxian Yang, Shengqi Wang, Shi Pan and Haoxiang Wang are with Tsinghua University, Beijing, China. (e-mails: {zx-yang23, shengqi-21, ps23, whx22}@mails.tsinghua.edu.cn).}%
\thanks{Hongshuai Li and Guanjun Li are with the Institute of Automation, Chinese Academy of Sciences, Beijing, China. (e-mails: lihongshuai24@mails.ucas.ac.cn, guanjun.li@nlpr.ia.ac.cn).}%
\thanks{Lin Li is with the Migu Beijing Research Institute, Beijing, China (e-mail: lilin@migu.chinamobile.com).}%
\thanks{Borong Lin is with the School of Architecture, Tsinghua University, Beijing, China. (e-mail: linbr@tsinghua.edu.cn).}%
\thanks{Jianhua Tao is with the Department of Automation, Tsinghua University, Beijing, China. (e-mail: jhtao@tsinghua.edu.cn).}%
\thanks{Zhengqi Wen and Tao Yu are with the BNRist, Tsinghua University, Beijing, China. (e-mails: zqwen@tsinghua.edu.cn, ytrock@mail.tsinghua.edu.cn).}%
}


\maketitle

\begin{abstract}
Fully immersive experiences that tightly integrate 6-DoF visual and auditory interaction are essential for virtual and augmented reality (VR/AR). 
While such experiences can be achieved through computer-generated content, constructing them directly from real-world captured videos remains largely unexplored. 
We introduce Immersive Volumetric Videos (IVV), a new volumetric media format designed to provide large 6-DoF interaction spaces, audiovisual feedback, and high-resolution, high-frame-rate dynamic content.
To support IVV construction, we present \textit{ImViD}, a multi-view, multi-modal dataset built upon a space-oriented capture philosophy. Our custom capture rig enables synchronized multi-view video-audio acquisition during motion, facilitating efficient capture of complex indoor and outdoor scenes with rich foreground--background interactions and challenging dynamics. The dataset provides 5K-resolution videos at 60 FPS with durations of 1--5 minutes, offering richer spatial, temporal, and multimodal coverage than existing benchmarks.

Leveraging this dataset, we develop a dynamic light field reconstruction framework built upon a Gaussian-based spatio-temporal representation, incorporating flow-guided sparse initialization, joint camera temporal calibration, and multi-term spatio-temporal supervision for robust and accurate modeling of complex motion. We further propose, to our knowledge, the first method for sound field reconstruction from such multi-view audiovisual data. Together, these components form a unified pipeline for immersive volumetric video production. 
Extensive benchmarks and immersive VR experiments demonstrate that our pipeline generates high-quality, temporally stable audiovisual volumetric content with large 6-DoF interaction spaces.
This work provides both a foundational definition and a practical construction methodology for immersive volumetric videos, facilitating future research in immersive media production. 
Project page: \url{https://github.com/Metaverse-AI-Lab-THU/ImViD}
\end{abstract}


\begin{figure}
    \centering
    \includegraphics[width=\linewidth]{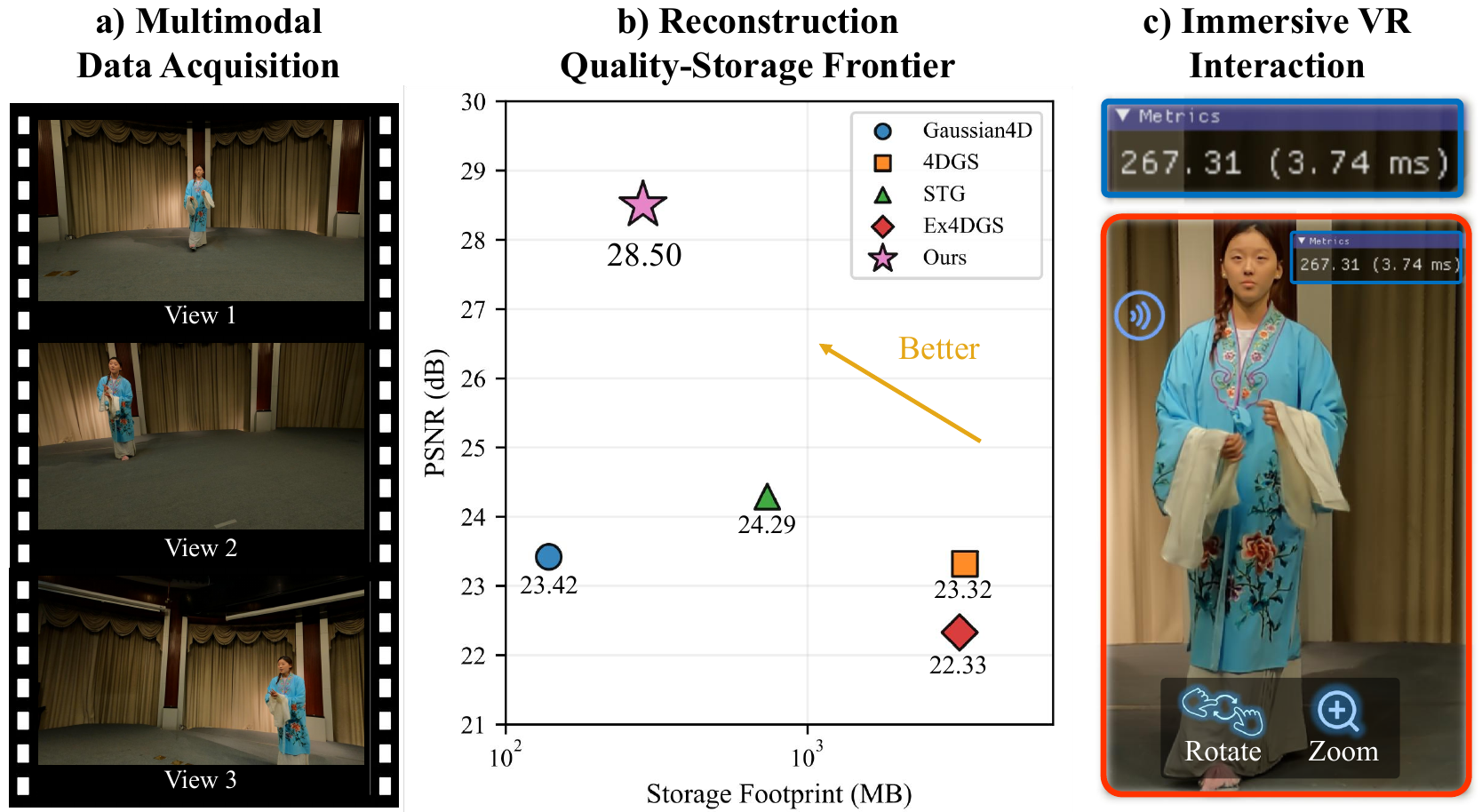}
    \caption{Equipped with a) Multimodal Capture, b) a State-of-the-art Audiovisual Reconstruction Framework, and c) Seamless 6-DoF VR Interaction, we achieve high-fidelity Immersive Volumetric Videos (IVV) construction.}
    \label{fig:teaser}
\end{figure}

\section{Introduction}
\label{sec:intro}
\IEEEPARstart{H}{igh-quality}, high-fidelity digital modeling of the real world has long been a central goal in computer vision and graphics. With recent advances in multi-view reconstruction and virtual reality technologies, generating visually compelling and interactive 3D content has become increasingly feasible, enabling applications such as holographic telepresence, immersive training, virtual tourism and cultural heritage preservation. 
More recently, generative models like Marble~\cite{marble} and Seedance 2.0~\cite{seedance} have demonstrated impressive capabilities in synthesizing novel views with consistent styles and plausible geometry from limited image inputs. However, despite their creative potential, these generative approaches still struggle to support free-viewpoint exploration of expansive environments or to faithfully record and restore complex dynamic content for authentic immersive viewing.
Achieving truly immersive user experiences that mirror the real world, therefore, remains an open challenge.

\textit{What constitutes the next generation of immersive media?}
Drawing on prior immersive systems~\cite{broxton2020immersive,apple2024} and user studies~\cite{narciso2019immersive,gracia}, we identify four technical pillars essential for deep immersion:
(i) Omnidirectional coverage of both foreground and background, 
(ii) High-quality 6-DoF interaction, 
(iii) Multi-modal experience with visual-auditory feedback, and 
(iv) High-frame-rate, long-duration dynamic content. 
We refer to a media format that simultaneously fulfills all these criteria as an \textbf{\emph{immersive volumetric video}}.

Despite steady progress in capture systems and reconstruction algorithms, existing solutions fall short of meeting all four requirements. Human- or object-centric datasets such as Panoptic Sports~\cite{Joo_2017_TPAMI}, ZJU-Mocap~\cite{peng2021neural}, and DiVa-360~\cite{lu2024diva} achieve dense 360$^\circ$ capture but lack complex backgrounds and multi-modal signals. Google Immersive Light Field Video~\cite{broxton2020immersive} enables high-quality 6-DoF rendering, yet remains limited in temporal duration, frame rate, and auditory immersion. Replay~\cite{shapovalov2023replay} introduces spatial audio but adopts a camera configuration that is less compatible with natural human viewing, restricting interaction quality. These limitations highlight the absence of an integrated capture and reconstruction paradigm tailored for immersive volumetric videos.

In parallel, advances in neural rendering and explicit scene representations have significantly improved the fidelity and efficiency of 3D reconstruction. NeRF~\cite{mildenhall2021nerf} and its subsequent extensions have demonstrated impressive visual quality, while 3D Gaussian Splatting (3DGS)~\cite{kerbl3DGaussianSplatting2023b} and its variants~\cite{lu2023scaffold,yan2024street,yang2024spec,malarzgaussian,zhang2024gaussian,cheng2024gaussianpro,li2024geogaussian} have enabled real-time rendering via explicit representations. Nevertheless, reconstructing dynamic immersive scenes remains particularly challenging. Existing dynamic light field methods, whether explicit Gaussian-based approaches~\cite{wu20244d,li2024spacetime,katsumata2024compactdynamic3dgaussian,kratimenos2023dynmf,lin2024gaussian} or implicit neural representations~\cite{li2022neural,song2023nerfplayer,wang2023mixed,attal2023hyperreel,fridovich2023k,cao2023hexplane,shao2023tensor4d}, struggle to balance foreground--background accuracy, temporal coherence, rendering efficiency, and storage costs in real-world, texture-rich environments with complex motion. Moreover, even millisecond-level temporal misalignment across multi-view cameras can cause noticeable artifacts in fast motion, an issue often overlooked in existing pipelines.

To better support immersive volumetric video research and address the challenges discussed above, we develop a mobile multi-view capture rig equipped with synchronized cameras and microphones, together with space-oriented acquisition strategies that combine dense static sampling with long-duration dynamic capture. Based on this system, we construct \textbf{\textit{ImViD}}, an audiovisual dataset comprising large-scale indoor and outdoor scenes with 360$^\circ$ coverage, high resolution (5K), high frame rate (60 FPS), and rich foreground--background interactions.
We further propose a robust dynamic light field reconstruction framework based on a Gaussian-based spatio-temporal representation to address complex motion and real-world capture imperfections. Our method integrates flow-guided sparse initialization to decouple static and dynamic regions, introduces camera-wise temporal calibration to explicitly compensate for residual sub-frame misalignment during joint optimization, and employs spatio-temporal supervision to enforce photometric, geometric, and motion consistency across long-duration sequences.
Together, these designs enable faithful reconstruction of temporally coherent 4D light fields under challenging in-the-wild conditions.
Finally, we incorporate a sound field reconstruction module to support consistent 6-DoF audiovisual immersion. A complete pipeline for immersive volumetric video construction is established.

Our main contributions can be summarized as follows:
\begin{itemize}[]
    \item We define \emph{Immersive Volumetric Videos (IVV)} as a new volumetric media for VR/AR applications and introduce \emph{ImViD}, a high-quality multi-view, multimodal dataset for immersive volumetric video. It features 360$^\circ$ foreground--background capture, long-duration video \& audio data under high resolution @5K and high frame rate @60 FPS. 
    
    \item We propose a robust dynamic light field reconstruction framework using a Gaussian-based spatio-temporal representation. By integrating flow-guided initialization, joint temporal calibration, and spatio-temporal supervision, our approach achieves high-fidelity, coherent reconstruction for in-the-wild motion, surpassing existing methods.

    \item We establish a comprehensive pipeline for IVV production (Fig.~\ref{fig:pipeline}), which, to our knowledge, is the first to jointly reconstruct dynamic light and sound fields from multi-view audiovisual captures. The effectiveness of our framework is validated through extensive experiments and immersive VR demonstrations, providing a practical methodology for high-quality immersive media creation.
\end{itemize}
 
\begin{figure*}
    \centering
    \includegraphics[width=\textwidth]{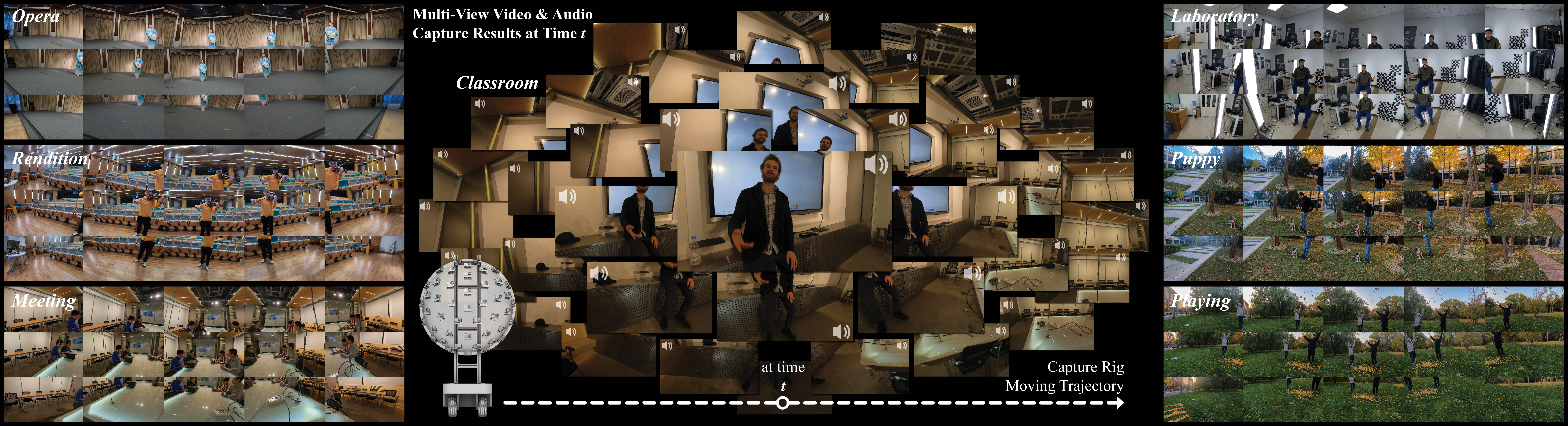}
    \caption{We introduce \textit{\textbf{{ImViD}}}, a dataset for immersive volumetric videos. 
\textit{ImViD} records dynamic scenes using a multi-view audio-video capture rig moving in a space-oriented manner, which provides a new benchmark for immersive volumetric video reconstruction and its application.}
    \label{fig:dataset}
\end{figure*}

\section{Related Work}
\subsection{Datasets for Dynamic Novel View Synthesis}
Early studies on dynamic scene reconstruction mainly focused on human avatars. Representative datasets like Human3.6M~\cite{ionescu2013human3}, Panoptic Sports~\cite{Joo_2017_TPAMI}, ZJU-Mocap~\cite{peng2021neural}, and Tensor-4D~\cite{shao2023tensor4d} primarily capture simple human motions with limited or no background content, limiting their applicability to immersive experiences. In the following, we review more complex datasets that incorporate environmental context, including both monocular and multi-view acquisitions.

Monocular acquisition systems are attractive due to their low cost and ease of deployment. Datasets such as HyperNeRF~\cite{park2021hypernerf}, Dynamic Scene Dataset~\cite{yoon2020novel}, and D2NeRF~\cite{wu2022ddnerf} use a mobile phone as the capture device for recording dynamic scenes through handheld motion. However, these datasets suffer from resolutions below 1080p, limited capture space (similar to fixed-point shooting), and durations under one minute. Although NeRF On-the-go~\cite{ren2024nerf} allows for larger capture ranges by walking while shooting, high-quality reconstructions are confined to the vicinity of the capture path, and the small field of view (FOV) limits prolonged observations of specific scene positions.

Multi-camera capture has gained significant attention for its ability to provide wider FOVs and richer details. For example, the Immersive Light Field dataset~\cite{broxton2020immersive} employs 46 cameras to capture 15 indoor and outdoor scenes, while Technicolor~\cite{sabater2017dataset} uses a 4$\times$4 camera rig for 12 indoor sequences. The UCSD Dynamic Scene Dataset~\cite{lin2021deep} consists of 96 outdoor videos focused on single-person activities captured by 10 cameras. The Plenoptic Dataset~\cite{li2022neural} uses 21 cameras for 6 indoor scenes. Similarly, datasets like~\cite{li2022streaming,lin2022efficient,wang2024masked} utilize 13, 18, and 24 cameras, respectively. Despite the increased camera counts, these systems rely on static capture setups, limiting them to frontal views and hindering 360$^\circ$ reconstruction. Moreover, the video sequences are typically short, with a maximum duration of 2 minutes (often less than 30 seconds) and a maximum resolution of 3840$\times$2160, which is insufficient for high-quality immersive VR experiences. 

In addition, the previously mentioned datasets, whether monocular or multi-view, lack sound recordings, despite the importance of multimodality for immersion. The Replay dataset~\cite{shapovalov2023replay} addresses this by focusing on long sequences with professional actors in familiar settings. It employs a ring of 8 static DSLR cameras paired with binaural microphones and 3 head-mounted GoPro cameras, providing 46 videos at 4K. However, aside from the head-mounted cameras, which can rotate slightly with head movements, all other cameras remain static. Furthermore, the DSLR arrangement does not align with human viewing habits in VR, making them unsuitable as benchmarks for novel view synthesis tasks.
The latest work~\cite{chen2024360+} presents a dataset of 28 scenes captured with a $360^\circ$ camera, each including multiple audio and video sequences. However, this dataset is constrained by a fixed-point shooting strategy, resulting in sparse viewpoints that hinder the reconstruction of high-quality dynamic scenes. Further comparisons between our work and these datasets can be found in Table~\ref{tab:datasets comparison}.

\subsection{Dynamic Light\&Sound Field Reconstruction} 
Temporal variations in input images make reconstructing dynamic scenes more challenging than static ones. The key issue is efficiently representing 4D scenes while ensuring spatiotemporal consistency and high accuracy. Traditional approaches address scene dynamics by estimating time-varying geometries, such as surfaces~\cite{collet2015high} and depth maps~\cite{kanade1997virtualized}. Recently, research efforts have shifted towards adapting neural radiance field representations to dynamic scenes. DyNeRF~\cite{li2022neural} pioneered dynamic scene reconstruction using neural radiance fields. Following DyNeRF, a series of works~\cite{li2022streaming,song2023nerfplayer,wang2023mixed,attal2023hyperreel,fridovich2023k,cao2023hexplane,shao2023tensor4d} have focused on improving the fidelity and efficiency of 4D neural representations. However, these methods commonly suffer from high memory consumption and the difficulty of modeling complex spatiotemporal dynamics. More recent work has extended 3DGS to dynamic scenes, enabling high-quality, real-time rendering. While modeling dynamic scenes as static ones is straightforward, methods relying solely on monocular data~\cite{jung2023deformable,liang2023gaufre} are not suitable for this paradigm and face challenges in adding constraints or priors to better supervise training. Other approaches leverage multi-view data as input. Works such as~\cite{luiten2023dynamic} and~\cite{sun20243dgstream} adopt frame-to-frame initialization to construct per-frame Gaussian point clouds, resulting in high memory consumption and temporal artifacts, including abrupt appearance and disappearance. 4DGS~\cite{wu20244d} introduces Hexplane to encode spatiotemporal information, using MLPs to predict changes in Gaussian properties. Subsequent works~\cite{duisterhof2023md,guo2024motion,li2024st} further add geometric and optical flow constraints to improve frame-to-frame motion but perform well only on short sequences with small actions. Other approaches~\cite{li2024spacetime,katsumata2024compactdynamic3dgaussian,kratimenos2023dynmf,lin2024gaussian,lee2024fully,wang2025freetimegs} employ the explicit parametric form (such as radial basis functions) of 3DGS to enhance spatiotemporal consistency. Some methods~\cite{yang2023real,duan20244d} focus on representation with dual quaternions, 4D spherical harmonics, or rotors for end-to-end training. More improvements, such as parallel training~\cite{shaw2024swings} and using codebooks and masks to enhance scene compactness~\cite{lee2024compact}, indicate the potential for efficient and high-quality representation of long-term complex dynamic scenes.


In addition to light field reconstruction, reconstructing the sound field from multimodal data is equally important, often referred to as novel-view acoustic synthesis. For the novel-view acoustic synthesis task, AV-NeRF~\cite{liang2023av} synthesizes spatial audio by leveraging visual features from rendered images. NeRAF~\cite{brunetto2024neraf} further decouples the camera and microphone poses used for training. AV-GS~\cite{bhosale2024av} incorporates a 3D scene representation during sound field synthesis, addressing the issue where previous methods overlooked the contribution of the broader 3D scene geometry. SOAF~\cite{gao2024soaf} considers occlusion in the light field when synthesizing the sound field. However, the above novel-view acoustic synthesis tasks typically require a large amount of multimodal training data collected from different locations, which is often user-unfriendly in practical applications. Moreover, these methods generally cannot solve the problem of sound field synthesis for moving sound sources. This paper proposes a training-free novel-view acoustic synthesis approach for our dataset, which also supports sound field synthesis for moving sound sources.

\begin{table*}[!t]
    \centering
    \caption{Existing real-world datasets for dynamic novel view synthesis. }
    \label{tab:datasets comparison}
    \renewcommand*{\arraystretch}{2.2}
    \setlength{\tabcolsep}{3pt}
    \resizebox{\textwidth}{!}{%
    \begin{tabular}{>{\raggedright\arraybackslash}m{2.4cm}ccccccccc>{\raggedright\arraybackslash}m{5.0cm}}
    \toprule
    \textbf{Datasets} & 
    \makecell{\textbf{No.}\\\textbf{Scene}} & 
    \makecell{\textbf{Outdoor/}\\\textbf{Indoor}} & 
    \textbf{Cameras} & 
    \textbf{Mobility} & 
    \textbf{Resolution} & 
    \textbf{Angles} & 
    \textbf{Duration} & 
    \textbf{FPS} & 
    \makecell{\textbf{Multi-}\\\textbf{modality}} & 
    \textbf{Content} \\
    \midrule
    PanopticSports~\cite{Joo_2017_TPAMI}                      & 65 & Indoor  & \textbf{480 cameras}                                            & Static                                    & 640$\times$480            & $360^\circ$                      & 5mins               & 25           & \XSolidBrush                & Human-centric actions \\
    Technicolor~\cite{sabater2017dataset}                     & 12 & Indoor  & 16 cameras                                             & Static                                    & 2048$\times$1088          & Frontal                   & 2s                  & 30           & \XSolidBrush                & Has a number of close-ups sequences, captured medium angle scenes and other animated scenes where the movement does not come from a human \\
    Immersive-Lightfield~\cite{broxton2020immersive}          & 15 & both    & 46 cameras                                             & Static                                    & 2560$\times$1920          & Frontal                   & 10--30s              & 30           & \XSolidBrush                & Simple and slow motion of human, animals, objects \\
    HyperNeRF~\cite{park2021hypernerf}                        & 17 & Indoor  & 1 hand-held phone                                      & \makecell[c]{Fixed-point\\Waving}                        & 1920$\times$1080          & Frontal                   & 30--60s              & 30           & \XSolidBrush                & Waving a mobile phone in front of a moving scene, object-centric \\
    Dynamic Scene Datasets (NVIDIA)~\cite{yoon2020novel}      & 8  & Outdoor & \makecell[c]{1 Mobile phone\\/12 cameras}              & \makecell[c]{Fixed-point\\Waving/Static}  & 1920$\times$1080          & Frontal                   & 5s                  & 60           & \XSolidBrush                & Simple body motions (facial, jump, etc.) \\
    UCSD Dynamic~\cite{lin2021deep}                           & 96 & Outdoor & 10 cameras                                             & Static                                    & 1920$\times$1080          & Frontal                   & 1--2mins             & \textbf{120} & \XSolidBrush                & Various visual effects and human interactions \\
    ZJU-Mocap~\cite{peng2021neural}                           & 10 & Indoor  & 21 cameras                                             & Static                                    & 1024$\times$1024          & $360^\circ$                      & 20s                 & 50           & \XSolidBrush                & Simple body motions (punch, kick, etc.) \\
    Plenoptic Dataset (DyNeRF/Neural 3D)~\cite{li2022neural}  & 6  & Indoor  & 21 cameras                                             & Static                                    & 2704$\times$2028          & Frontal                   & 10--30s              & 30           & \XSolidBrush                & Contains high specularity, translucency and transparency objects, motions with changing topology, selfcast moving shadows, volumetric effects, various lighting conditions and multiple people moving around in open living room space \\
    D2NeRF~\cite{wu2022ddnerf}                                & 10 & Indoor  & dual-hold phone                                        & \makecell[c]{Fixed-point\\Waving}                        & 1920$\times$1080          & Frontal                   & 5s                  & 30           & \XSolidBrush                & Contains more challenging scenarios with rapid motion and non-trivial dynamic shadows \\
    iPhone Datasets~\cite{gao2022monocular}                   & 14 & both    & \makecell[c]{1 hand-held\\phone/2 cameras }            & \makecell[c]{Fixed-point\\Waving/Static}  & 640$\times$480            & Frontal                   & 8--15s               & 30/60        & \XSolidBrush                & Featuring non-repetitive motion, from various categories such as generic objects, humans, and pets \\
    Meetroom Datasets~\cite{li2022streaming}                  & 4  & Indoor  & 13 cameras                                             & Static                                    & 1280$\times$720           & Frontal                   & 10s                 & 30           & \XSolidBrush                & One or three persons have discussion, working, trimming in a meeting room \\
    ENeRF-Outdoor~\cite{lin2022efficient}                     & 4  & Outdoor & 18 cameras                                             & Static                                    & 1920$\times$1080          & Frontal                   & 20--40s              & 30           & \XSolidBrush                & Complex human motions \\
    Replay~\cite{shapovalov2023replay}                        & 46 & Indoor  & 12 cameras                                             & Static                                    & 3840$\times$2160          & $360^\circ$                      & 5mins               & 30           & \checkmark (Audio)          & Dancing, chatting, playing video games, unwrapping presents, playing ping pong \\
    Campus Datasets~\cite{wang2024masked}                     & 6  & Outdoor & 24 cameras                                             & Static                                    & 3840$\times$2160          & Frontal                   & 5--10s               & 30           & \XSolidBrush                & Includes more realistic observations such as pedestrians, moving cars, and grasses with people playing \\
    MoDGS~\cite{liu2024modgs}                                 & 6  & both    & 1 cameras                                              & Static                                    & \textbf{--}        & Frontal                   & --                  & --           & \XSolidBrush                & Contains diverse subjects like skating, a dog eating food, YOGA, etc. \\
    DiVa-360~\cite{lu2024diva}                                & 53 & Indoor  & 53 cameras                                             & Static                                    & 1280$\times$720           & Frontal                   & 51s                 & 120          & \checkmark (Audio)          & For Object-centric tasks. Contains dynamic objects and intricate hand-object interactions \\
    NeRF On-the-go~\cite{ren2024nerf}                         & 12 & both    & 1 hand-held phone                                      & Moveable                                  & 4032$\times$3024          & $360^\circ$                      & 5--10s               & 30           & \XSolidBrush                & Including 10 outdoor and 2 indoor scenes, features a wide range of dynamic objects including pedestrians, cyclists, strollers, toys, cars, robots, and trams, along with diverse occlusion ratios ranging from 5\% to 30\% \\
    360+X~\cite{chen2024360+}                                 & 28 & both    & \makecell[c]{1 $360^\circ$ cams\\1 Spectacles cam}            & Static                                    & 5760$\times$2880          & $360^\circ$                      & 10s                 & 30           & \checkmark (Audio)          & Capture in 17 cities across 5 countries. Panoptic perspective to scene understanding with audio \\
    \textbf{ImViD(Ours)}                                      & 7  & \textbf{both} & 39 cameras                              & \textbf{Moveable}                         & \textbf{5312$\times$2988} & \textbf{Frontal/$360^\circ$}     & \textbf{1--5mins}    & 60           & \textbf{\checkmark (Audio)} & Seven common indoor and outdoor scenes in daily life, including opera, face-to-face communication, teaching, discussion, music performance, interaction with pets, and playing. Each scene has high-quality synchronized multi-view video and audio with a duration of more than 1 minute, and contains rich elements such as various small objects, glass, and changes in light and shadow \\
    \bottomrule
    \end{tabular}%
    }
\end{table*}

\begin{figure*}[!t]
    \centering
    \includegraphics[width=\textwidth]{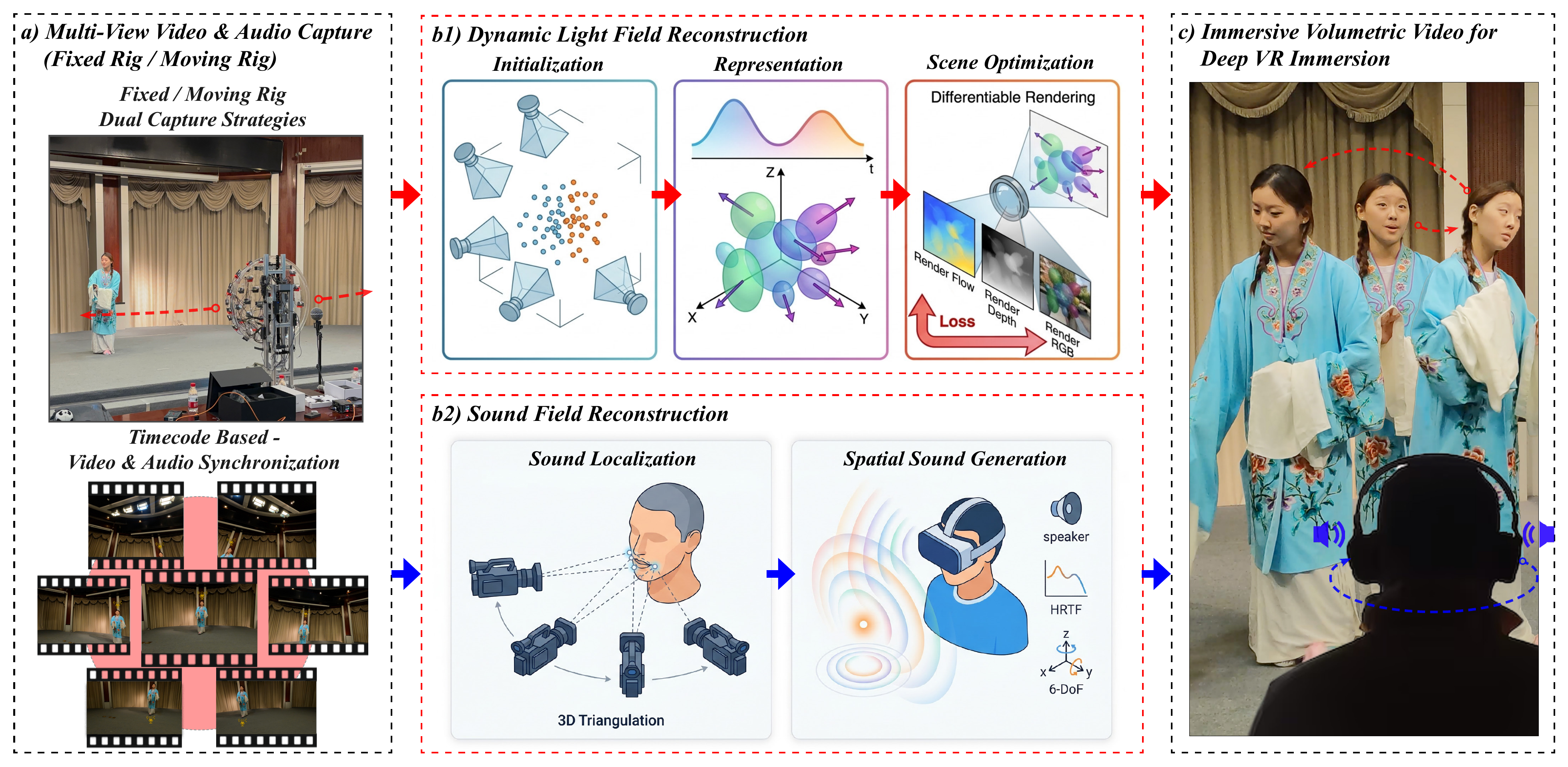}
    \caption{The pipeline to realize the multimodal 6-DoF immersive VR experiences. We applied a carefully designed rig to \textbf{(a)} simultaneously capture multi-view video and audio. The \textbf{(b1)} presents our proposed dynamic light field reconstruction framework while \textbf{(b2)} demonstrates the construction process of sound field, details can be seen in Sec.~\ref{sec:method}. Ultimately, we achieve a 6-DoF audiovisual experience \textbf{(c)}, demonstrate the effectiveness of both our dataset and the proposed pipeline.}
    \label{fig:pipeline}
\end{figure*}

\section{\textit{ImViD} Dataset}
Our goal is to construct a real-world immersive volumetric video dataset that jointly captures foreground and background content, thereby supporting research on spatial video reconstruction and VR/AR applications. To this end, we carefully design the entire data acquisition pipeline as described in~\cite{Yang_2025_CVPR}, including hardware configuration, scene selection, capture strategies, and data processing, ensuring that the elements of interest to researchers are preserved to the greatest extent. 
With informed consent obtained from all participants involved in dynamic scenes prior to data collection, the released \textit{ImViD} dataset comprises a total of 7 large scenes, including 5 indoor and 2 outdoor environments. The captured content covers a diverse set of everyday activities, such as opera performances, meetings, teaching scenarios, and outdoor leisure activities. Each scene provides a 360$^\circ$ coverage of the environment through static images, along with multi-view, time-synchronized video and audio sequences captured at 5K resolution and 60 FPS. Two complementary capture strategies are employed to capture dynamic scenes with both fine-grained motion details and extended spatial coverage, along with calibrated camera poses.

\begin{figure}[!t]
  \centering
  \includegraphics[width=\linewidth]{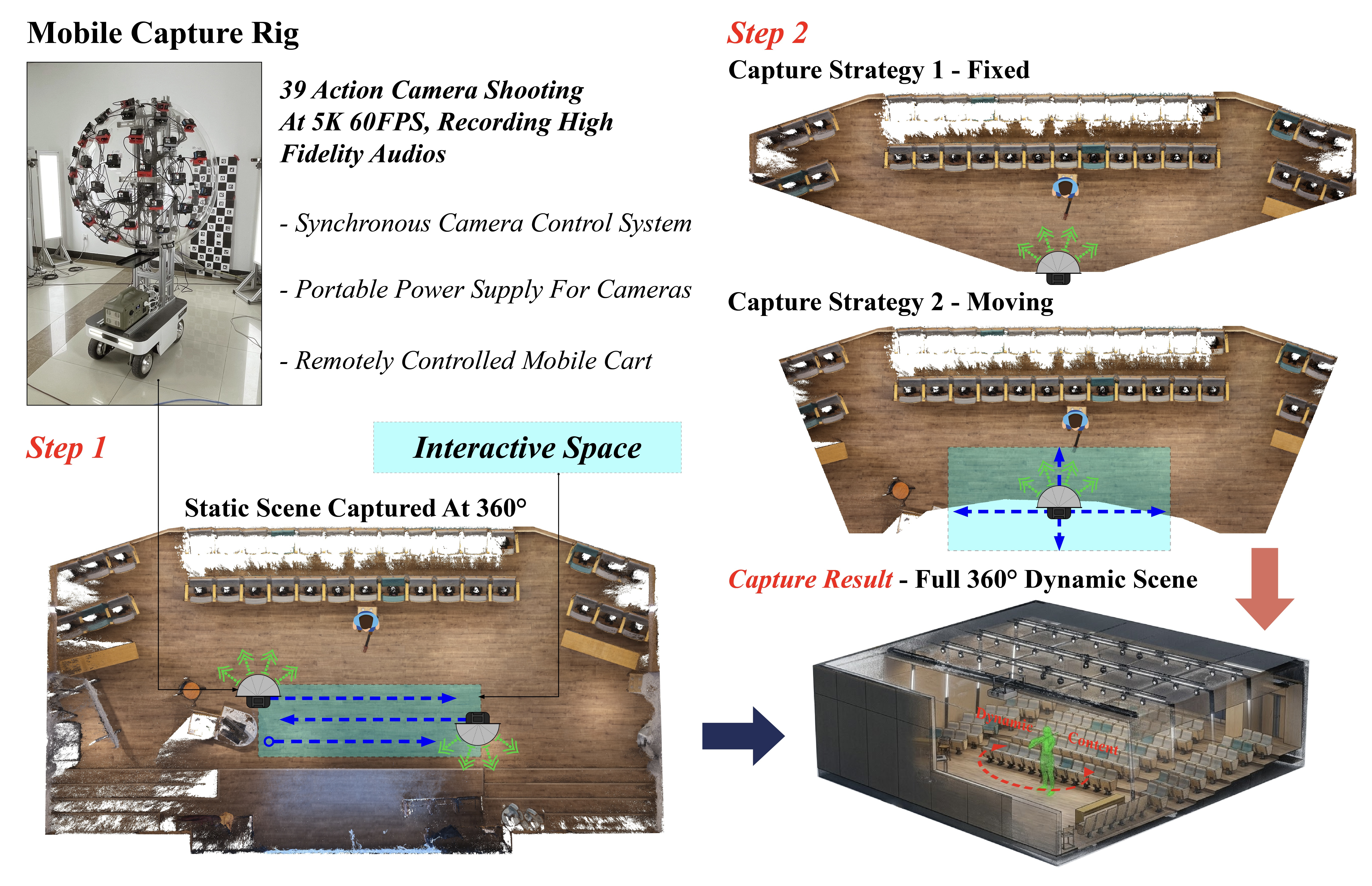}
  \caption{Our rig support two kinds of capturing strategies for high resolution, high frame rate and $360^\circ$ dynamic data.}
  \label{fig:data capture}
\end{figure}

\subsection{Data Acquisition}
Handheld monocular cameras offer flexible motion but only sparse viewpoints from various locations. In contrast, fixed camera arrays, while stationary, offer dense perspectives within a limited range. We aim to combine the advantages of both to design an effective data collection system and strategy for a fully immersive VR experience. As shown in Fig.~\ref{fig:data capture}, our system consists of 39 GoPro cameras installed on a hemispherical surface mounted on a remotely controllable mobile platform, all of which are connected to a PC for synchronized triggering and control. The rig is carefully designed to be positioned at approximately human eye height, and the camera are oriented outward to emulate a natural inside-looking-out perspective consistent with human viewing. This system can synchronously collect about 1,000 images at 5K resolution within a space of at least 6 m³ in 2 min. It further enables synchronized video and audio recording at 5K resolution and 60 FPS for up to 30 min (limited by heat dissipation). Based on this hardware setup, we adopt the following capture strategy:

\noindent\textbf{Step 1:} Efficient, high-quality, high-density $360^\circ$ image acquisition of static environments.

\noindent\textbf{Step 2:} Continuous synchronous collection of dynamic scenes under two different strategies.

i) Fixed-point shooting. The cart remains stationary during the capture, focusing on capturing dynamic details.

ii) Moving shooting. The cart slowly moves while the scene is in motion, providing additional levels of detail and a larger exploratory space.

Throughout the capture process, world time codes are strictly synchronized across all devices. Camera parameters, including focal length, exposure, and white balance, are manually calibrated per-scene through iterative adjustment to minimize hardware-induced variations in data quality. To ensure high-fidelity audio, camera noise reduction coefficients are optimized, and external environmental noise is strictly controlled during acquisition.

\begin{table*}[!t]
    \centering
    \caption{Detailed Statistics of the \textit{ImViD} Dataset.}
    \label{tab:dataset_statistic}
    \resizebox{\linewidth}{!}{%
    \begin{tabular}{lcccccccc}
    \toprule
    Name & \#Camera & \#Static Views & \#Take & \#Strategy & Density ($m^3/s$) & Viewing Space & Avg. Duration & Storage (GB) \\ 
    \midrule
    Scene 1 Opera & 39 & 1152 & 2 & 1 & -- & $180^{\circ}$ & 3min22s & 226 \\
    Scene 2 Laboratory & 39 & 1225 & 2 & 2 & 0.10 & $360^{\circ}$ & 1min42s & 137.3 \\
    Scene 3 Classroom & 39 & 1223 & 2 & 2 & 0.10 & $360^{\circ}$ & 4min42s & 497 \\
    Scene 4 Meeting & 39 & 1223 & 1 & 1 & -- & $360^{\circ}$ & 3min16s & 114 \\
    Scene 5 Rendition & 39 & 1620 & 4 & 2 & 0.10 & $360^{\circ}$ & 2min02s & 516 \\
    Scene 6 Puppy & 39 & 1404 & 3 & 2 & 0.10 & $360^{\circ}$ & 1min50s & 359 \\
    Scene 7 Playing & 39 & 1224 & 2 & 2 & 0.10 & $360^{\circ}$ & 1min10s & 220 \\
    \midrule
    \textbf{Total} & -- & -- & \textbf{16} & -- & -- & -- & \textbf{38min46s} & \textbf{2069.3} \\ 
    \bottomrule
    \end{tabular}%
    }\par \vspace{2pt}
    \parbox{0.95\linewidth}{\footnotesize \textit{Note:} \textbf{\#Take} represents the number of episodes. \textbf{\#Strategy} indicates the capture strategy in Step 2 (1: fixed-point only; 2: both fixed and moving). \textbf{Density} denotes the spatiotemporal captured volume per second.}
\end{table*}

\subsection{Data Processing}
\label{sec:data_processing}
The data processing pipeline is designed to ensure high-precision geometric and temporal alignment across multimodal sequences.
For Step 1 images, we perform Structure-from-Motion (SfM) using COLMAP~\cite{schoenberger2016sfm}. Camera intrinsic priors and the known relative arrangement of the rig are incorporated to accelerate calibration and improve stability. This procedure establishes a globally consistent coordinate system and produces a sparse geometric reference for each scene.

For the video and audio from Step~2, synchronization across all cameras is first verified via embedded world time codes. Accurate temporal alignment is critical, as synchronization quality directly affects the fidelity of dynamic reconstruction and the spatial consistency of the sound field. The resulting synchronization error is approximately 5~ms.
For fixed-point sequences, cameras at Frame~0 are registered to the calibrated static scene through feature matching and triangulation, yielding intrinsic and extrinsic parameters within the same global coordinate system. This guarantees seamless spatial integration of dynamic content with the 360$^\circ$ background. Subsequent frames inherit these parameters, with sparse point clouds generated via frame-wise triangulation.
For moving sequences, the static scene is treated as a geometric backbone. At each time step, camera poses are estimated through feature matching with the static environment, using the previous frame as initialization for optimization to improve efficiency and temporal smoothness in long-duration captures. Since the rig is not perfectly rigid and may undergo slight deformation during motion, per-frame camera parameters are optimized rather than enforcing a single rigid transformation across the sequence.

Given that the audiovisual data is captured by integrated GoPro units, the spatial offset between the lens and the internal microphone is negligible relative to the scene scale. Accordingly, the estimated camera poses are used as the microphone positions for sound field reconstruction.

\subsection{Dataset Analysis}
\noindent\textbf{Diversity.} 
Our dataset, as depicted in Table~\ref{tab:dataset_statistic}, includes 7 common indoor and outdoor scenes, each comprising one or more takes, totaling 16 video sequences. Each scene is rich in foreground and background elements, meticulously crafted to preserve the authenticity of the environment. The characters, clothing, and motions within each scene exhibit a diverse range, set against varying ambient lighting conditions. We also included areas of interest, such as tables, chairs, and windows, as much as possible for researchers.

\noindent\textbf{Quality.} 
Leveraging a total of 39 cameras, we captured dynamic scenes at a resolution of 5312$\times$2988 and 60 FPS. The images of the static environment are 5568$\times$4176.

\noindent\textbf{Duration.}
As outlined in Table~\ref{tab:dataset_statistic}, each scene in our dataset contains at least 1 minute of footage. The longest sequence in our forthcoming public dataset spans approximately 5 minutes. The total processed data duration amounts to 38 minutes and 46 seconds, including 139,560 frames per camera.

\noindent\textbf{Coverage.}
\begin{figure}
    \centering
    \includegraphics[width=\linewidth]{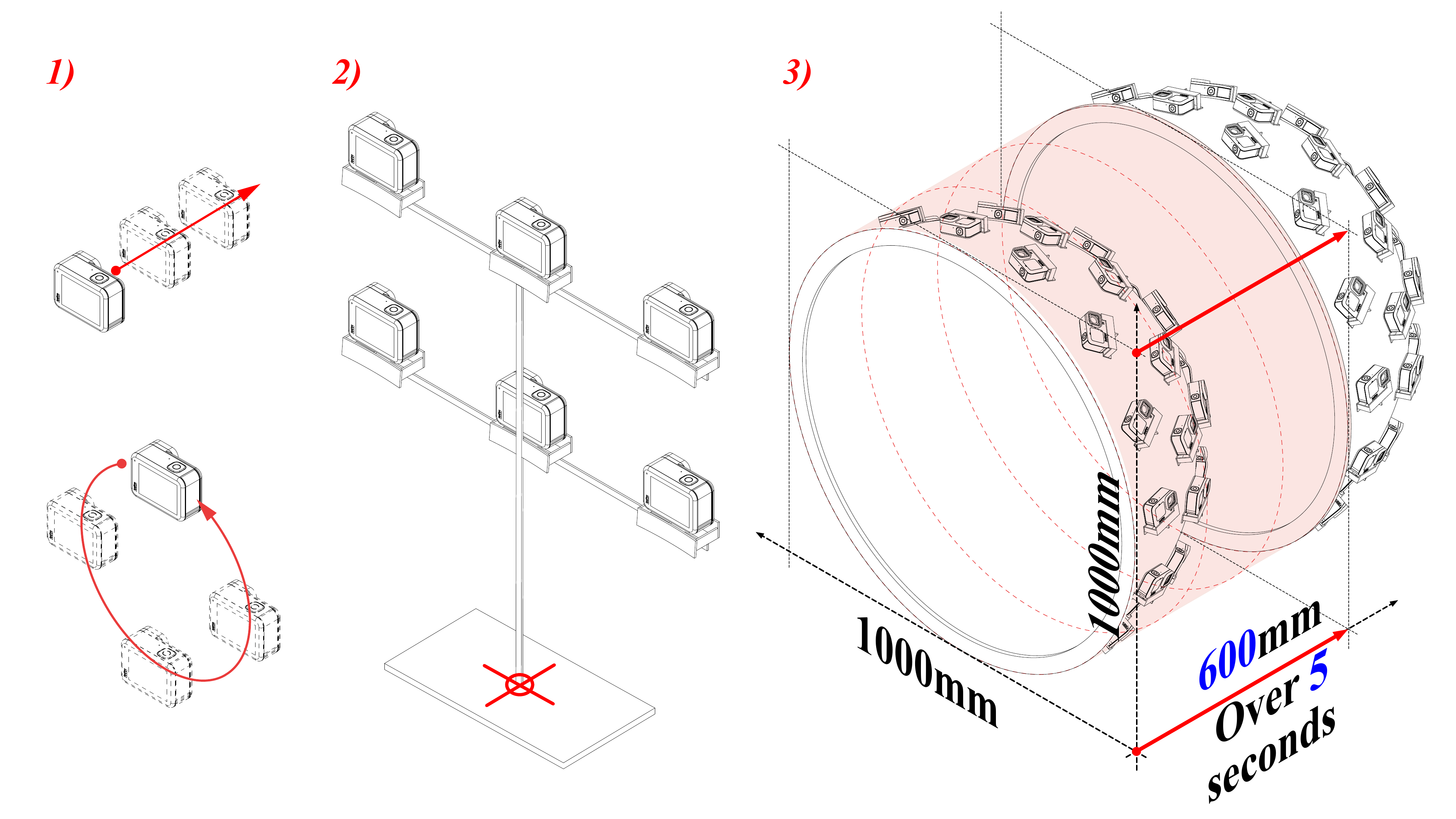}
    \caption{Calculation method for \textbf{spatiotemporal capture density}. 1) the capture strategy of handheld monocular camera 2) represents the fixed camera array 3) Our rig moving covers a volume of $0.6\pi\times0.5^2\ \mathrm{m}^3$ over 5 seconds.}
    \label{fig:spatialtemporaldensity}
\end{figure}
As highlighted in Table~\ref{tab:dataset_statistic}, we conducted image acquisition from over 1,000 viewpoints before recording dynamic content, thereby offering a $180^\circ$--$360^\circ$ viewing space in VR. To characterize the mobile rig, we propose a metric termed `spatiotemporal capture density', quantifying the captured volume covered by our rig per second, as shown in Fig.~\ref{fig:spatialtemporaldensity}. This metric, unavailable in current datasets (where static camera arrays are deemed 0 and handheld monocular cameras follow a single trajectory without capturing volume), registers at approximately 0.10 $m^3/s$ in our dataset.

\section{Method}
\label{sec:method}
\begin{figure*}[!t]
    \centering
    \includegraphics[width=\textwidth]{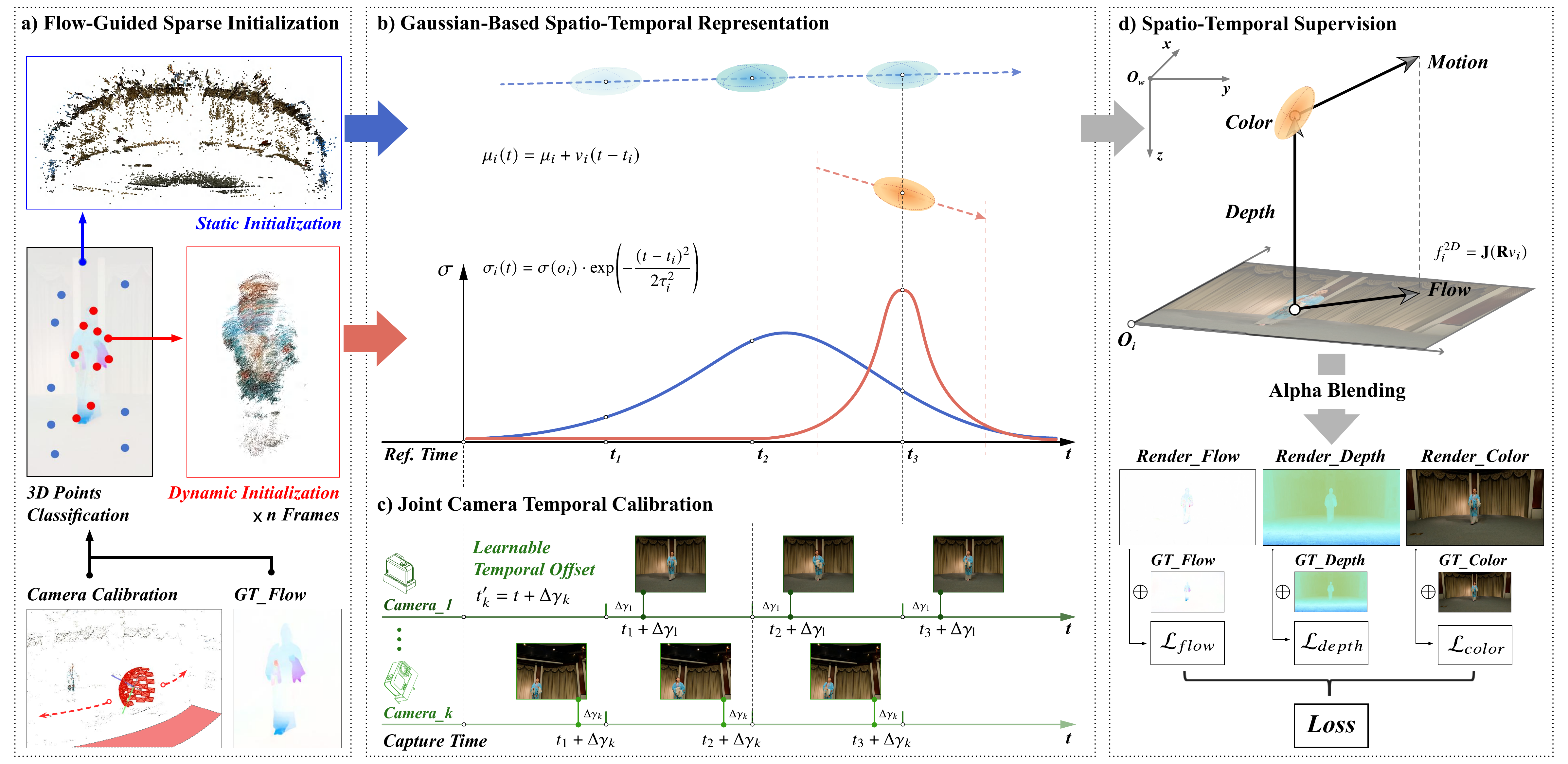}
    \caption{\textbf{Overview of the proposed Dynamic Light Field Reconstruction method.} 
    The pipeline starts with \textbf{Flow-Guided Sparse Initialization (a)}, which leverages SfM geometry and optical flow priors to decouple static and dynamic regions, initializing static primitives globally and dynamic ones per frame to reduce redundancy. 
    The scene is then represented using a \textbf{Gaussian-Based Spatio-Temporal Representation (b)}, encoding spatial geometry, appearance, and temporal dynamics, where motion is modeled by linear velocity and visibility by Gaussian-modulated temporal opacity.
    In the final stage, we jointly optimize the scene parameters and perform \textbf{Joint Camera Temporal Calibration (c)} to refine per-camera temporal offsets for sub-frame alignment.
    Rendered color, depth, and optical flow maps are further supervised through \textbf{Spatio-Temporal Supervision (d)}, enforcing photometric, geometric, and motion consistency.
    }
    \label{fig:pipeline_method}
\end{figure*}

Evolving from the preliminary framework in~\cite{Yang_2025_CVPR}, our proposed framework comprises two main components:
(i) a novel Gaussian-based dynamic light field reconstruction method, as illustrated in Fig.~\ref{fig:pipeline_method}, which incorporates flow-guided sparse initialization strategy to decouple static and dynamic regions, joint camera temporal calibration mechanism to address residual sub-frame misalignment, and multi-term spatio-temporal supervision to simultaneously regularize photometric, geometric, and motion consistency for high-fidelity rendering;
(ii) a sound field reconstruction method that synthesizes spatial audio to complement the visual realism, ensuring coherent and immersive audiovisual engagement.

\subsection{Dynamic Light Field Reconstruction}

\subsubsection{Gaussian-based Spatio-temporal Representation}
Building upon 3D Gaussian Splatting (3DGS)~\cite{kerbl3DGaussianSplatting2023b}, we represent a dynamic scene as a set of spatio-temporal Gaussian primitives $\mathcal{G}=\{G_i\}_{i=1}^N$. 
Each primitive encodes spatial geometry, appearance, and temporal dynamics, and is parameterized as
\begin{equation}
G_i = \{ \mu_i, q_i, s_i, f_i, o_i, v_i, \tau_i, t_i \},
\end{equation}
where $\mu_i \in \mathbb{R}^3$, $q_i \in \mathbb{R}^4$, and $s_i \in \mathbb{R}^3$ denote the canonical center position, rotation quaternion, and scaling factors respectively.
$f_i$ represents the Spherical Harmonics (SH) coefficients for view-dependent color, and $o_i$ is the base opacity logit.
To model temporal behavior, we introduce a velocity vector $v_i \in \mathbb{R}^3$, a temporal extent $\tau_i$ controlling visibility duration, and a temporal center $t_i$, which denotes the peak visibility timestamp, serving as the temporal reference.

At time $t$, the spatial center of each primitive evolves linearly as:
\begin{equation}
\mu_i(t) = \mu_i + v_i (t - t_i).
\end{equation}
Temporal visibility is modeled by  Gaussian decay on the opacity:
\begin{equation}
\sigma_i(t) = \sigma(o_i)\exp\!\left(-\frac{(t - t_i)^2}{\tau_i^2}\right),
\end{equation}
where $\sigma(\cdot)$ denotes the sigmoid function.
With fixed rotation $q_i$ and scale $s_i$, each spatio-temporal Gaussian primitive at time $t$ effectively reduces to a standard 3D Gaussian $\{\mu_i(t), q_i, s_i, f_i, \sigma_i(t)\}$, enabling real-time rendering via the original tile-based rasterization pipeline of 3DGS.

\subsubsection{Flow-Guided Sparse Initialization}
In complex dynamic scenes, objects may abruptly appear or disappear, making initialization from a single-frame point cloud (e.g., Gaussian4D~\cite{wu20244d}, 4DGS~\cite{yang2023real}) insufficient and often leading to blur and floating artifacts.
STG~\cite{li2024spacetime} alleviates this issue by aggregating sparse point clouds across frames, but simple concatenation introduces substantial redundancy and hampers optimization efficiency.
Therefore, we propose an effective initialization strategy to reduce primitive redundancy and accelerate convergence.

Leveraging SfM-calibrated geometry and dense optical flow estimated by VideoFlow~\cite{Shi_2023_ICCV}, we classify each triangulated 3D point by querying the optical flow magnitude at its projected 2D location across multiple views.
A point is labeled as \emph{dynamic} if the sampled flow exceeds a threshold $\epsilon_f$ in any view; otherwise, it is considered \emph{static}.
Based on this classification, we adopt different initialization strategies:

\noindent\textbf{Dynamic Initialization:}
Dynamic points are instantiated on a per-frame basis, with frame-aligned temporal centers and compact temporal extents to capture localized motion.

\noindent\textbf{Static Initialization:}
Static points are initialized once from a reference frame (typically frame~0), using fixed temporal centers and larger temporal extents to ensure temporal stability.

We unify static and dynamic primitives across all frames to construct the initial representation. This strategy yields a sparse yet expressive set of spatio-temporal Gaussian primitives, enabling accurate modeling of dynamic regions while maintaining representation compactness for static areas.

\subsubsection{Joint Camera Temporal Calibration}
Accurate temporal alignment across cameras is critical for sharp 4D reconstruction, especially in practical \textit{in-the-wild} capture setups.
As illustrated in Fig.~\ref{fig:pipeline_method}(c), sub-frame timing discrepancies caused by electronic trigger latency and sensor readout delays can introduce noticeable inter-view misalignment.
Although hardware-based world time codes typically achieve millisecond-level synchronization, this accuracy is often insufficient for high-fidelity dynamic reconstruction, leading to blurring or ghosting artifacts in fast-moving regions if left unmodeled.

To address this issue, we introduce a learnable per-camera temporal offset $\Delta \gamma_k \in \mathbb{R}$ to model residual deviations from the nominal capture time.
Given the synchronized timestamp $t$, the effective capture time of camera $k$ is defined as
\begin{equation}
t'_k = t + \Delta \gamma_k.
\end{equation}

The offset $\Delta \gamma_k$ is shared across all frames of the same camera, capturing camera-specific yet temporally consistent latency.
By jointly optimizing these offsets together with the spatio-temporal Gaussian primitives, rendered observations are temporally realigned to better match the underlying scene dynamics.
An $L_2$ regularization term $\mathcal{L}_{reg}$ is applied to constrain excessive temporal drift, ensuring stable optimization while remaining close to the hardware-synchronized timestamps.

\subsubsection{Spatio-temporal Supervision}
We further design a multi-term supervision scheme to better guide the joint optimization of spatio-temporal Gaussian primitives and camera temporal offsets. 
By integrating photometric, geometric, and motion constraints, the proposed supervision encourages temporally coherent geometry and visually faithful appearance.

\noindent\textbf{Photometric Reliability.} To compensate for view-dependent photometric inconsistencies across cameras, we apply a camera-specific learnable bilateral grid~\cite{wang2024bilateral} to the rendered image, producing a color-aligned output $\hat{\mathcal{I}}_{rend}$.
The reconstruction is supervised using a photometric loss that combines an $\ell_1$ term with DSSIM:
\begin{equation}
\begin{aligned}
\mathcal{L}_{color} = \;& (1 - \lambda_{dssim}) \lVert \hat{\mathcal{I}}_{rend} - \mathcal{I}_{gt} \rVert_1 \\
& + \lambda_{dssim} \mathcal{L}_{\text{DSSIM}}(\hat{\mathcal{I}}_{rend}, \mathcal{I}_{gt}).
\end{aligned}
\end{equation}

\noindent\textbf{Geometric Consistency.}
To reduce geometric ambiguity in textureless regions, we incorporate monocular depth priors estimated by Depth-Anything-V2~\cite{depth_anything_v2}. Following~\cite{kerbl3DGaussianSplatting2023b,kerbl2024hierarchical}, we first align the raw predicted depth $\mathcal{D}_{pred}$ to the sparse COLMAP SfM points using a per-frame linear transformation, yielding the scale-consistent depth $\hat{\mathcal{D}}_{pred}$. We then supervise the rendered depth $\mathcal{D}_{rend}$ using an $\ell_1$ loss:
\begin{equation}
\mathcal{L}_{depth} = \lVert \mathcal{D}_{rend} - \hat{\mathcal{D}}_{pred} \rVert_1.
\end{equation}

\noindent\textbf{Motion Continuity.}
Although the velocity parameter $v_i$ allows Gaussians to model dynamic behavior, recovering accurate 3D motion solely from photometric supervision remains challenging, particularly in areas involving complex motion or textureless surfaces.
In such areas, weak photometric gradients offer insufficient constraints, allowing Gaussians to exhibit unstable or drifting motion. For example, in static yet textureless regions, primitives initialized with zero velocity may continue to move arbitrarily, as the rendered appearance remains visually consistent with the ground truth despite the underlying physical inaccuracy.
To enforce physically plausible dynamics, we align projected 3D velocities with observed 2D optical flow.
Specifically, a world-space velocity $v_i$ is transformed into the camera coordinate system using the rotation matrix $\mathbf{R}$ and projected onto the image plane via the Jacobian $\mathbf{J}$:
\begin{equation}
f_i^{2D} = \mathbf{J} (\mathbf{R} v_i), \quad \text{where } \mathbf{J} =
\begin{bmatrix}
\frac{f_x}{Z} & 0 & -\frac{f_x X}{Z^2} \\[4pt]
0 & \frac{f_y}{Z} & -\frac{f_y Y}{Z^2}
\end{bmatrix}.
\end{equation}
Here, $(X, Y, Z)$ denotes the Gaussian center in camera space, and $(f_x, f_y)$ denotes the focal lengths.
Similar to color rendering, we aggregate these projected $2\text{D}$ velocities $f_i^{2D}$ using alpha blending to obtain the pixel-wise rasterized optical flow $\mathcal{F}_{rend} = \sum_{i \in \mathcal{N}} f_i^{2D} \alpha_i \prod_{j=1}^{i-1} (1 - \alpha_j)$.
We then supervise the velocity parameters by minimizing the endpoint error (EPE) between $\mathcal{F}_{rend}$ and the pseudo-ground-truth flow $\mathcal{F}_{pred}$ estimated using VideoFlow~\cite{Shi_2023_ICCV}:
\begin{equation}
\mathcal{L}_{flow} = \lVert \mathcal{F}_{rend} - \mathcal{F}_{pred} \rVert_2.
\end{equation}

\noindent\textbf{Total Loss.}
The overall objective is defined as a weighted sum of the above terms and regularization:
\begin{equation}
\begin{aligned}
\mathcal{L}_{total} =\;&
\lambda_{color}\mathcal{L}_{color}
+ \lambda_{perc}\mathcal{L}_{perc}
+ \lambda_{depth}\mathcal{L}_{depth} \\
&+ \lambda_{flow}\mathcal{L}_{flow}
+ \lambda_{reg}\mathcal{L}_{reg},
\end{aligned}
\end{equation}
where each weight $\lambda$ controls the contribution of its corresponding term, and $\mathcal{L}_{perc}$ denotes the LPIPS (AlexNet) perceptual loss~\cite{zhang2018perceptual}.

\subsection{Sound Field Reconstruction}
\begin{figure}
    \centering
    \includegraphics[width=\linewidth]{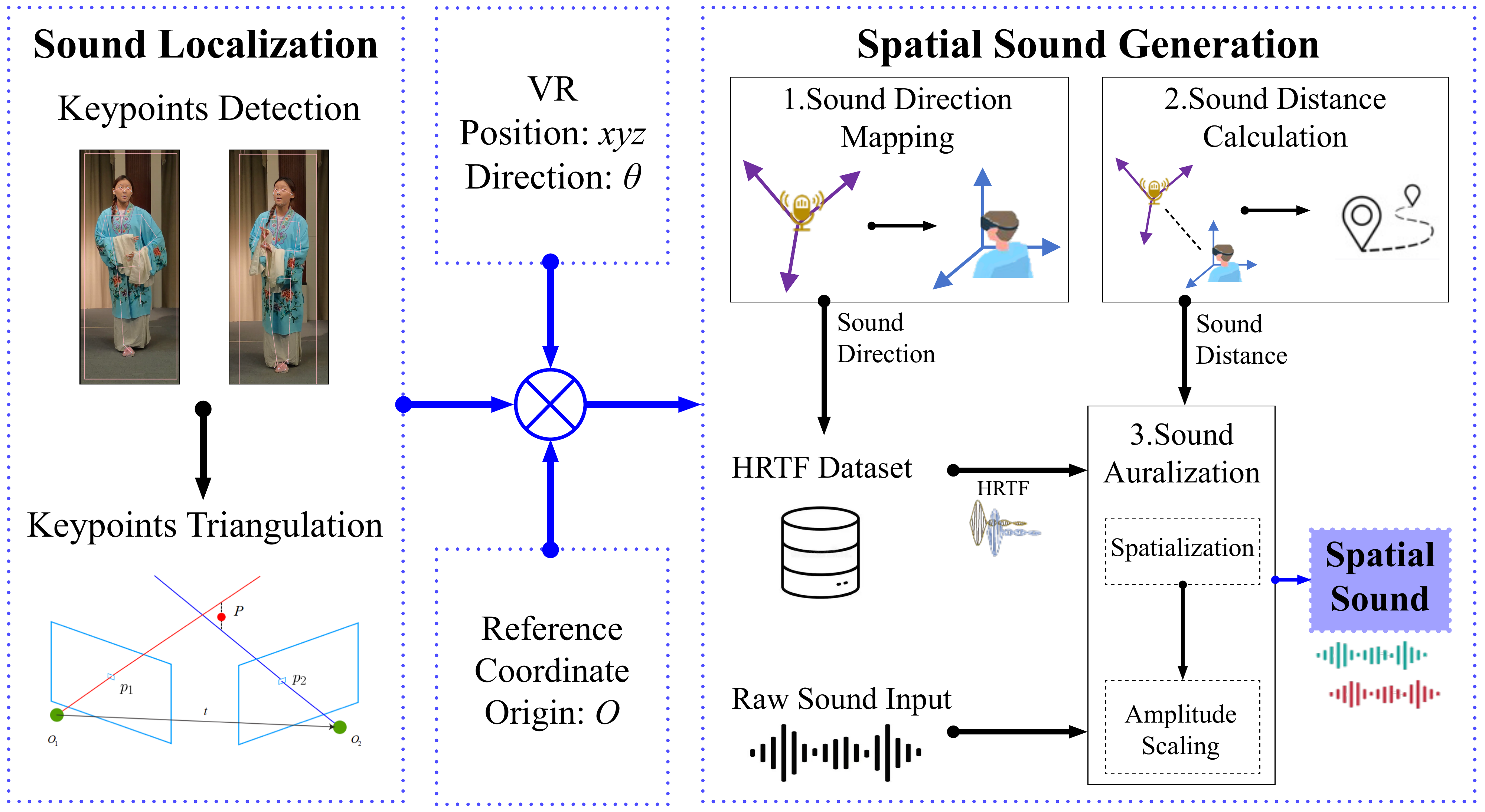}
    \caption{\textbf{Overview of the proposed sound field reconstruction framework.} The pipeline consists of two main modules. \textbf{Sound Localization} estimates the 3D coordinates of the sound source. \textbf{Spatial Sound Generation} then synthesizes binaural audio based on the VR user's real-time pose.}
    \label{fig:soundfield}
\end{figure}

To complete the immersive experience, we complement our high-fidelity visual reconstruction with a spatial audio synthesis framework, shown in Fig.~\ref{fig:soundfield}. Taking the recording microphone as the origin of the coordinate system, we denote the coordinates of the sound source as $s(t)=\{x_s(t),y_s(t)\}$, the user wearing the VR device as $l(t)=\{x_l(t), y_l(t)\}$, and the angle of deviation from the y-axis in the counterclockwise direction as $\theta_l(t)$, where $t$ is the time index. Based on these premises, generating the user's binaural spatial audio can be divided into three parts: sound direction mapping, sound distance mapping, and sound auralization. 
Note that we assume there is only one dominant sound source (omnidirectional emitting) to simplify the sound field reconstruction. 

In the sound direction mapping part, we calculate the direction of the sound source relative to the user's ears, $\theta_s(t)$,
\begin{equation}
    \theta_s(t) = \arccos\frac{\mathbf{v}_1(t)\mathbf{v}_2(t)}{|\mathbf{v}_1(t)||\mathbf{v}_2(t)|},
\end{equation}
where
\begin{align}
\mathbf{v}_1(t) &= [x_s(t)-x_l(t), y_s(t)-y_l(t)]^T,\\
\mathbf{v}_2(t) &= [-\sin{\left(\theta_l(t)\right)}, \cos{\left(\theta_l(t)\right)}]^T,
\end{align}
In the sound distance mapping part, we calculate the scaling of the sound source at the user's ears, relative to the original recording. We denote this scaling parameter as $\lambda(t)$,
\begin{align}
    \lambda(t) = \frac{\sqrt{x_s^2(t)+y_s^2(t)}}{\sqrt{(x_s(t)-x_l(t))^2+(y_s(t)-y_l(t))^2}}
\end{align}
In the sound auralization part, we convert the original recording into spatial audio based on $\theta_s(t)$ and $\lambda(t)$.
In the short-time Fourier transform (STFT) domain, we denote the audio from the left and right ears as $A_L(t,f)$ and $A_R(t,f)$, respectively, where $f$ is the frequency index. Then $A_L(t,f)$ and $A_R(t,f)$ can be written as
\begin{align}
    A_L(t,f) &= F_L(\theta_s(t)) F_l(t,f)  A_O(t,f) \label{left}\\   A_R(t,f) &= F_R(\theta_s(t)) F_l(t,f) A_O(t,f), \label{right}
\end{align}
where $F_L(\theta_s(t))$ and $F_R(\theta_s(t))$ are head-related transfer functions (HRTF) based on $\theta_s(t)$, $F_l(t)$ denotes the room impulse response (RIR) at the user's location, and $A_O(t,f)$ is the sound emitted from the speaker in the STFT domain.
We utilize $\theta_s(t)$ to retrieve $F_L(\theta_s(t))$ and $F_R(\theta_s(t))$ from the SADIE II dataset~\cite{armstrong2018perceptual}.
In order to model $F_l(t)$ and $A_O(t,f)$, we make the following assumptions: $F_l(t)$ changes between different locations at the same time index $t$ depending only on the parameter $\lambda(t)$, i.e.,
\begin{align}
    F_l(t,f) = \lambda(t) F(t,f),
\end{align}
where $F(t)$ is the RIR at the recording microphone.
Using the above assumption, Eq.(\ref{left}) and Eq.(\ref{right}) can be rewritten as 
\begin{align}
        A_L(t,f) &= \lambda(t) F_L(\theta_s(t)) A(t,f), \\   
        A_R(t,f) &= \lambda(t) F_R(\theta_s(t)) A(t,f),
\end{align}
where $A(t,f)=F(t,f) A_O(t,f)$ is the audio at the recording microphone.

\subsection{Implementation Details}
Our dynamic light field reconstruction framework is implemented in PyTorch and optimized using Adam for 30 epochs, where each epoch corresponds to one full random permutation of the training frames.
The spatio-temporal parameters are optimized with distinct learning rates: 
$2\times10^{-3}$ for velocity $v_i$, 
$1\times10^{-5}$ for temporal centers $t_i$, 
and $3\times10^{-2}$ for temporal extents $\tau_i$.
All remaining parameters follow the same learning rate schedule in vanilla 3DGS. 
The loss weights are set as follows:
$\lambda_{dssim}=0.2$, 
$\lambda_{color}=1.0$, 
$\lambda_{perc}=0.1$, 
$\lambda_{depth}=1.0$, 
$\lambda_{flow}=1.0$, 
and $\lambda_{reg}=10^{-4}$.
For flow-guided initialization, we use a fixed flow threshold $\epsilon_f=0.1$ across all scenes. 

For sound field reconstruction, we estimate the 3D sound source location by tracking the speaker's mouth across frames using RTMPose~\cite{jiang2023rtmpose}. Multi-view 2D keypoints are triangulated to obtain the 3D position, which is treated as a point sound source. The microphone of the central camera serves as the recording microphone and defines the origin of the sound field coordinate system.

\section{Experimental Settings}
\paragraph{Datasets}

We conduct experiments on five dynamic view synthesis datasets,
including our proposed \textit{ImViD} and four public benchmarks.
\textit{N3V}~\cite{li2022neural} contains 6 indoor scenes captured with 19--21 cameras at a resolution of 2704$\times$2028 and 30~FPS.
\textit{MPEG-GSC}~\cite{Li_2025_CVPR_GIFStream} includes two indoor sequences with fast motions, captured by 20--30 cameras at approximately 1080p resolution and 30~FPS.
\textit{MeetRoom}~\cite{li2022streaming} comprises 4 indoor scenes recorded using 13 cameras at 1280$\times$720 and 30~FPS.
\textit{Google Immersive}~\cite{broxton2020immersive} provides 15 indoor and outdoor scenes captured with 46 cameras at 2560$\times$1920 and 30~FPS.
For each scene, we select 300 frames for evaluation.

\paragraph{Baselines}
To evaluate both the effectiveness of our \textit{ImViD} dataset and the performance of our dynamic light field reconstruction framework, we conduct comparative experiments with four recent representative Gaussian-based dynamic reconstruction methods:
\textit{Gaussian4D}~\cite{wu20244d}, \textit{4DGS}~\cite{yang2023real},
\textit{STG}~\cite{li2024spacetime}, and \textit{Ex4DGS}~\cite{lee2024fully}.

\paragraph{Evaluation Protocol}
We follow a unified evaluation protocol to ensure fair and reproducible comparisons across all methods.

\noindent\textbf{Hardware:} 
All dynamic light field reconstruction experiments are conducted on a single NVIDIA A100 GPU.

\noindent\textbf{Resolution:} 
For \textit{MeetRoom} and \textit{MPEG-GSC}, we preserve the original input resolution, as these datasets are captured at relatively low resolutions, and retaining the native scale better reveals reconstruction details.
For \textit{N3V} and \textit{Google Immersive}, we apply 2$\times$ downsampling following common practice in prior works~\cite{wu20244d,yang2023real,li2024spacetime,lee2024fully}. All methods on \textit{ImViD} are also evaluated using 2$\times$ downsampled inputs.
    
\noindent\textbf{Train-Test split:} 
Consistent with standard dynamic view synthesis benchmarks, we hold out a single viewpoint for testing and use the remaining views for training. 
Specifically, Camera~10 is used for \textit{MPEG-GSC}, Camera~1 for \textit{Google Immersive}, and Camera~0 for \textit{ImViD}, \textit{N3V}, and \textit{MeetRoom}.

\noindent\textbf{Metrics:} 
We report PSNR, SSIM, and LPIPS (AlexNet) as the primary quantitative evaluation metrics.

\section{Results and Analysis}

\subsection{Dynamic Light Field Reconstruction}

\subsubsection{Quantitative Evaluation}

As shown in Table~\ref{tab:quantitative_comparison_stacked}, our dynamic scene reconstruction method consistently establishes state-of-the-art performance across all five datasets on the evaluated metrics (PSNR, SSIM, and LPIPS). 
Notably, on the \textit{ImViD} dataset, which represents the most challenging in-the-wild scenario, our method achieves substantial gains, with an average PSNR improvement of 4.21~dB over the strongest baseline and a significant reduction in LPIPS from 0.153 to 0.078.
While certain baseline methods outperform ours on individual metrics in specific scenes, these differences are largely attributable to scene characteristics and methodological biases. For example, on \textit{N3V}, Gaussian4D benefits from an additional static stage, leading to stronger background fitting in scenes with limited motion and marginally higher PSNR.

Our approach achieves a superior trade-off between reconstruction quality and storage overhead. As shown in Table~\ref{tab:efficiency}, our method maintains a compact storage footprint while delivering significantly higher rendering quality than existing baselines in complex \textit{ImViD} scenes. Furthermore, our implementation reaches an average rendering speed of 200 FPS on an NVIDIA RTX 4090 GPU.

\begin{table*}[!t]
    \centering
    \caption{\textbf{Quantitative comparison on Gaussian-based dynamic light field reconstruction method across multiple datasets.} All metrics are calculated on the test views across all frames (300 frames for each scene). The \colorbox[HTML]{FE996B}{best}, \colorbox[HTML]{FFCE93}{second} and \colorbox[HTML]{FFF2CC}{third} results are highlighted with color blocks.}
    \label{tab:quantitative_comparison_stacked}
    \renewcommand{\arraystretch}{0.9}

    \resizebox{\linewidth}{!}{%
    \begin{tabular}{p{2.3cm} ccc ccc ccc ccc ccc}
    \toprule
    \multirow{2}{*}{(a) \textbf{ImViD}} & \multicolumn{3}{c}{Gaussian4D} & \multicolumn{3}{c}{4DGS} & \multicolumn{3}{c}{STG} & \multicolumn{3}{c}{Ex4DGS} & \multicolumn{3}{c}{\textbf{Ours}} \\
    \cmidrule(lr){2-4} \cmidrule(lr){5-7} \cmidrule(lr){8-10} \cmidrule(lr){11-13} \cmidrule(lr){14-16}
     & PSNR$\uparrow$ & SSIM$\uparrow$ & LPIPS$\downarrow$ & PSNR$\uparrow$ & SSIM$\uparrow$ & LPIPS$\downarrow$ & PSNR$\uparrow$ & SSIM$\uparrow$ & LPIPS$\downarrow$ & PSNR$\uparrow$ & SSIM$\uparrow$ & LPIPS$\downarrow$ & PSNR$\uparrow$ & SSIM$\uparrow$ & LPIPS$\downarrow$ \\
    \midrule
    Scene 1 Opera & \third{25.61} & 0.873 & 0.206 & 25.01 & \third{0.877} & \third{0.174} & \second{26.30} & \second{0.899} & \second{0.169} & 23.08 & 0.835 & 0.219 & \best{33.51} & \best{0.916} & \best{0.070} \\
    Scene 2 Laboratory & \third{22.15} & \third{0.908} & 0.099 & 21.02 & \third{0.910} & \third{0.081} & \second{23.40} & \second{0.920} & \second{0.067} & 20.29 & 0.896 & 0.106 & \best{31.10} & \best{0.941} & \best{0.034} \\
    Scene 5 Rendition & 25.78 & 0.848 & 0.206 & \second{26.23} & \second{0.865} & \third{0.158} & \third{26.21} & \third{0.861} & \second{0.145} & 24.73 & 0.857 & 0.159 & \best{27.84} & \best{0.884} & \best{0.073} \\
    Scene 6 Puppy & 20.15 & 0.408 & 0.553 & 21.03 & 0.568 & 0.299 & \second{21.25} & \second{0.597} & \third{0.231} & \third{21.21} & \third{0.586} & \second{0.226} & \best{21.53} & \best{0.607} & \best{0.135} \\
    \midrule
    Average & \third{23.42} & 0.759 & 0.266 & 23.32 & \third{0.805} & \third{0.178} & \second{24.29} & \second{0.819} & \second{0.153} & 22.33 & 0.794 & \third{0.178} & \best{28.50} & \best{0.837} & \best{0.078} \\
    \bottomrule
    \end{tabular}%
    }

    \vspace{4pt}

    \resizebox{\linewidth}{!}{%
    \begin{tabular}{p{2.3cm} ccc ccc ccc ccc ccc}
    \toprule
    \multirow{2}{*}{(b) \textbf{N3V}} & \multicolumn{3}{c}{Gaussian4D} & \multicolumn{3}{c}{4DGS} & \multicolumn{3}{c}{STG} & \multicolumn{3}{c}{Ex4DGS} & \multicolumn{3}{c}{\textbf{Ours}} \\
    \cmidrule(lr){2-4} \cmidrule(lr){5-7} \cmidrule(lr){8-10} \cmidrule(lr){11-13} \cmidrule(lr){14-16}
     & PSNR$\uparrow$ & SSIM$\uparrow$ & LPIPS$\downarrow$ & PSNR$\uparrow$ & SSIM$\uparrow$ & LPIPS$\downarrow$ & PSNR$\uparrow$ & SSIM$\uparrow$ & LPIPS$\downarrow$ & PSNR$\uparrow$ & SSIM$\uparrow$ & LPIPS$\downarrow$ & PSNR$\uparrow$ & SSIM$\uparrow$ & LPIPS$\downarrow$ \\
    \midrule
    Coffee Martini & \best{28.62} & \third{0.909} & 0.083 & 27.32 & 0.893 & 0.097 & 27.55 & \second{0.913} & \third{0.075} & \second{28.12} & \best{0.915} & \second{0.067} & \third{27.69} & 0.908 & \best{0.044} \\
    Cook Spinach & 32.39 & 0.947 & 0.049 & \third{32.59} & 0.946 & 0.054 & \second{33.01} & \second{0.955} & \third{0.042} & 32.34 & \second{0.955} & \second{0.043} & \best{34.61} & \best{0.958} & \best{0.021} \\
    Cut Roasted Beef & 27.48 & 0.932 & 0.059 & 29.59 & 0.932 & 0.057 & \second{33.20} & \second{0.955} & \third{0.043} & \third{32.04} & \third{0.953} & \second{0.035} & \best{35.38} & \best{0.960} & \best{0.021} \\
    Flame Salmon & \best{28.86} & 0.912 & 0.081 & 27.84 & 0.896 & 0.094 & \second{28.78} & \second{0.920} & \second{0.069} & \third{28.69} & \best{0.921} & \third{0.066} & 28.27 & \third{0.916} & \best{0.039} \\
    Flame Steak & 32.28 & 0.954 & 0.040 & 31.05 & 0.950 & 0.047 & \second{33.68} & \second{0.964} & \second{0.031} & \third{33.32} & \third{0.962} & \third{0.036} & \best{36.19} & \best{0.967} & \best{0.018} \\
    Sear Steak & \third{33.27} & 0.955 & 0.040 & 32.66 & 0.952 & 0.050 & \second{33.68} & \second{0.964} & \second{0.033} & 32.77 & \third{0.961} & \second{0.033} & \best{37.16} & \best{0.968} & \best{0.016} \\
    \midrule
    Average & 30.48 & 0.935 & 0.059 & 30.18 & 0.928 & 0.067 & \second{31.65} & \second{0.945} & \third{0.049} & \third{31.21} & \second{0.945} & \second{0.047} & \best{33.22} & \best{0.946} & \best{0.027} \\
    \bottomrule
    \end{tabular}%
    }

    \vspace{4pt}

    \resizebox{\linewidth}{!}{%
    \begin{tabular}{p{2.3cm} ccc ccc ccc ccc ccc}
    \toprule
    \multirow{2}{*}{(c) \textbf{MPEG-GSC}} & \multicolumn{3}{c}{Gaussian4D} & \multicolumn{3}{c}{4DGS} & \multicolumn{3}{c}{STG} & \multicolumn{3}{c}{Ex4DGS} & \multicolumn{3}{c}{\textbf{Ours}} \\
    \cmidrule(lr){2-4} \cmidrule(lr){5-7} \cmidrule(lr){8-10} \cmidrule(lr){11-13} \cmidrule(lr){14-16}
     & PSNR$\uparrow$ & SSIM$\uparrow$ & LPIPS$\downarrow$ & PSNR$\uparrow$ & SSIM$\uparrow$ & LPIPS$\downarrow$ & PSNR$\uparrow$ & SSIM$\uparrow$ & LPIPS$\downarrow$ & PSNR$\uparrow$ & SSIM$\uparrow$ & LPIPS$\downarrow$ & PSNR$\uparrow$ & SSIM$\uparrow$ & LPIPS$\downarrow$ \\
    \midrule
    Bartender & 32.29 & 0.918 & 0.097 & \second{35.31} & \second{0.945} & \second{0.052} & \third{34.88} & \second{0.945} & \third{0.058} & 32.80 & 0.930 & 0.076 & \best{36.16} & \best{0.948} & \best{0.026} \\
    CBA & 25.52 & 0.897 & 0.171 & \third{29.85} & 0.944 & \third{0.055} & \second{31.99} & \best{0.949} & \second{0.045} & 28.07 & 0.923 & 0.069 & \best{32.17} & \best{0.949} & \best{0.028} \\
    \midrule
    Average & 28.91 & 0.908 & 0.134 & \third{32.58} & \third{0.945} & \third{0.054} & \second{33.44} & \second{0.947} & \second{0.052} & 30.44 & 0.927 & 0.073 & \best{34.17} & \best{0.949} & \best{0.027} \\
    \bottomrule
    \end{tabular}%
    }

    \vspace{4pt}

    \resizebox{\linewidth}{!}{%
    \begin{tabular}{p{2.3cm} ccc ccc ccc ccc ccc}
    \toprule
    \multirow{2}{*}{(d) \textbf{MeetRoom}} & \multicolumn{3}{c}{Gaussian4D} & \multicolumn{3}{c}{4DGS} & \multicolumn{3}{c}{STG} & \multicolumn{3}{c}{Ex4DGS} & \multicolumn{3}{c}{\textbf{Ours}} \\
    \cmidrule(lr){2-4} \cmidrule(lr){5-7} \cmidrule(lr){8-10} \cmidrule(lr){11-13} \cmidrule(lr){14-16}
     & PSNR$\uparrow$ & SSIM$\uparrow$ & LPIPS$\downarrow$ & PSNR$\uparrow$ & SSIM$\uparrow$ & LPIPS$\downarrow$ & PSNR$\uparrow$ & SSIM$\uparrow$ & LPIPS$\downarrow$ & PSNR$\uparrow$ & SSIM$\uparrow$ & LPIPS$\downarrow$ & PSNR$\uparrow$ & SSIM$\uparrow$ & LPIPS$\downarrow$ \\
    \midrule
    Discussion & 30.64 & 0.952 & 0.041 & \third{30.80} & \third{0.958} & \third{0.038} & \second{31.77} & \second{0.965} & \second{0.031} & 28.53 & 0.936 & 0.055 & \best{35.01} & \best{0.967} & \best{0.018} \\
    Trimming & 28.76 & 0.948 & 0.044 & 28.20 & \third{0.954} & \third{0.039} & \second{32.64} & \second{0.963} & \second{0.036} & \third{30.33} & 0.948 & 0.050 & \best{33.33} & \best{0.964} & \best{0.018} \\
    VRheadset & \third{29.99} & 0.948 & 0.044 & 29.38 & \third{0.950} & \third{0.039} & \second{31.23} & \best{0.959} & \second{0.036} & 27.92 & 0.938 & 0.059 & \best{32.08} & \best{0.959} & \best{0.020} \\
    \midrule
    Average & \third{29.80} & 0.949 & 0.043 & 29.46 & \third{0.954} & \third{0.039} & \second{31.88} & \second{0.962} & \second{0.034} & 28.93 & 0.941 & 0.055 & \best{33.47} & \best{0.963} & \best{0.019} \\
    \bottomrule
    \end{tabular}%
    }

    \vspace{4pt}

    \resizebox{\linewidth}{!}{%
    \begin{tabular}{p{2.3cm} ccc ccc ccc ccc ccc}
    \toprule
    \multirow{2}{*}{\makecell[l]{(e) \textbf{Google}\\ \textbf{Immersive}}} & \multicolumn{3}{c}{Gaussian4D} & \multicolumn{3}{c}{4DGS} & \multicolumn{3}{c}{STG} & \multicolumn{3}{c}{Ex4DGS} & \multicolumn{3}{c}{\textbf{Ours}} \\
    \cmidrule(lr){2-4} \cmidrule(lr){5-7} \cmidrule(lr){8-10} \cmidrule(lr){11-13} \cmidrule(lr){14-16}
     & PSNR$\uparrow$ & SSIM$\uparrow$ & LPIPS$\downarrow$ & PSNR$\uparrow$ & SSIM$\uparrow$ & LPIPS$\downarrow$ & PSNR$\uparrow$ & SSIM$\uparrow$ & LPIPS$\downarrow$ & PSNR$\uparrow$ & SSIM$\uparrow$ & LPIPS$\downarrow$ & PSNR$\uparrow$ & SSIM$\uparrow$ & LPIPS$\downarrow$ \\
    \midrule
    02\_Flames & 19.83 & 0.752 & 0.354 & \second{30.70} & 0.922 & 0.106 & \third{30.48} & \second{0.932} & \second{0.063} & 27.70 & \third{0.924} & \third{0.081} & \best{31.85} & \best{0.940} & \best{0.031} \\
    03\_Dog & 18.51 & 0.632 & 0.550 & \second{23.25} & \second{0.753} & \third{0.325} & 20.89 & 0.726 & 0.415 & \third{22.87} & \third{0.736} & \second{0.222} & \best{25.18} & \best{0.801} & \best{0.102} \\
    10\_Face\_Paint & 23.61 & 0.878 & 0.305 & \third{29.65} & 0.929 & 0.156 & \second{31.86} & \best{0.964} & \second{0.058} & 26.94 & \third{0.954} & \third{0.074} & \best{32.48} & \second{0.957} & \best{0.034} \\
    \midrule
    Average & 20.65 & 0.754 & 0.403 & \second{27.87} & 0.868 & 0.196 & \third{27.74} & \second{0.874} & \third{0.179} & 25.84 & \third{0.871} & \second{0.126} & \best{29.84} & \best{0.899} & \best{0.056} \\
    \bottomrule
    \end{tabular}%
    }
    \par \vspace{2pt}
\end{table*}

\begin{table}[!t]
\centering
\caption{Quantitative results and storage footprint on the \textit{ImViD} dataset (300 frames for each scene).}
\label{tab:efficiency}
\renewcommand{\arraystretch}{0.9}
\footnotesize
\resizebox{\linewidth}{!}{%
\begin{tabular}{llccccc}
\toprule
Method & & \makecell{Scene 1 \\ Opera} & \makecell{Scene 2 \\ Laboratory} & \makecell{Scene 5 \\ Rendition} & \makecell{Scene 6 \\ Puppy} & Average \\
\midrule
\multirow{2}{*}{Gaussian4D} & PSNR $\uparrow$ & \third{25.61} & \third{22.15} & 25.78 & 20.15 & \third{23.42} \\
& Size (MB) $\downarrow$ & \best{107} & \best{174} & \best{92} & \best{183} & \best{139.0} \\
\midrule
\multirow{2}{*}{4DGS} & PSNR $\uparrow$ & 25.01 & 21.02 & \second{26.23} & 21.03 & 23.32 \\
& Size (MB) $\downarrow$ & 1802 & 4832 & 1403 & 5233 & 3317.5 \\
\midrule
\multirow{2}{*}{STG} & PSNR $\uparrow$ & \second{26.30} & \second{23.40} & \third{26.21} & \second{21.25} & \second{24.29} \\
& Size (MB) $\downarrow$ & 540 & 542 & \third{366} & 1493 & 735.3 \\
\midrule
\multirow{2}{*}{Ex4DGS} & PSNR $\uparrow$ & 23.08 & 20.29 & 24.73 & \third{21.21} & 22.33 \\
& Size (MB) $\downarrow$ & 2449 & 714 & 674 & 8918 & 3188.8 \\
\midrule
\multirow{2}{*}{\textbf{Ours}} & PSNR $\uparrow$ & \best{33.51} & \best{31.10} & \best{27.84} & \best{21.53} & \best{28.50} \\
& Size (MB) $\downarrow$ & \second{210} & \second{209} & \second{140} & \second{581} & \second{285.0} \\
\bottomrule
\end{tabular}%
}
\end{table}

\subsubsection{Qualitative Assessment}
\begin{figure*}[!t]
    \centering
    \setlength{\tabcolsep}{1pt}
    \renewcommand{\arraystretch}{0.5}

    \begin{tabular}{c c c c c c c}
        & \small{Gaussian4D} & \small{4DGS} & \small{STG} & \small{Ex4DGS} & \textbf{\small{Ours}} & \small{Ground Truth}  \\
        
        \rotatebox[origin=c]{90}{\small \textbf{Opera}} &
        \parbox{0.155\textwidth}{\begin{minipage}{\linewidth}\centering\setlength{\fboxsep}{0pt}\setlength{\fboxrule}{1.2pt}\includegraphics[width=\linewidth]{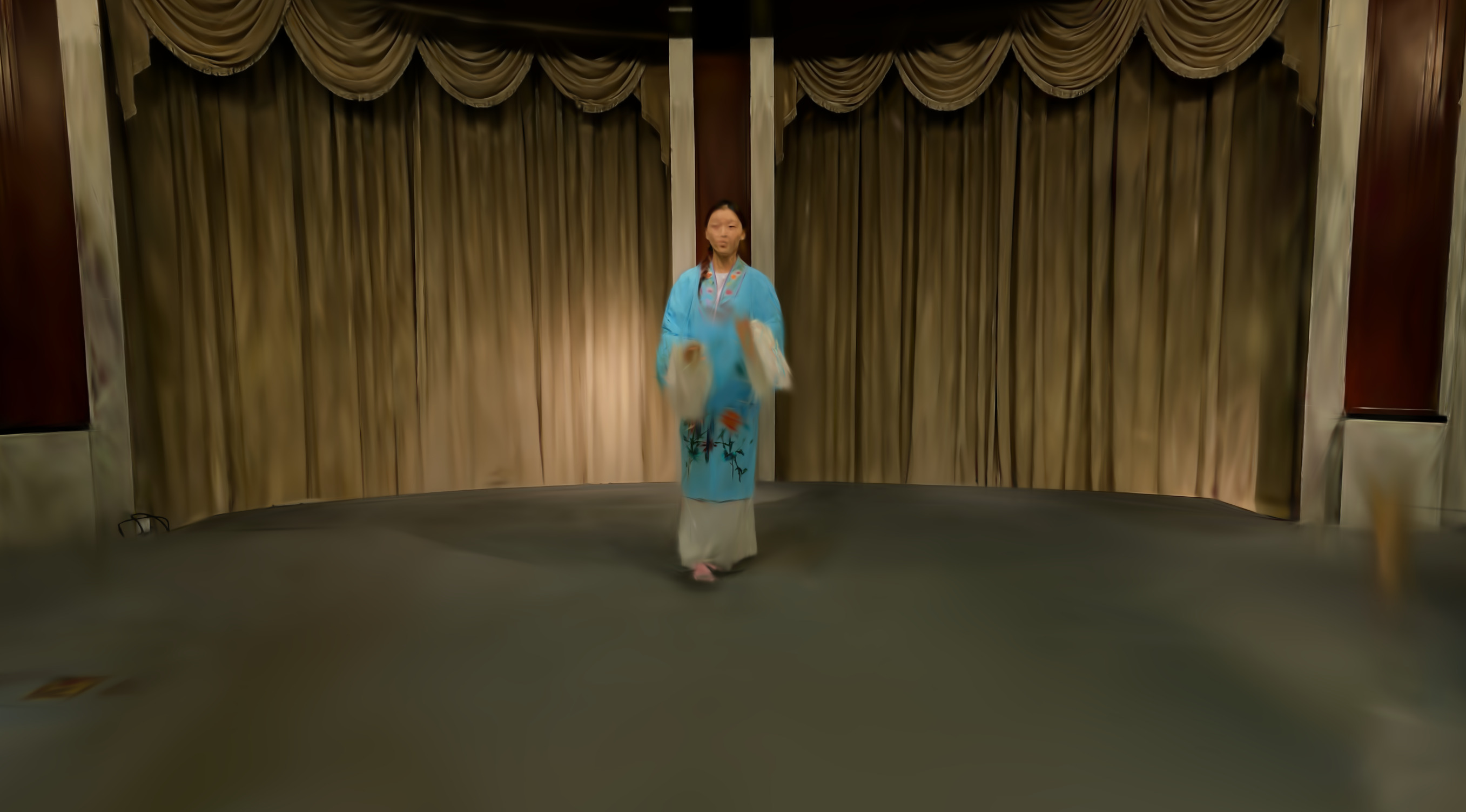} \\\vspace{1.5pt}\fcolorbox[HTML]{00FFFF}{00FFFF}{\includegraphics[width=\dimexpr0.49\linewidth-2\fboxrule\relax]{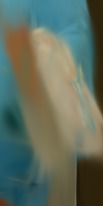}}\hfill\fcolorbox[HTML]{FF69B4}{FF69B4}{\includegraphics[width=\dimexpr0.49\linewidth-2\fboxrule\relax]{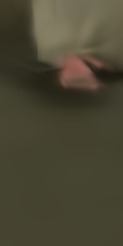}}\end{minipage}} &
        \parbox{0.155\textwidth}{\begin{minipage}{\linewidth}\centering\setlength{\fboxsep}{0pt}\setlength{\fboxrule}{1.2pt}\includegraphics[width=\linewidth]{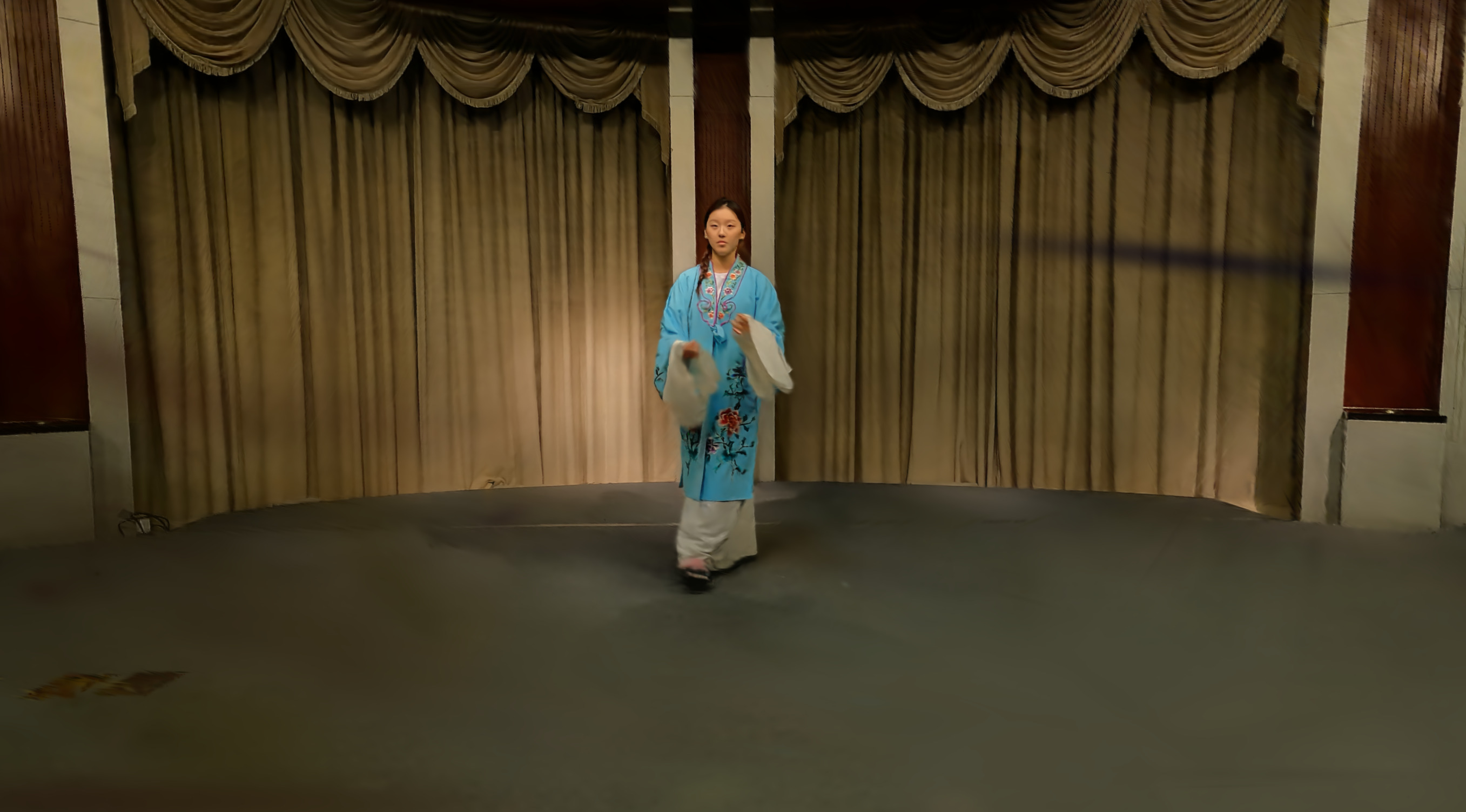} \\\vspace{1.5pt}\fcolorbox[HTML]{00FFFF}{00FFFF}{\includegraphics[width=\dimexpr0.49\linewidth-2\fboxrule\relax]{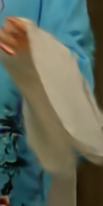}}\hfill\fcolorbox[HTML]{FF69B4}{FF69B4}{\includegraphics[width=\dimexpr0.49\linewidth-2\fboxrule\relax]{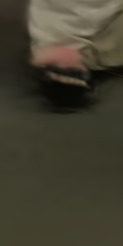}}\end{minipage}} &
        \parbox{0.155\textwidth}{\begin{minipage}{\linewidth}\centering\setlength{\fboxsep}{0pt}\setlength{\fboxrule}{1.2pt}\includegraphics[width=\linewidth]{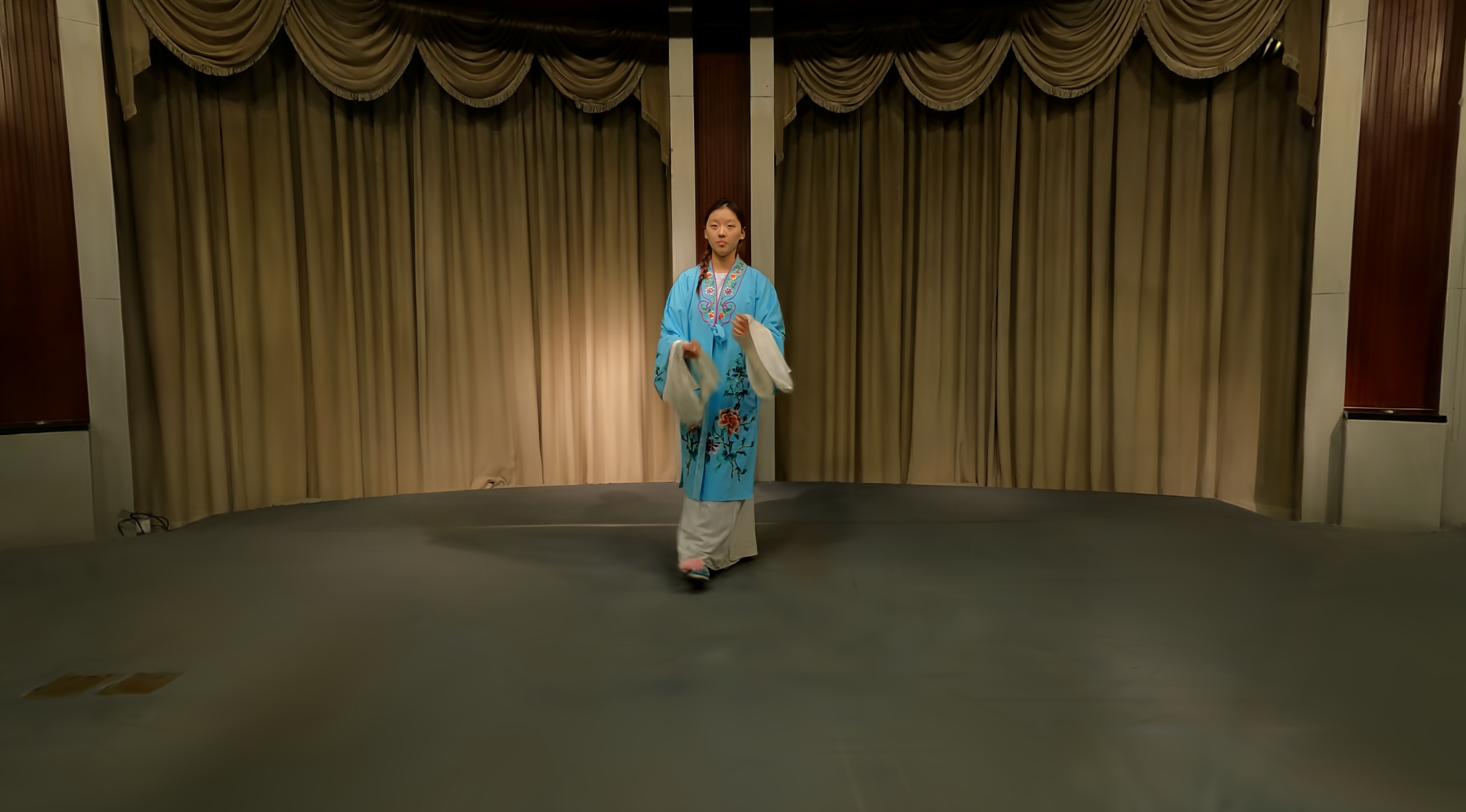} \\\vspace{1.5pt}\fcolorbox[HTML]{00FFFF}{00FFFF}{\includegraphics[width=\dimexpr0.49\linewidth-2\fboxrule\relax]{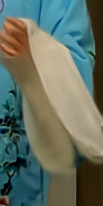}}\hfill\fcolorbox[HTML]{FF69B4}{FF69B4}{\includegraphics[width=\dimexpr0.49\linewidth-2\fboxrule\relax]{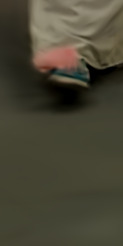}}\end{minipage}} &
        \parbox{0.155\textwidth}{\begin{minipage}{\linewidth}\centering\setlength{\fboxsep}{0pt}\setlength{\fboxrule}{1.2pt}\includegraphics[width=\linewidth]{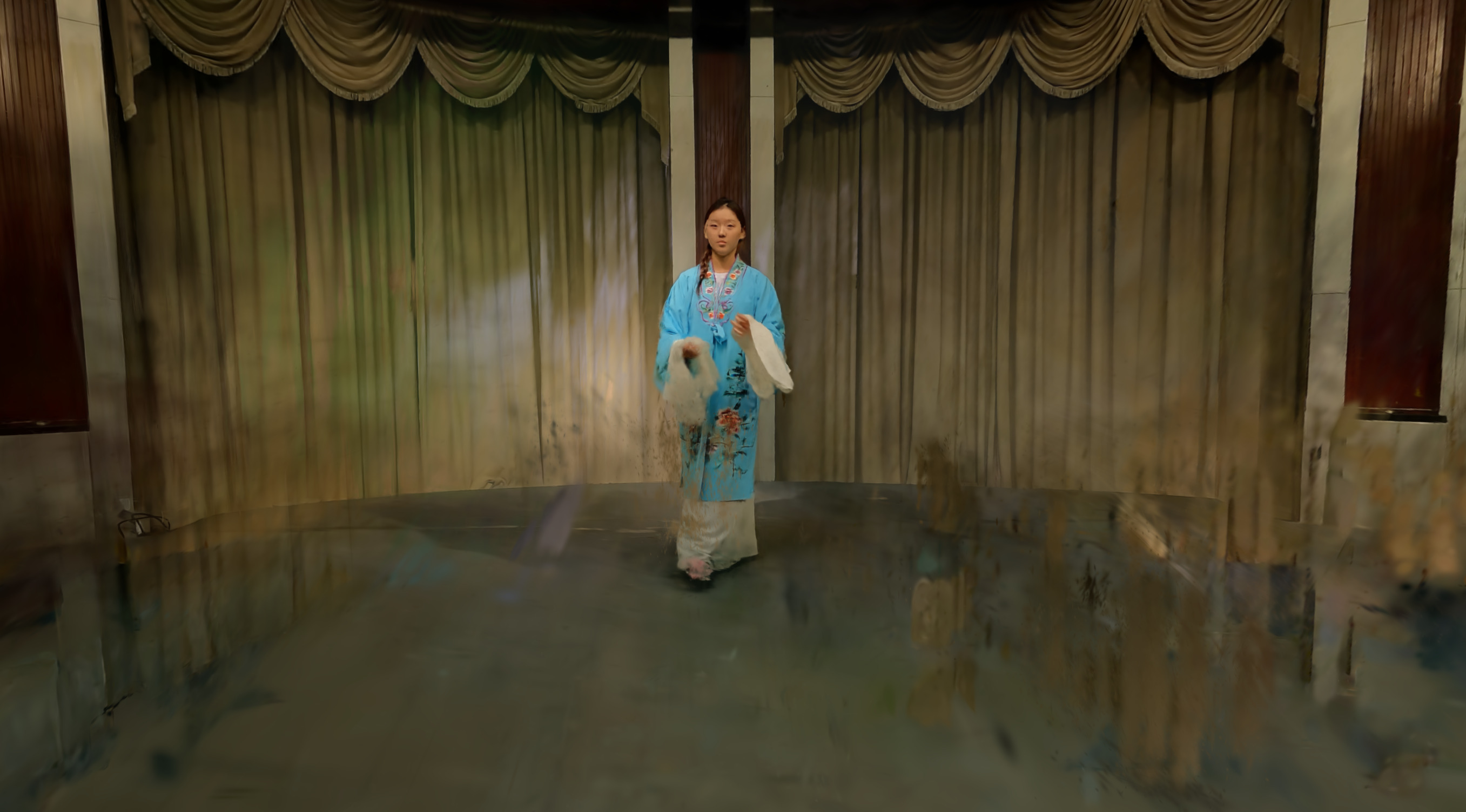} \\\vspace{1.5pt}\fcolorbox[HTML]{00FFFF}{00FFFF}{\includegraphics[width=\dimexpr0.49\linewidth-2\fboxrule\relax]{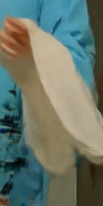}}\hfill\fcolorbox[HTML]{FF69B4}{FF69B4}{\includegraphics[width=\dimexpr0.49\linewidth-2\fboxrule\relax]{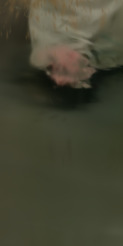}}\end{minipage}} &
        \parbox{0.155\textwidth}{\begin{minipage}{\linewidth}\centering\setlength{\fboxsep}{0pt}\setlength{\fboxrule}{1.2pt}\includegraphics[width=\linewidth]{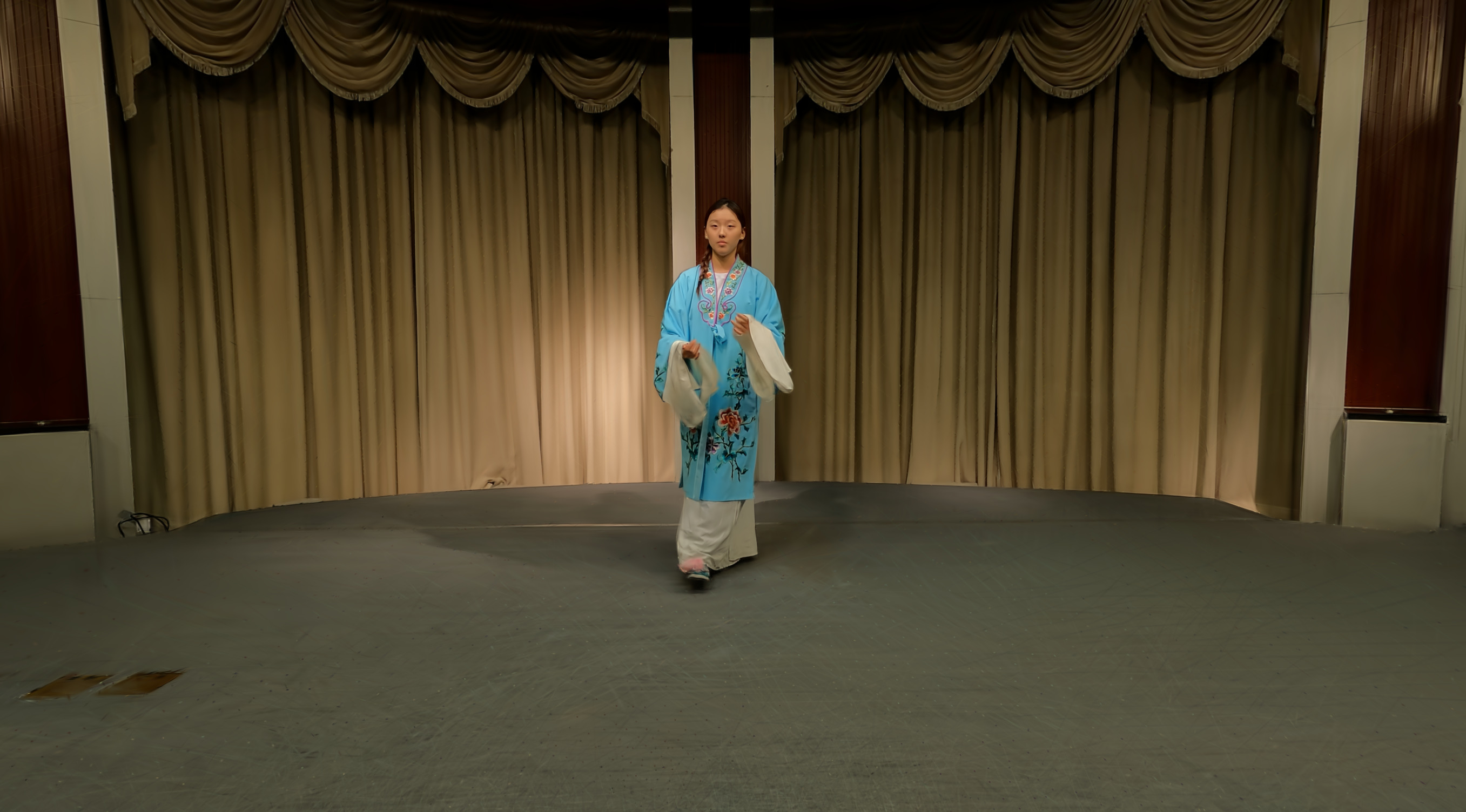} \\\vspace{1.5pt}\fcolorbox[HTML]{00FFFF}{00FFFF}{\includegraphics[width=\dimexpr0.49\linewidth-2\fboxrule\relax]{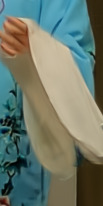}}\hfill\fcolorbox[HTML]{FF69B4}{FF69B4}{\includegraphics[width=\dimexpr0.49\linewidth-2\fboxrule\relax]{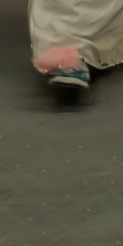}}\end{minipage}} &
        \parbox{0.155\textwidth}{\begin{minipage}{\linewidth}\centering\setlength{\fboxsep}{0pt}\setlength{\fboxrule}{1.2pt}\includegraphics[width=\linewidth]{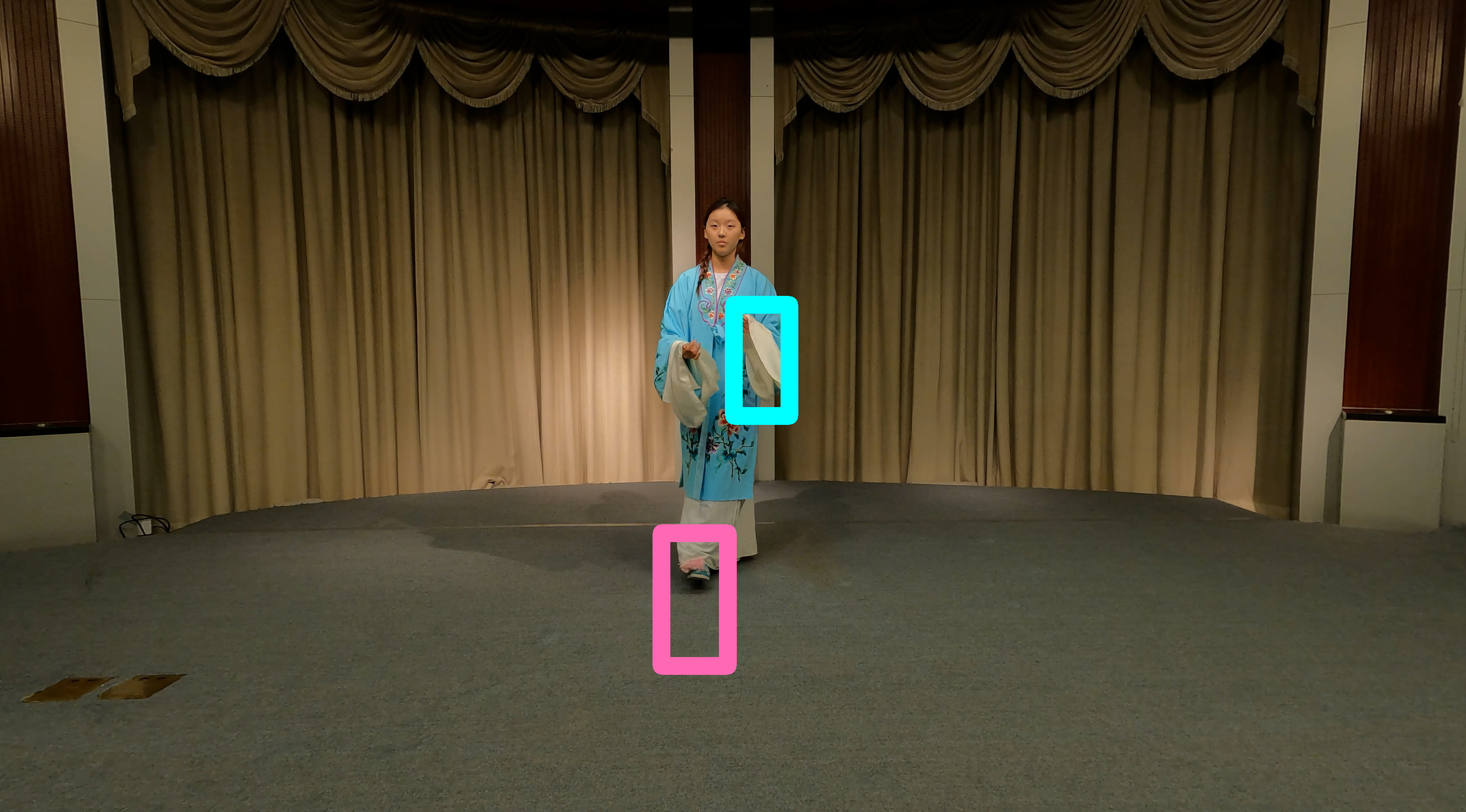} \\\vspace{1.5pt}\fcolorbox[HTML]{00FFFF}{00FFFF}{\includegraphics[width=\dimexpr0.49\linewidth-2\fboxrule\relax]{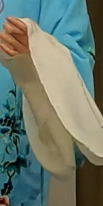}}\hfill\fcolorbox[HTML]{FF69B4}{FF69B4}{\includegraphics[width=\dimexpr0.49\linewidth-2\fboxrule\relax]{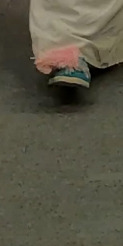}}\end{minipage}} \\
        
        \multicolumn{5}{c}{\vspace{3mm}} \\

        \rotatebox[origin=c]{90}{\small \textbf{Laboratory}} &
        \parbox{0.155\textwidth}{\begin{minipage}{\linewidth}\centering\setlength{\fboxsep}{0pt}\setlength{\fboxrule}{1.2pt}\includegraphics[width=\linewidth]{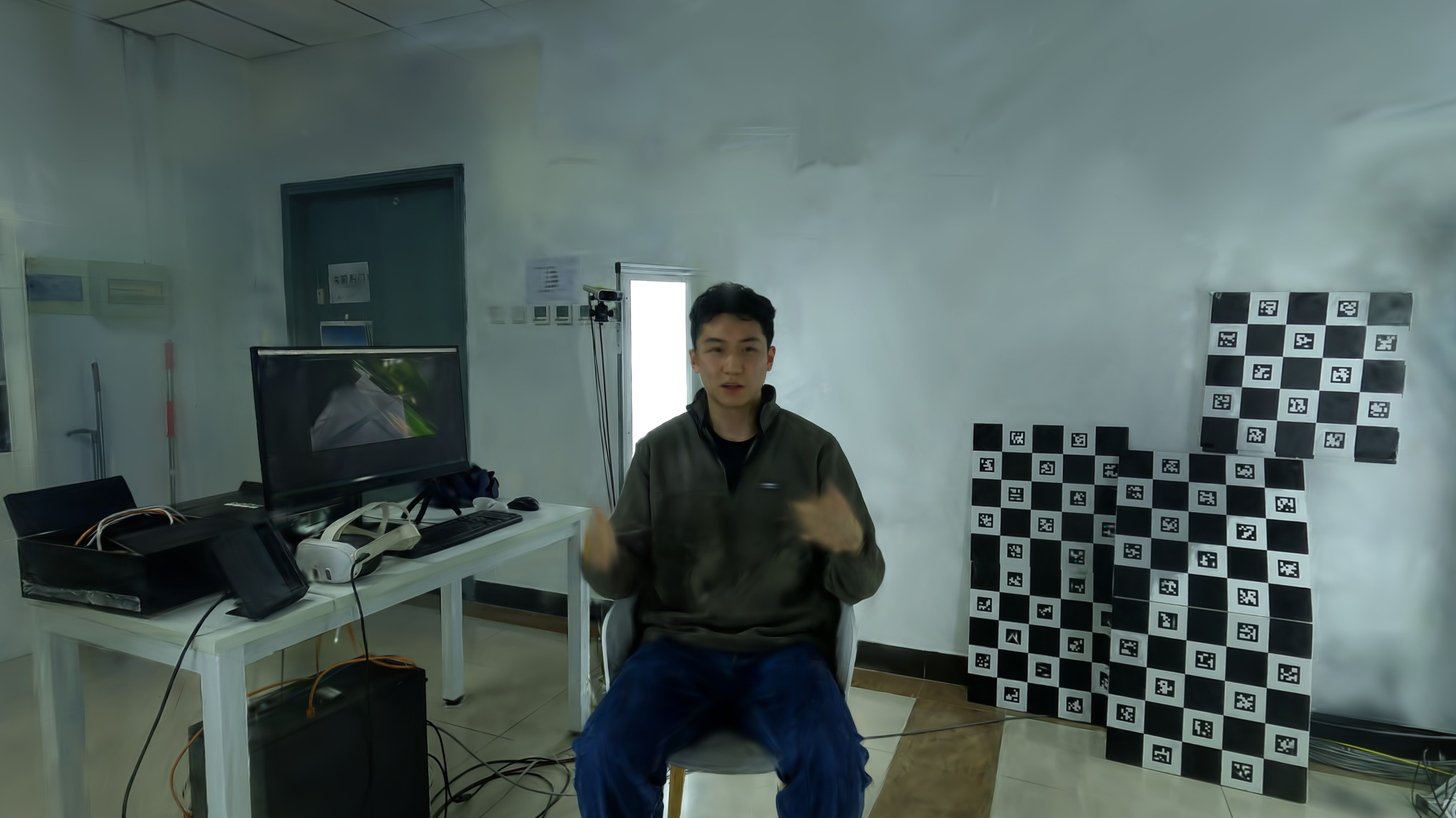} \\\vspace{1.5pt}\fcolorbox[HTML]{00FFFF}{00FFFF}{\includegraphics[width=\dimexpr\linewidth-2\fboxrule\relax]{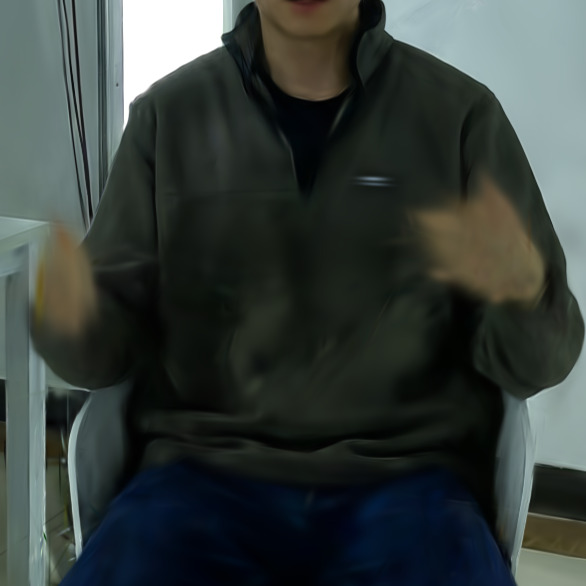}}\end{minipage}} &
        \parbox{0.155\textwidth}{\begin{minipage}{\linewidth}\centering\setlength{\fboxsep}{0pt}\setlength{\fboxrule}{1.2pt}\includegraphics[width=\linewidth]{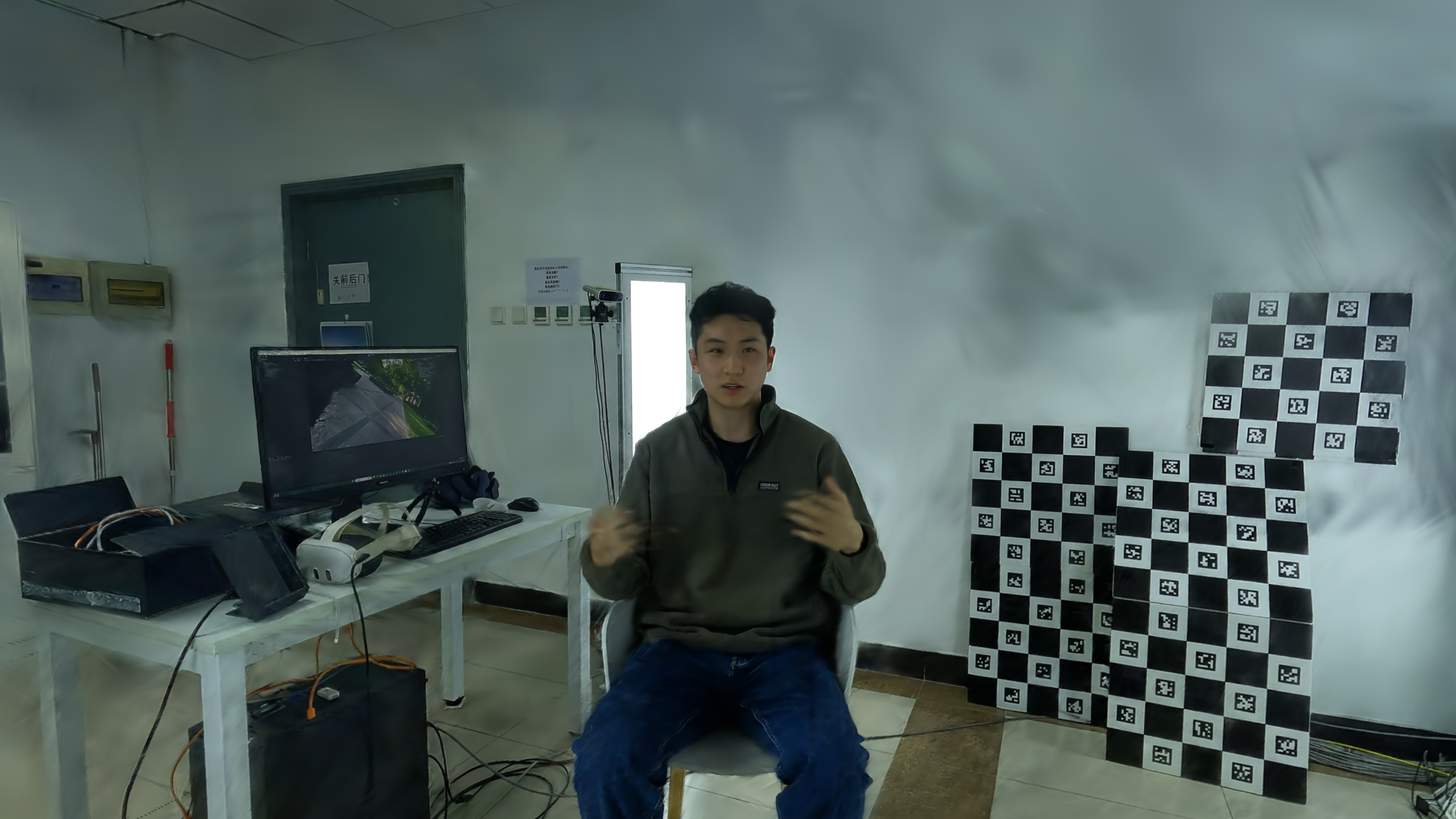} \\\vspace{1.5pt}\fcolorbox[HTML]{00FFFF}{00FFFF}{\includegraphics[width=\dimexpr\linewidth-2\fboxrule\relax]{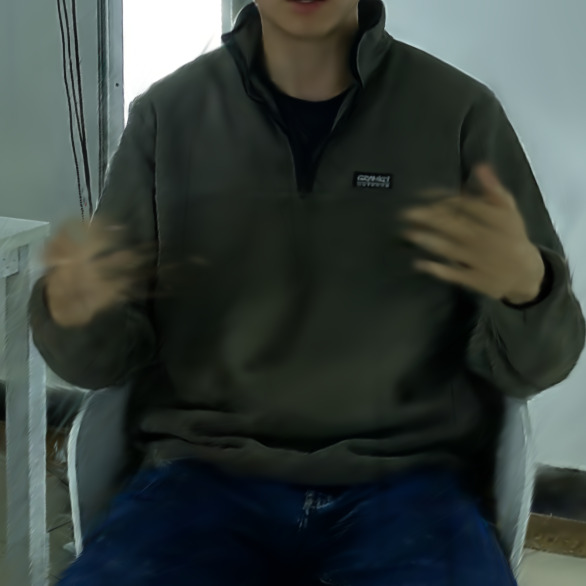}}\end{minipage}} &
        \parbox{0.155\textwidth}{\begin{minipage}{\linewidth}\centering\setlength{\fboxsep}{0pt}\setlength{\fboxrule}{1.2pt}\includegraphics[width=\linewidth]{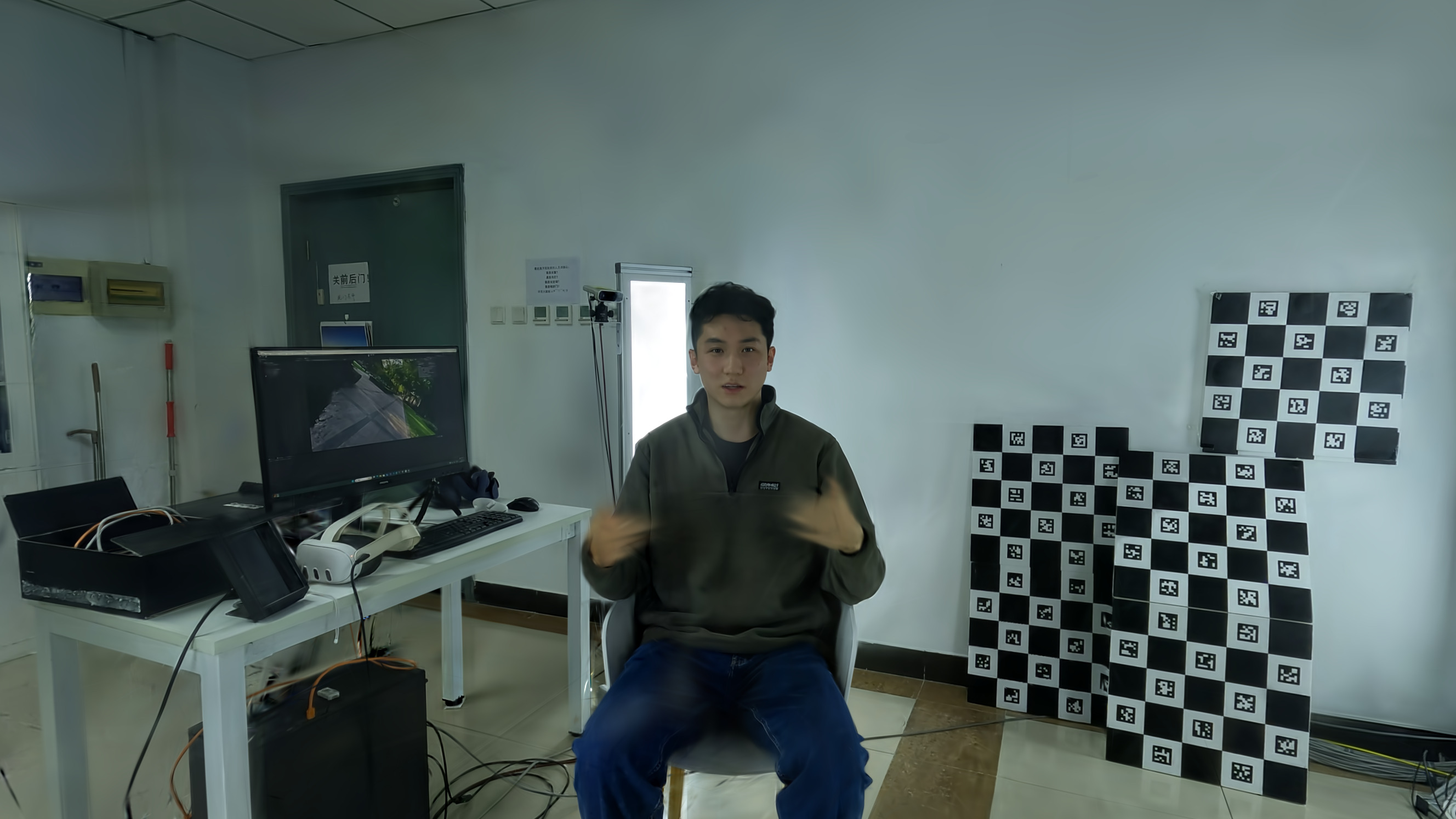} \\\vspace{1.5pt}\fcolorbox[HTML]{00FFFF}{00FFFF}{\includegraphics[width=\dimexpr\linewidth-2\fboxrule\relax]{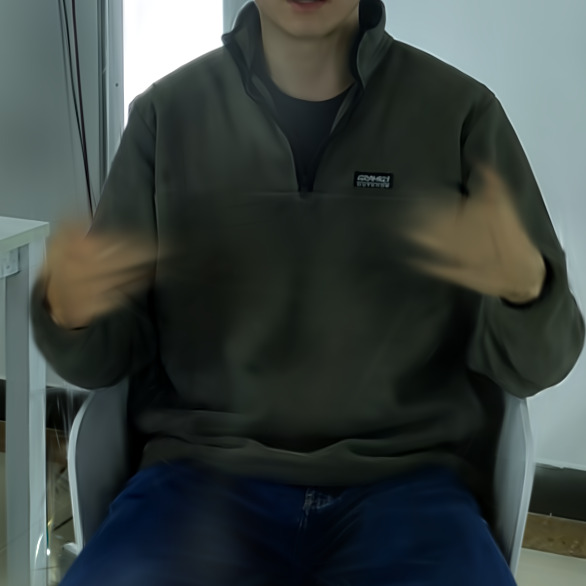}}\end{minipage}} &
        \parbox{0.155\textwidth}{\begin{minipage}{\linewidth}\centering\setlength{\fboxsep}{0pt}\setlength{\fboxrule}{1.2pt}\includegraphics[width=\linewidth]{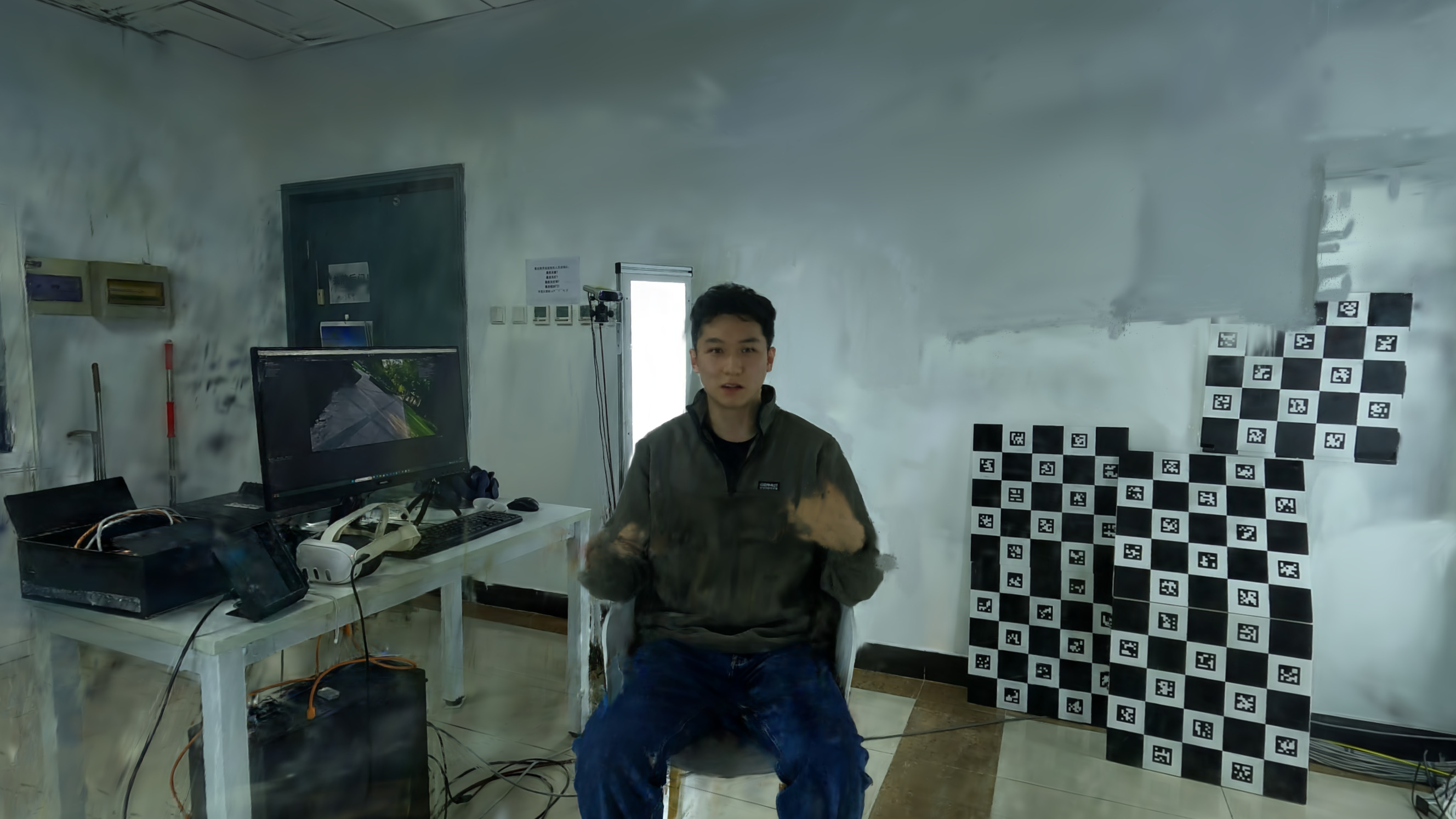} \\\vspace{1.5pt}\fcolorbox[HTML]{00FFFF}{00FFFF}{\includegraphics[width=\dimexpr\linewidth-2\fboxrule\relax]{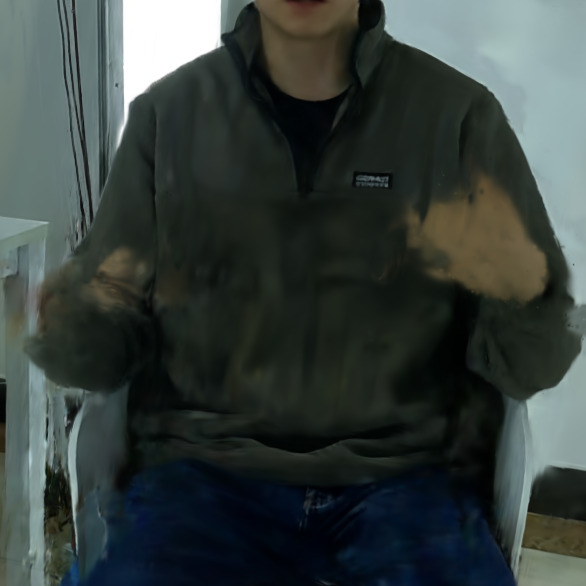}}\end{minipage}} &
        \parbox{0.155\textwidth}{\begin{minipage}{\linewidth}\centering\setlength{\fboxsep}{0pt}\setlength{\fboxrule}{1.2pt}\includegraphics[width=\linewidth]{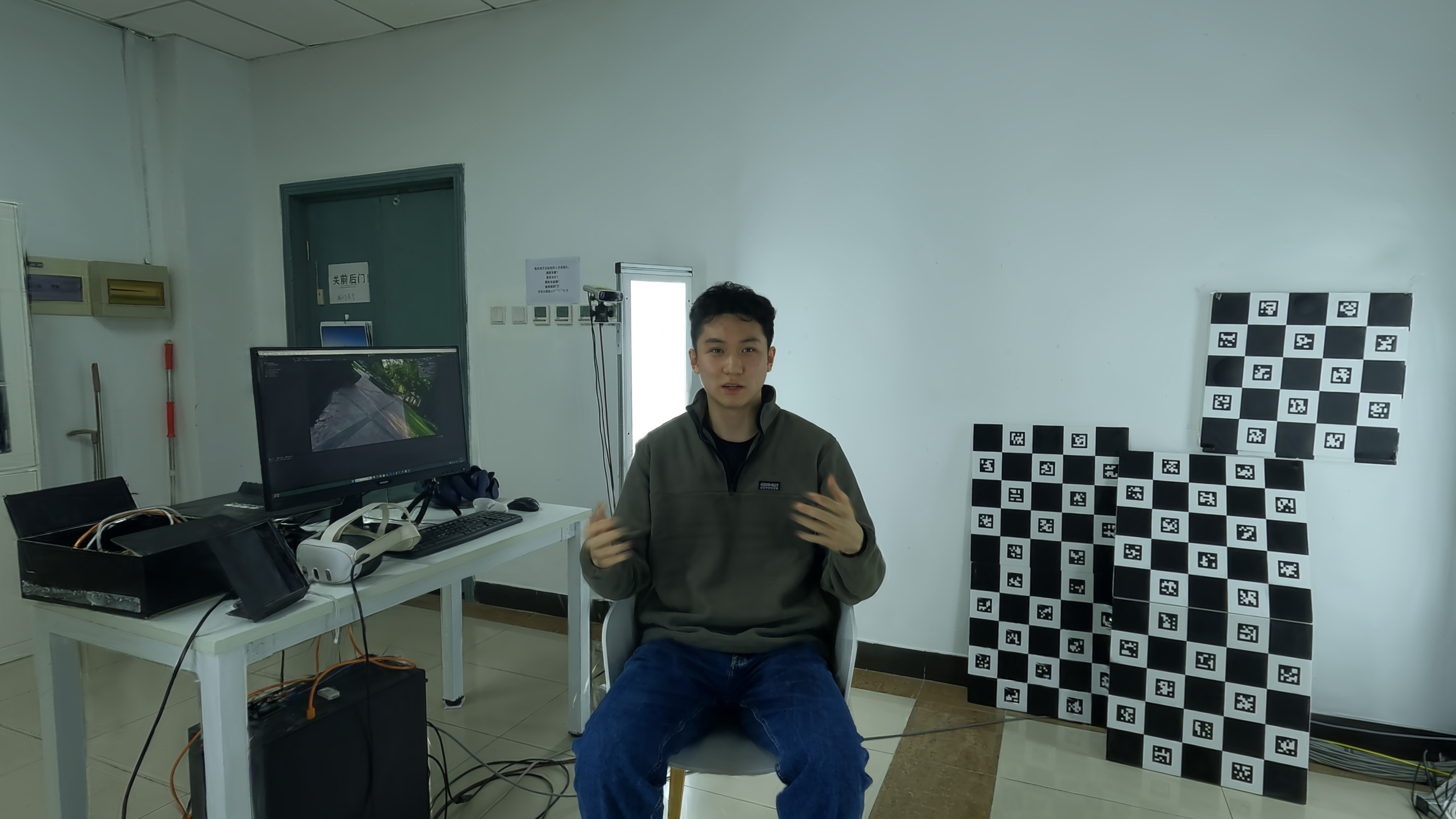} \\\vspace{1.5pt}\fcolorbox[HTML]{00FFFF}{00FFFF}{\includegraphics[width=\dimexpr\linewidth-2\fboxrule\relax]{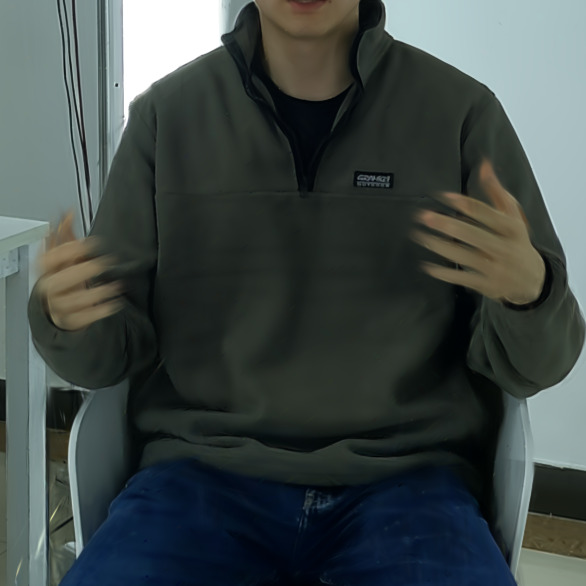}}\end{minipage}} &
        \parbox{0.155\textwidth}{\begin{minipage}{\linewidth}\centering\setlength{\fboxsep}{0pt}\setlength{\fboxrule}{1.2pt}\includegraphics[width=\linewidth]{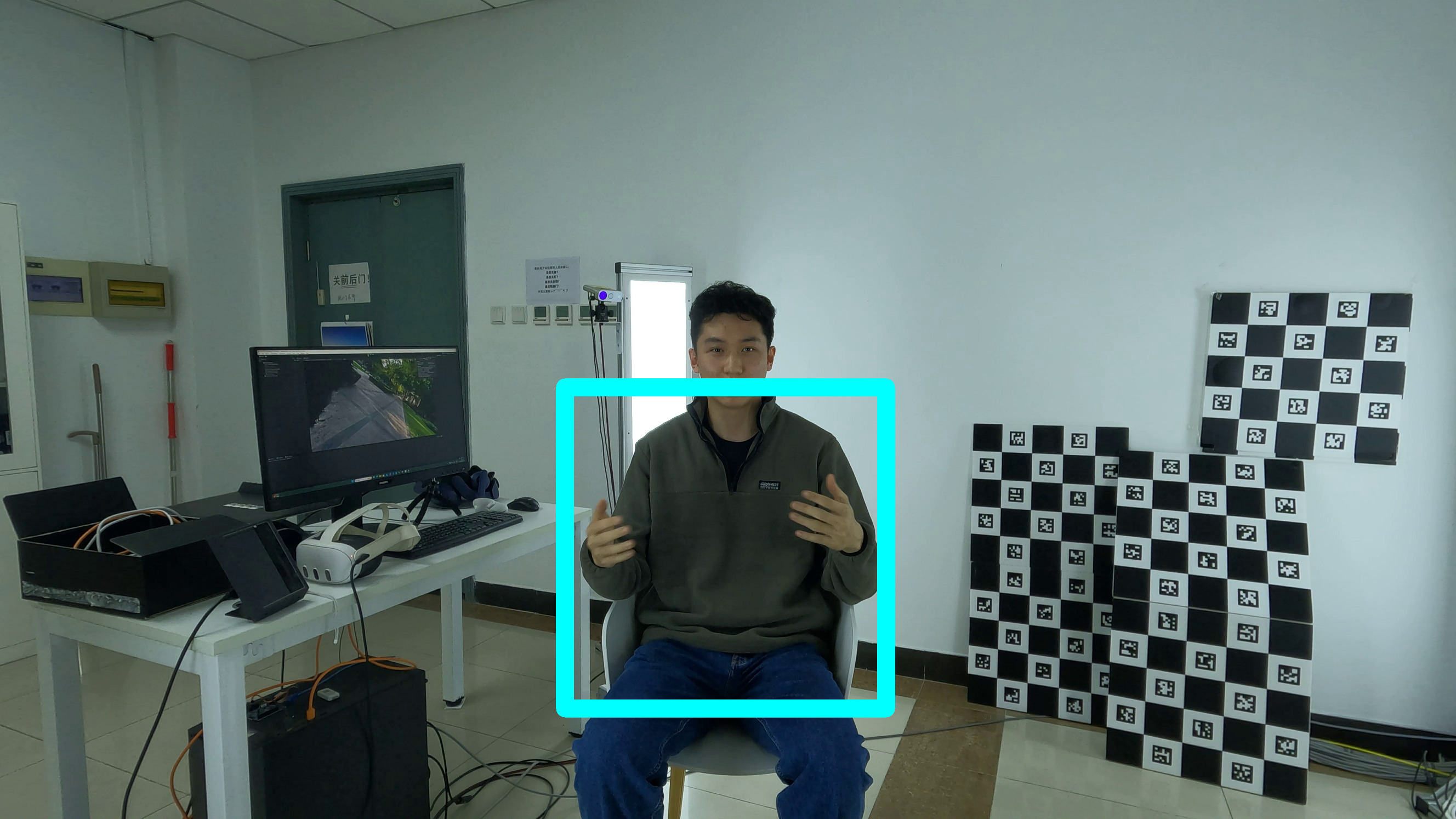} \\\vspace{1.5pt}\fcolorbox[HTML]{00FFFF}{00FFFF}{\includegraphics[width=\dimexpr\linewidth-2\fboxrule\relax]{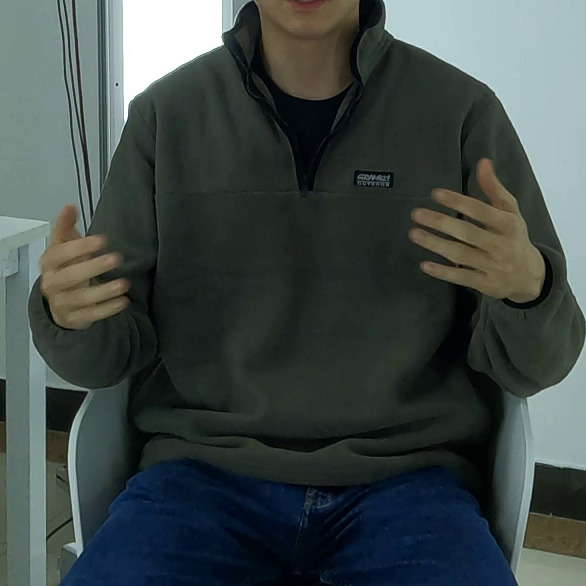}}\end{minipage}} \\

        \multicolumn{5}{c}{\vspace{3mm}} \\

        \rotatebox[origin=c]{90}{\small \textbf{Rendition}} &
        \parbox{0.155\textwidth}{\begin{minipage}{\linewidth}\centering\setlength{\fboxsep}{0pt}\setlength{\fboxrule}{1.2pt}\includegraphics[width=\linewidth]{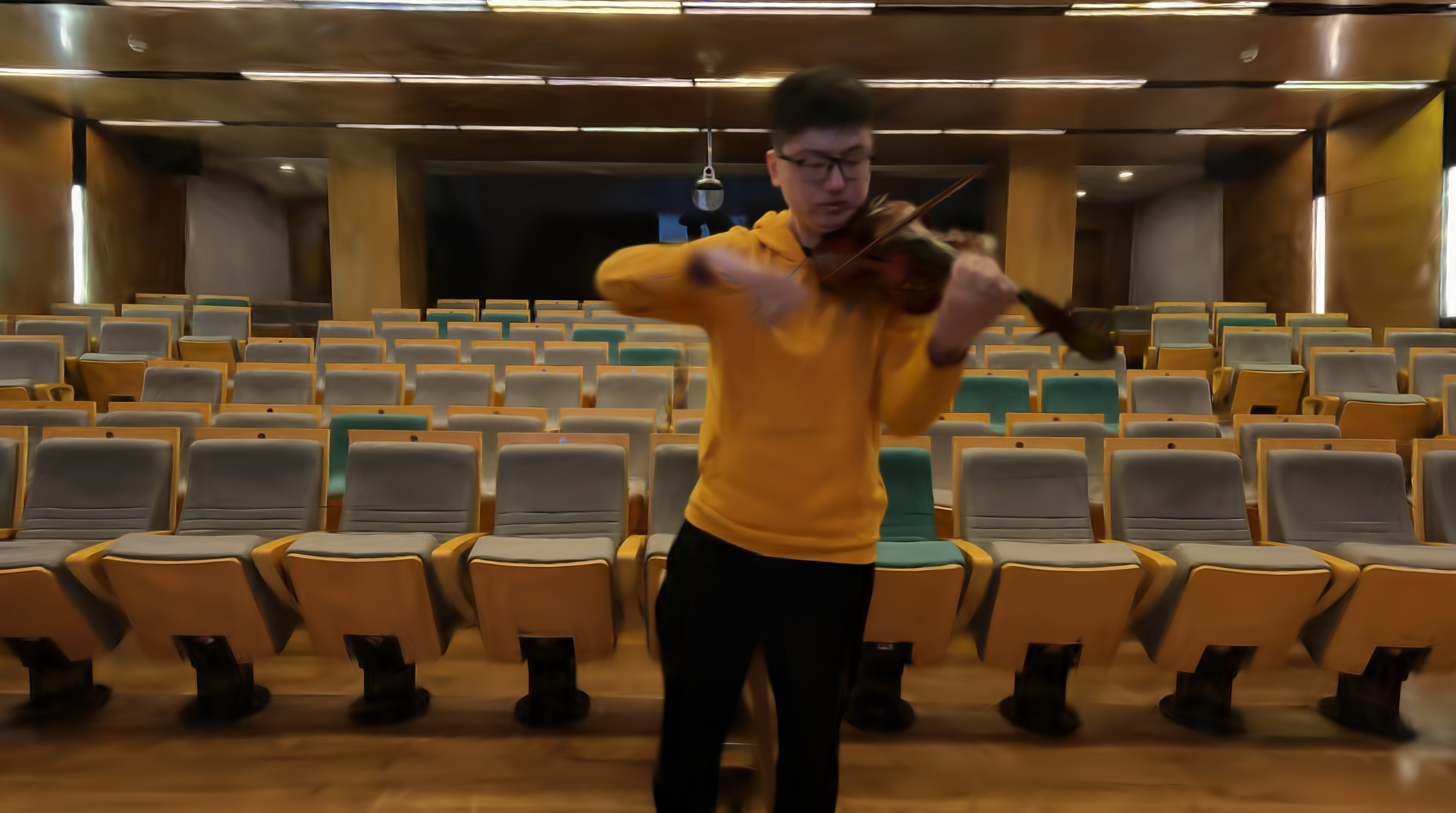} \\\vspace{1.5pt}\fcolorbox[HTML]{00FFFF}{00FFFF}{\includegraphics[width=\dimexpr\linewidth-2\fboxrule\relax]{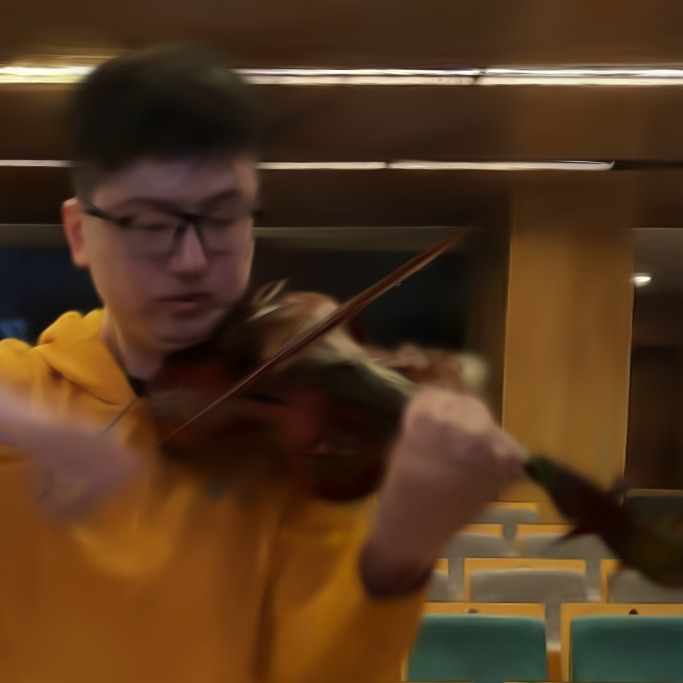}}\end{minipage}} &
        \parbox{0.155\textwidth}{\begin{minipage}{\linewidth}\centering\setlength{\fboxsep}{0pt}\setlength{\fboxrule}{1.2pt}\includegraphics[width=\linewidth]{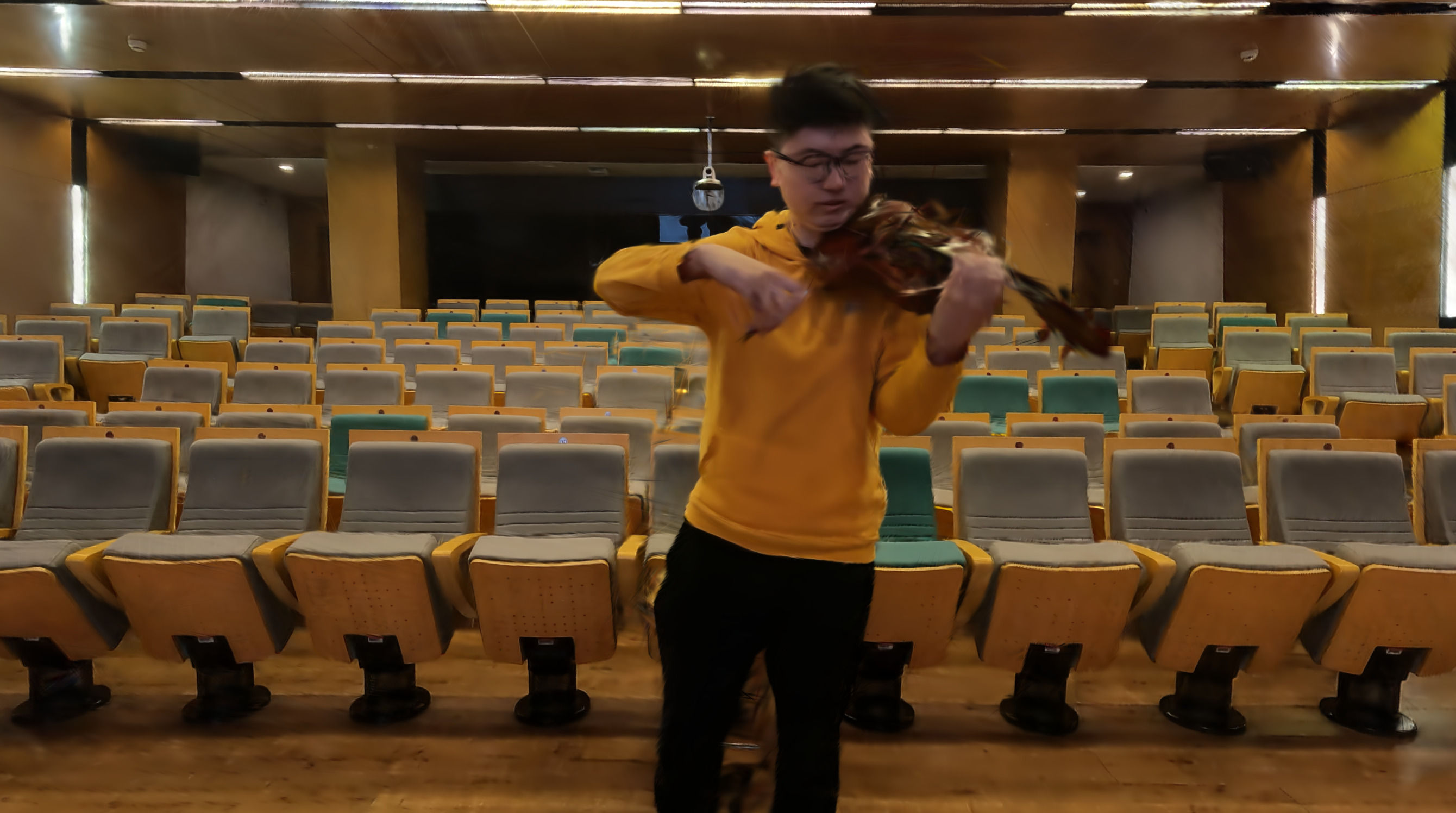} \\\vspace{1.5pt}\fcolorbox[HTML]{00FFFF}{00FFFF}{\includegraphics[width=\dimexpr\linewidth-2\fboxrule\relax]{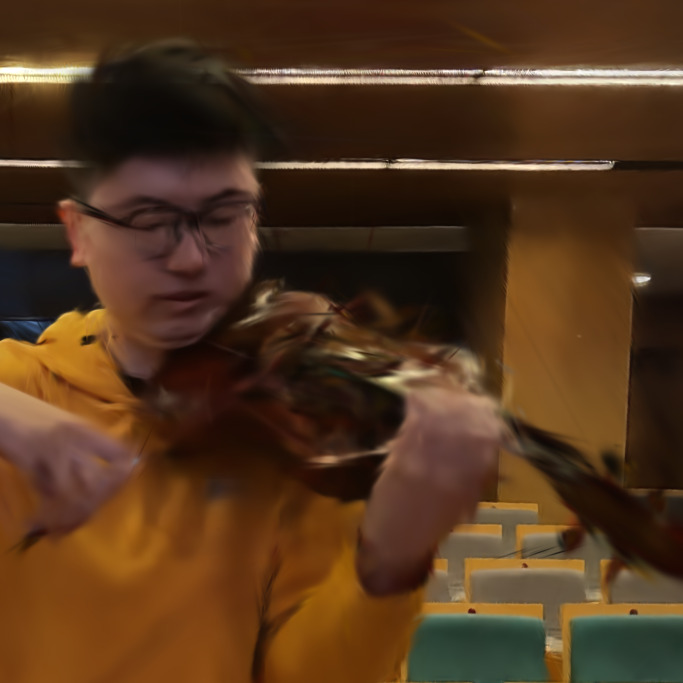}}\end{minipage}} &
        \parbox{0.155\textwidth}{\begin{minipage}{\linewidth}\centering\setlength{\fboxsep}{0pt}\setlength{\fboxrule}{1.2pt}\includegraphics[width=\linewidth]{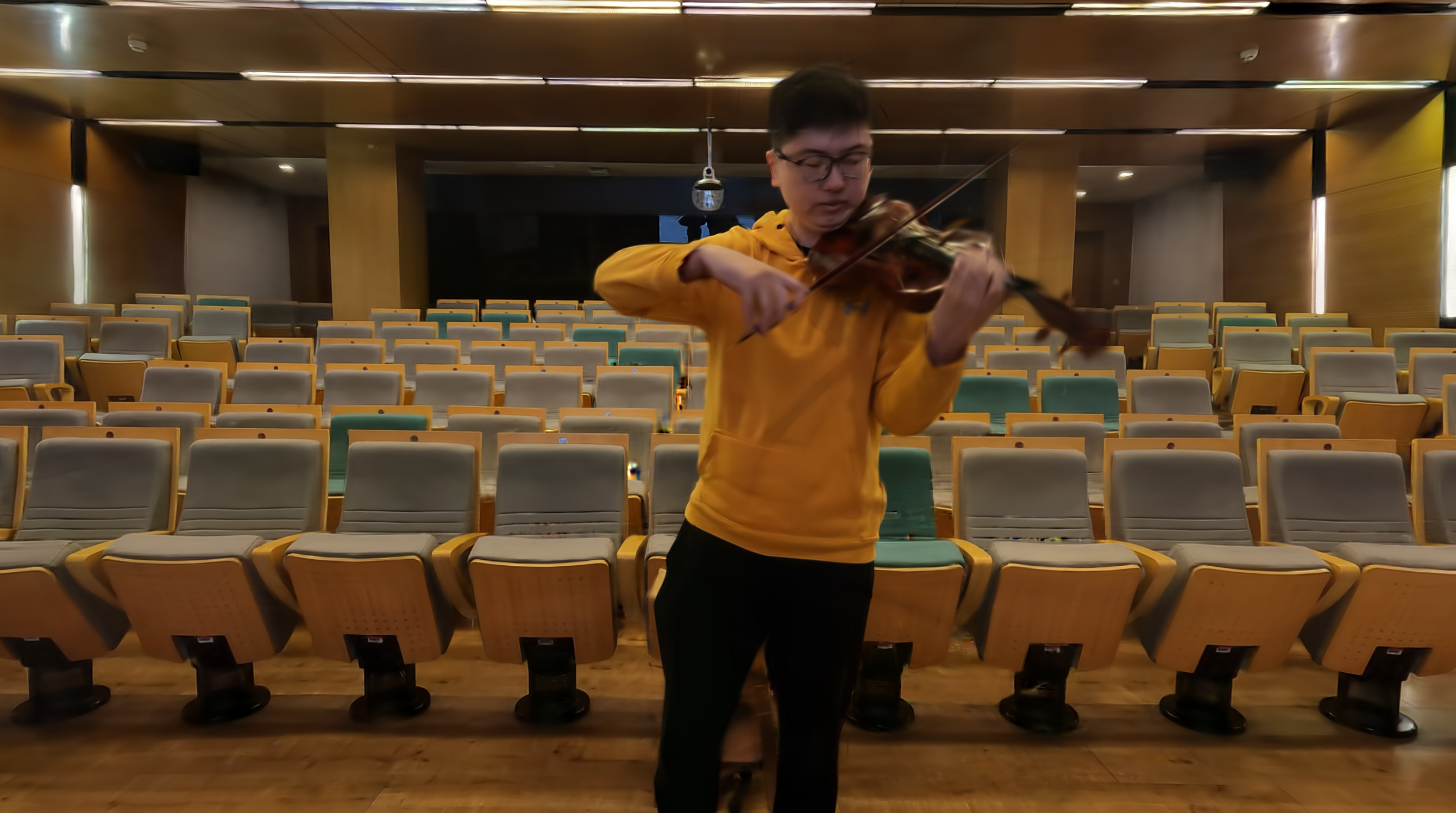} \\\vspace{1.5pt}\fcolorbox[HTML]{00FFFF}{00FFFF}{\includegraphics[width=\dimexpr\linewidth-2\fboxrule\relax]{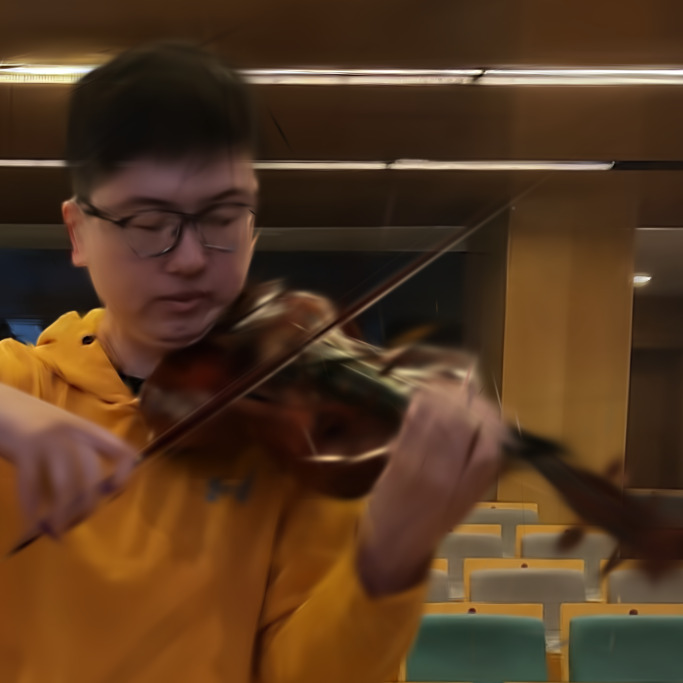}}\end{minipage}} &
        \parbox{0.155\textwidth}{\begin{minipage}{\linewidth}\centering\setlength{\fboxsep}{0pt}\setlength{\fboxrule}{1.2pt}\includegraphics[width=\linewidth]{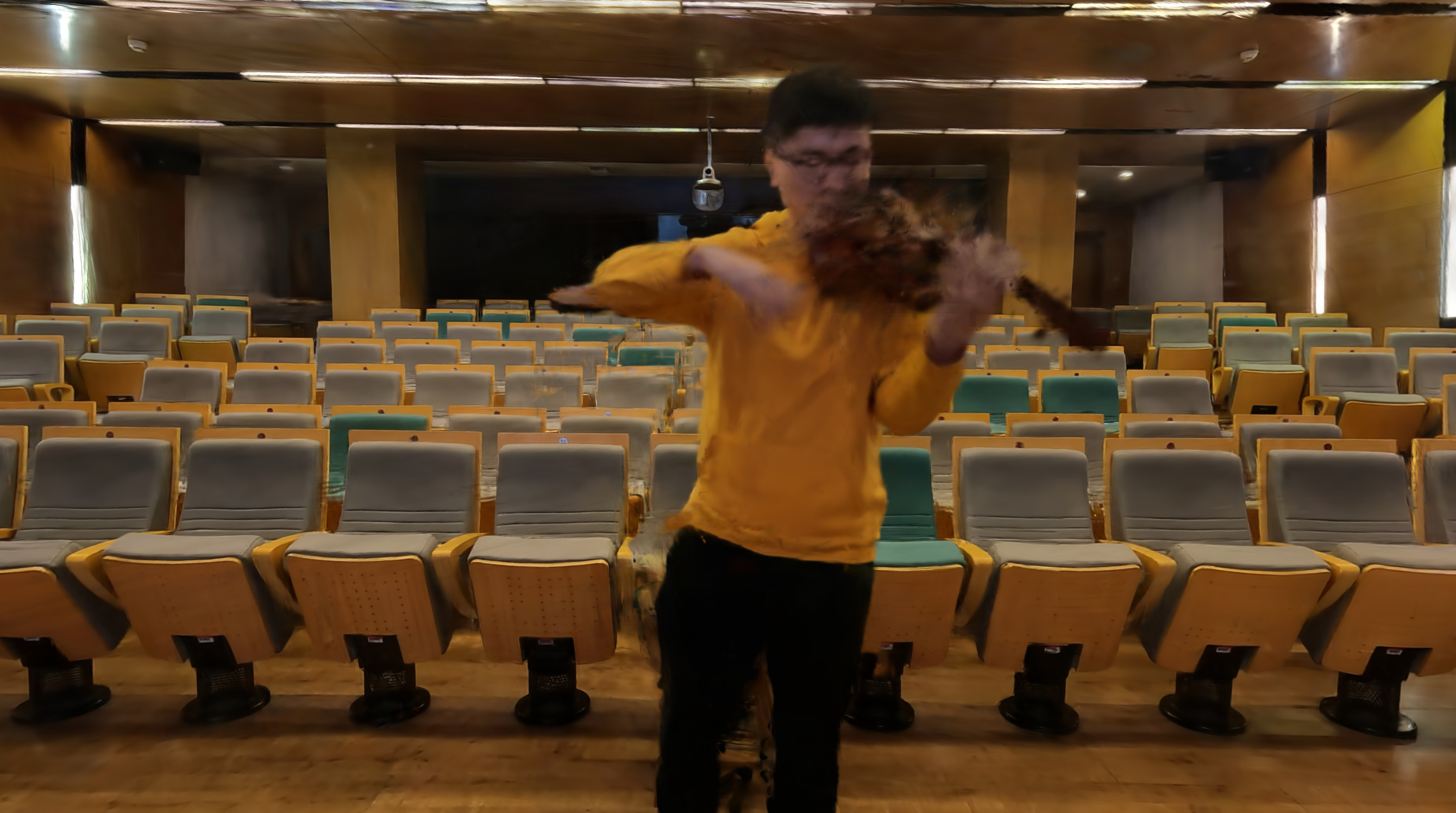} \\\vspace{1.5pt}\fcolorbox[HTML]{00FFFF}{00FFFF}{\includegraphics[width=\dimexpr\linewidth-2\fboxrule\relax]{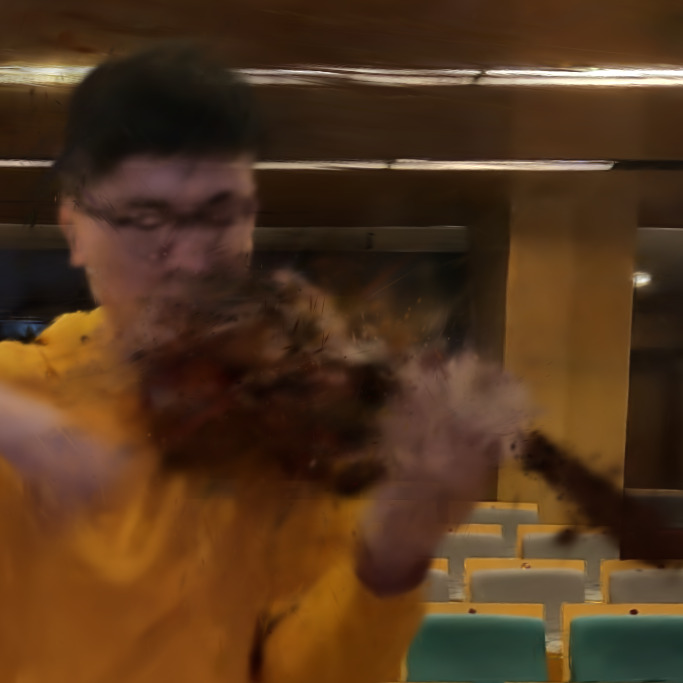}}\end{minipage}} &
        \parbox{0.155\textwidth}{\begin{minipage}{\linewidth}\centering\setlength{\fboxsep}{0pt}\setlength{\fboxrule}{1.2pt}\includegraphics[width=\linewidth]{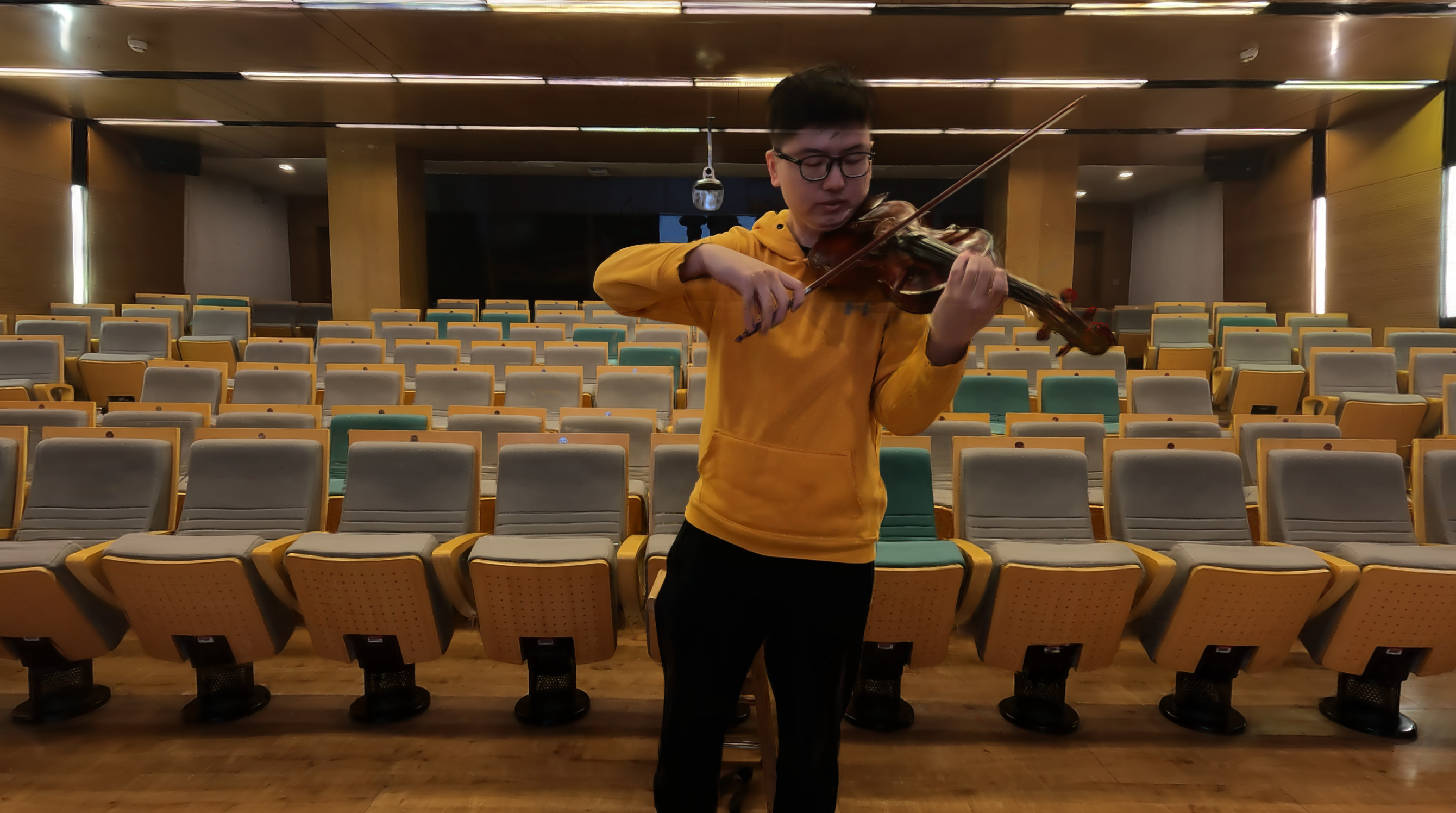} \\\vspace{1.5pt}\fcolorbox[HTML]{00FFFF}{00FFFF}{\includegraphics[width=\dimexpr\linewidth-2\fboxrule\relax]{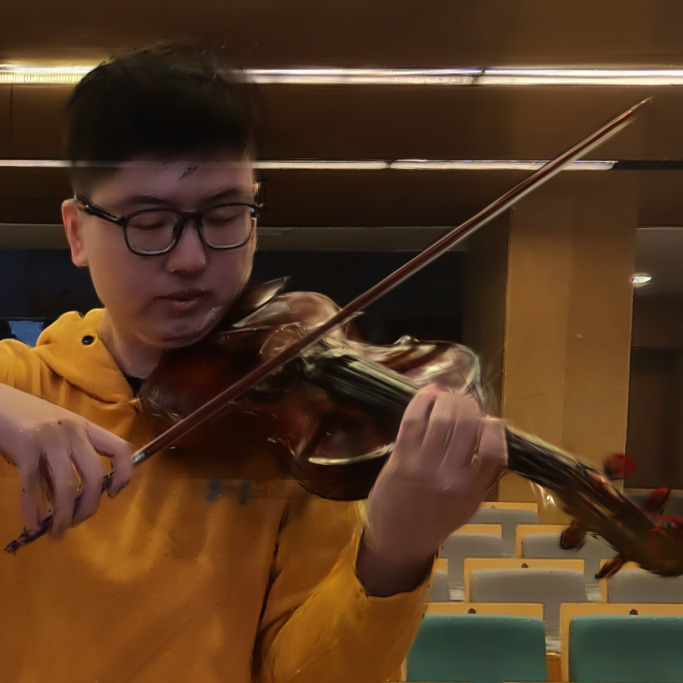}}\end{minipage}} &
        \parbox{0.155\textwidth}{\begin{minipage}{\linewidth}\centering\setlength{\fboxsep}{0pt}\setlength{\fboxrule}{1.2pt}\includegraphics[width=\linewidth]{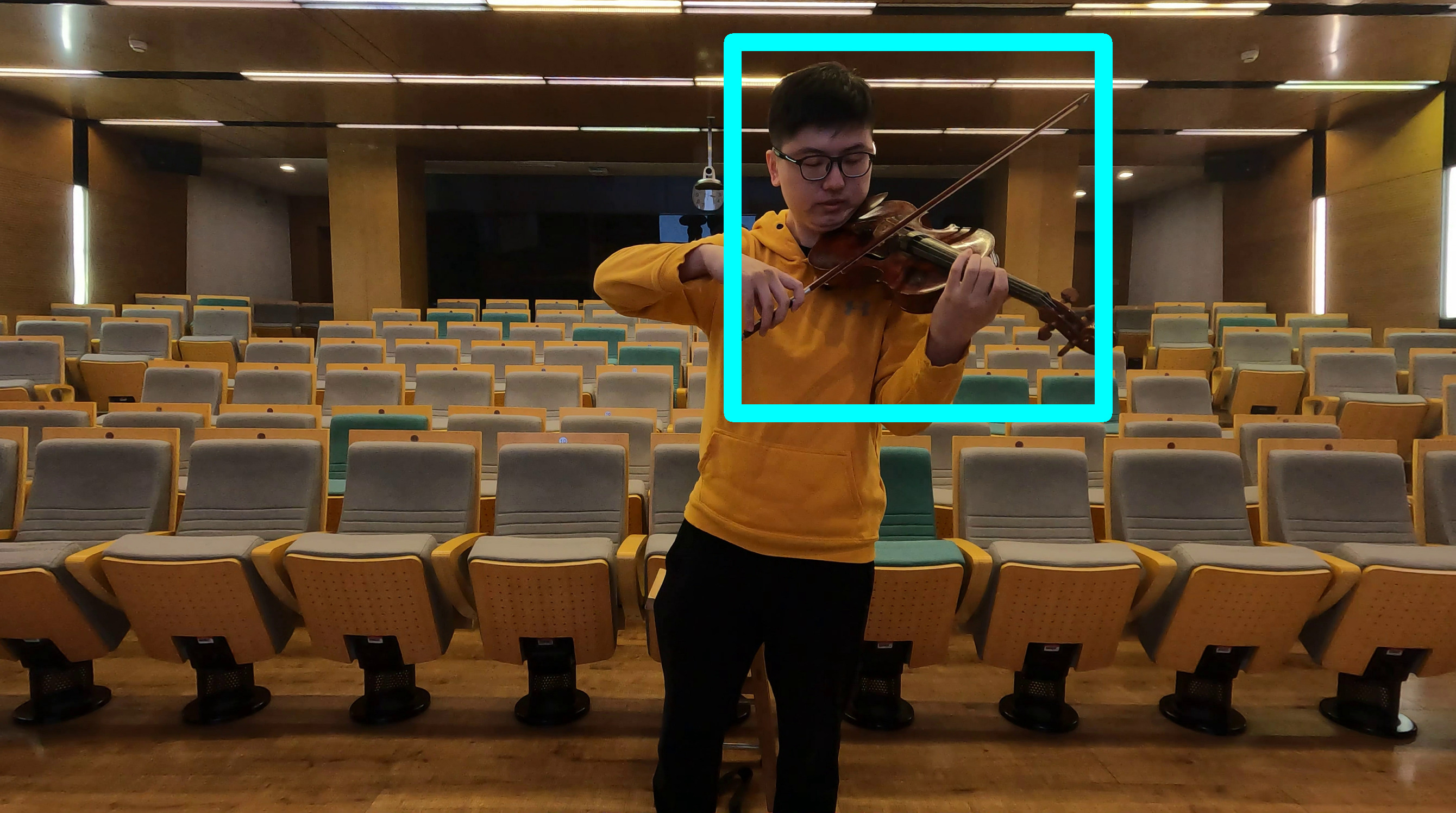} \\\vspace{1.5pt}\fcolorbox[HTML]{00FFFF}{00FFFF}{\includegraphics[width=\dimexpr\linewidth-2\fboxrule\relax]{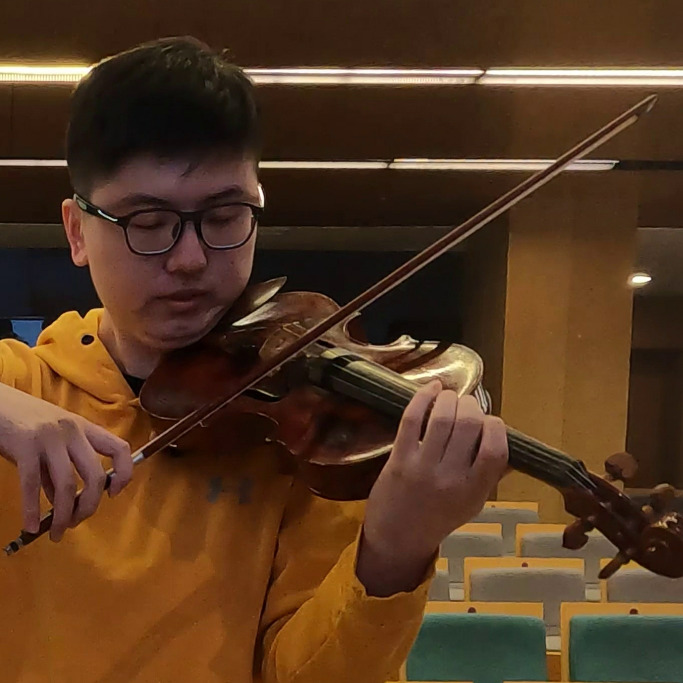}}\end{minipage}} \\

        \multicolumn{5}{c}{\vspace{3mm}} \\

        \rotatebox[origin=c]{90}{\small \textbf{Puppy}} &
        \parbox{0.155\textwidth}{\begin{minipage}{\linewidth}\centering\setlength{\fboxsep}{0pt}\setlength{\fboxrule}{1.2pt}\includegraphics[width=\linewidth]{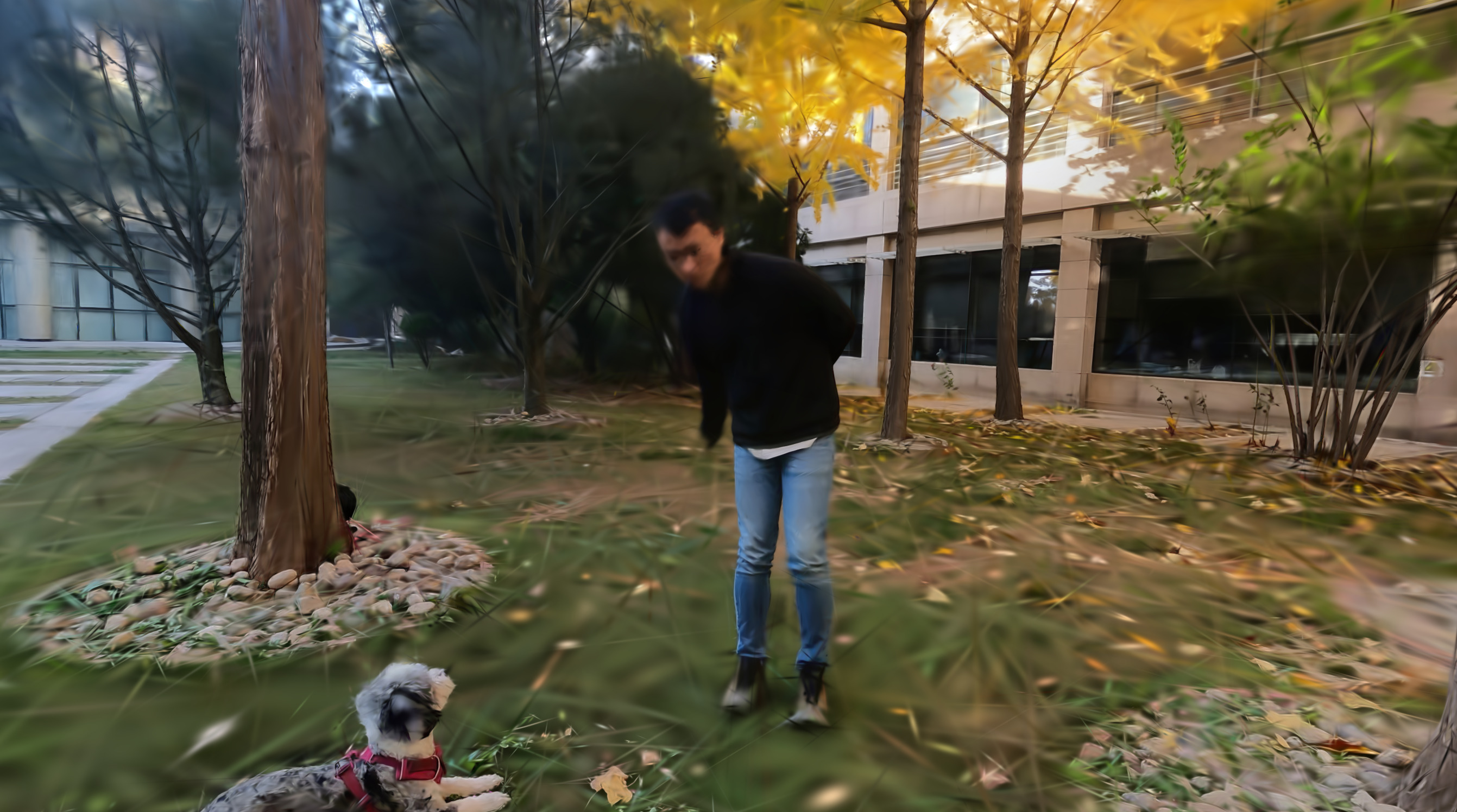} \\\vspace{1.5pt}\fcolorbox[HTML]{00FFFF}{00FFFF}{\includegraphics[width=\dimexpr0.49\linewidth-2\fboxrule\relax]{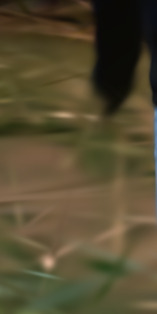}}\hfill\fcolorbox[HTML]{FF69B4}{FF69B4}{\includegraphics[width=\dimexpr0.49\linewidth-2\fboxrule\relax]{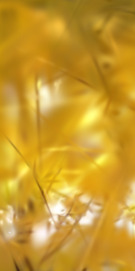}}\end{minipage}} &
        \parbox{0.155\textwidth}{\begin{minipage}{\linewidth}\centering\setlength{\fboxsep}{0pt}\setlength{\fboxrule}{1.2pt}\includegraphics[width=\linewidth]{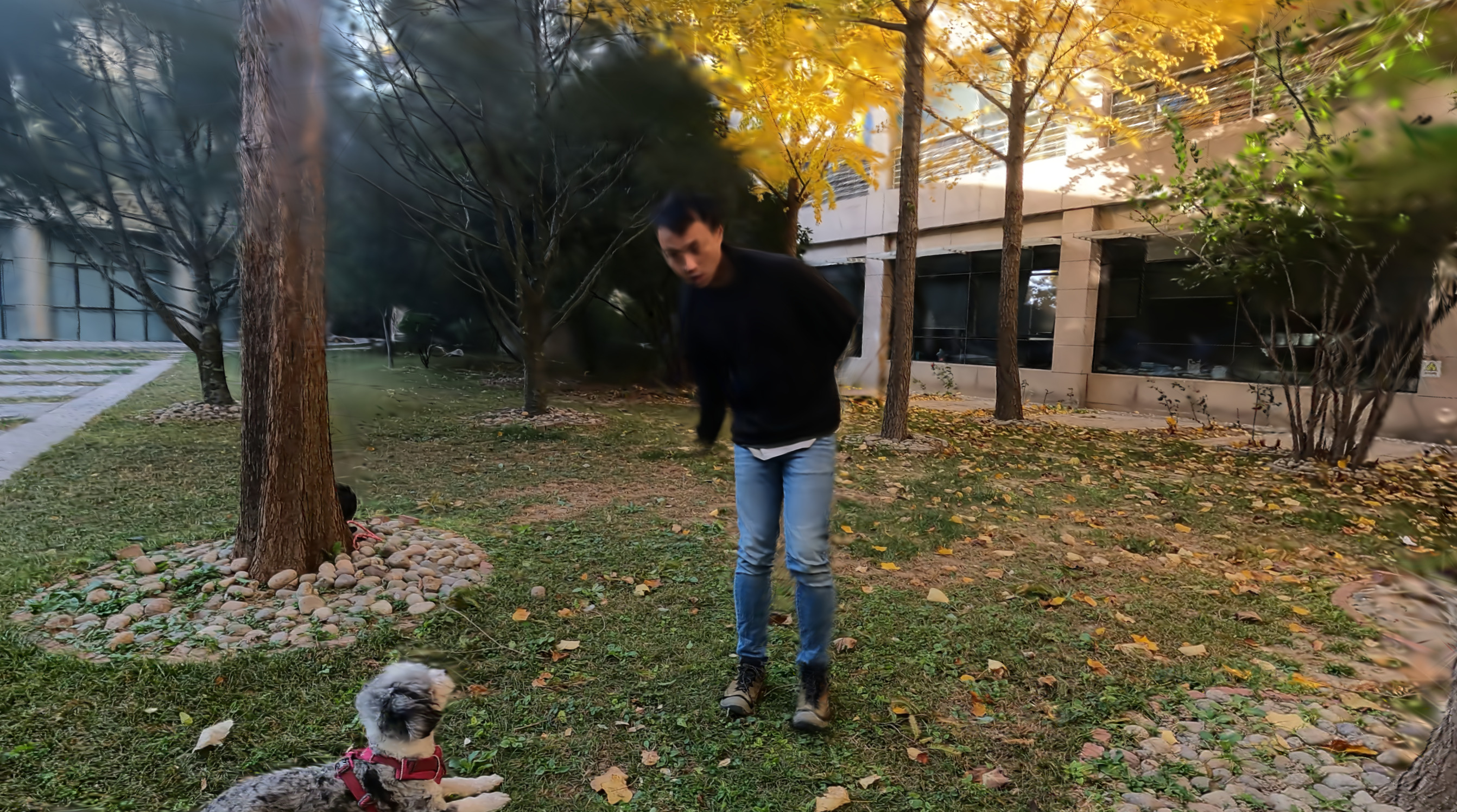} \\\vspace{1.5pt}\fcolorbox[HTML]{00FFFF}{00FFFF}{\includegraphics[width=\dimexpr0.49\linewidth-2\fboxrule\relax]{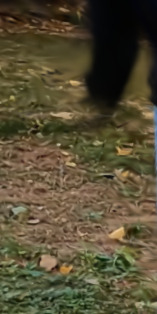}}\hfill\fcolorbox[HTML]{FF69B4}{FF69B4}{\includegraphics[width=\dimexpr0.49\linewidth-2\fboxrule\relax]{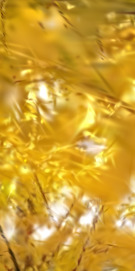}}\end{minipage}} &
        \parbox{0.155\textwidth}{\begin{minipage}{\linewidth}\centering\setlength{\fboxsep}{0pt}\setlength{\fboxrule}{1.2pt}\includegraphics[width=\linewidth]{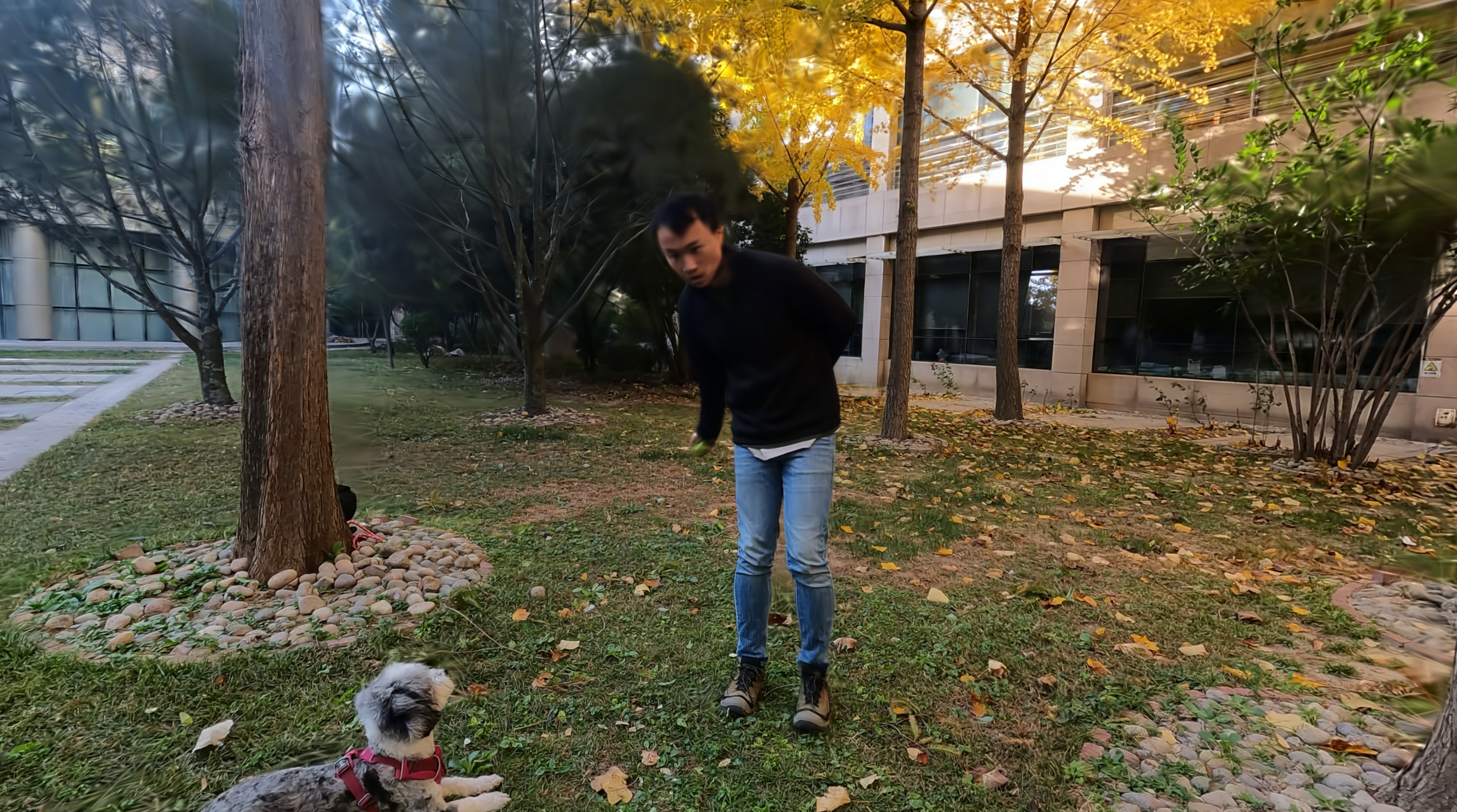} \\\vspace{1.5pt}\fcolorbox[HTML]{00FFFF}{00FFFF}{\includegraphics[width=\dimexpr0.49\linewidth-2\fboxrule\relax]{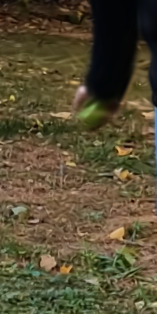}}\hfill\fcolorbox[HTML]{FF69B4}{FF69B4}{\includegraphics[width=\dimexpr0.49\linewidth-2\fboxrule\relax]{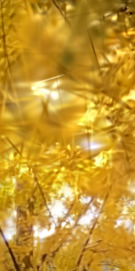}}\end{minipage}} &
        \parbox{0.155\textwidth}{\begin{minipage}{\linewidth}\centering\setlength{\fboxsep}{0pt}\setlength{\fboxrule}{1.2pt}\includegraphics[width=\linewidth]{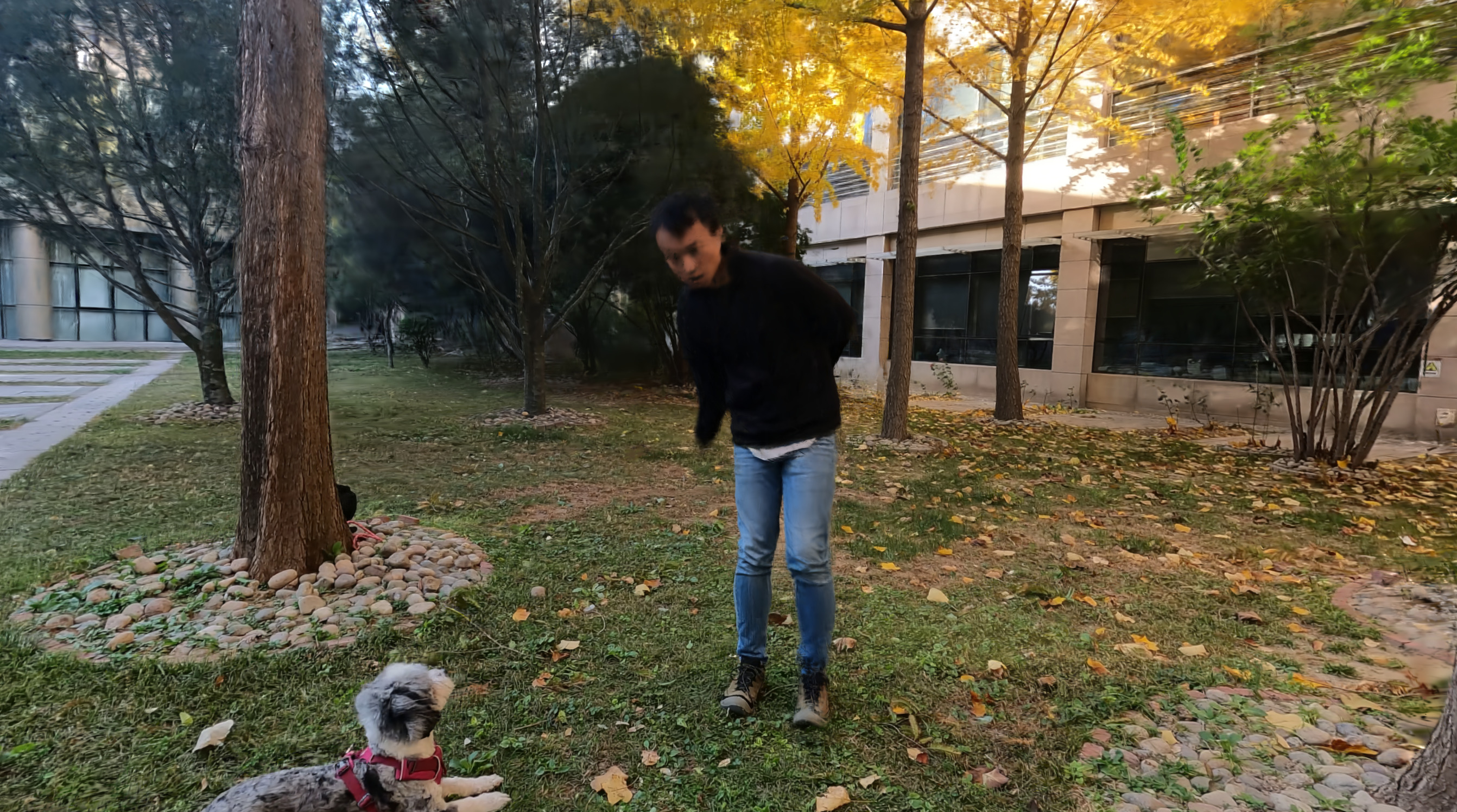} \\\vspace{1.5pt}\fcolorbox[HTML]{00FFFF}{00FFFF}{\includegraphics[width=\dimexpr0.49\linewidth-2\fboxrule\relax]{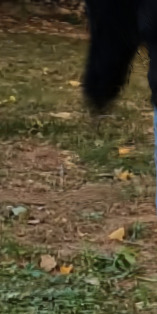}}\hfill\fcolorbox[HTML]{FF69B4}{FF69B4}{\includegraphics[width=\dimexpr0.49\linewidth-2\fboxrule\relax]{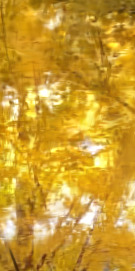}}\end{minipage}} &
        \parbox{0.155\textwidth}{\begin{minipage}{\linewidth}\centering\setlength{\fboxsep}{0pt}\setlength{\fboxrule}{1.2pt}\includegraphics[width=\linewidth]{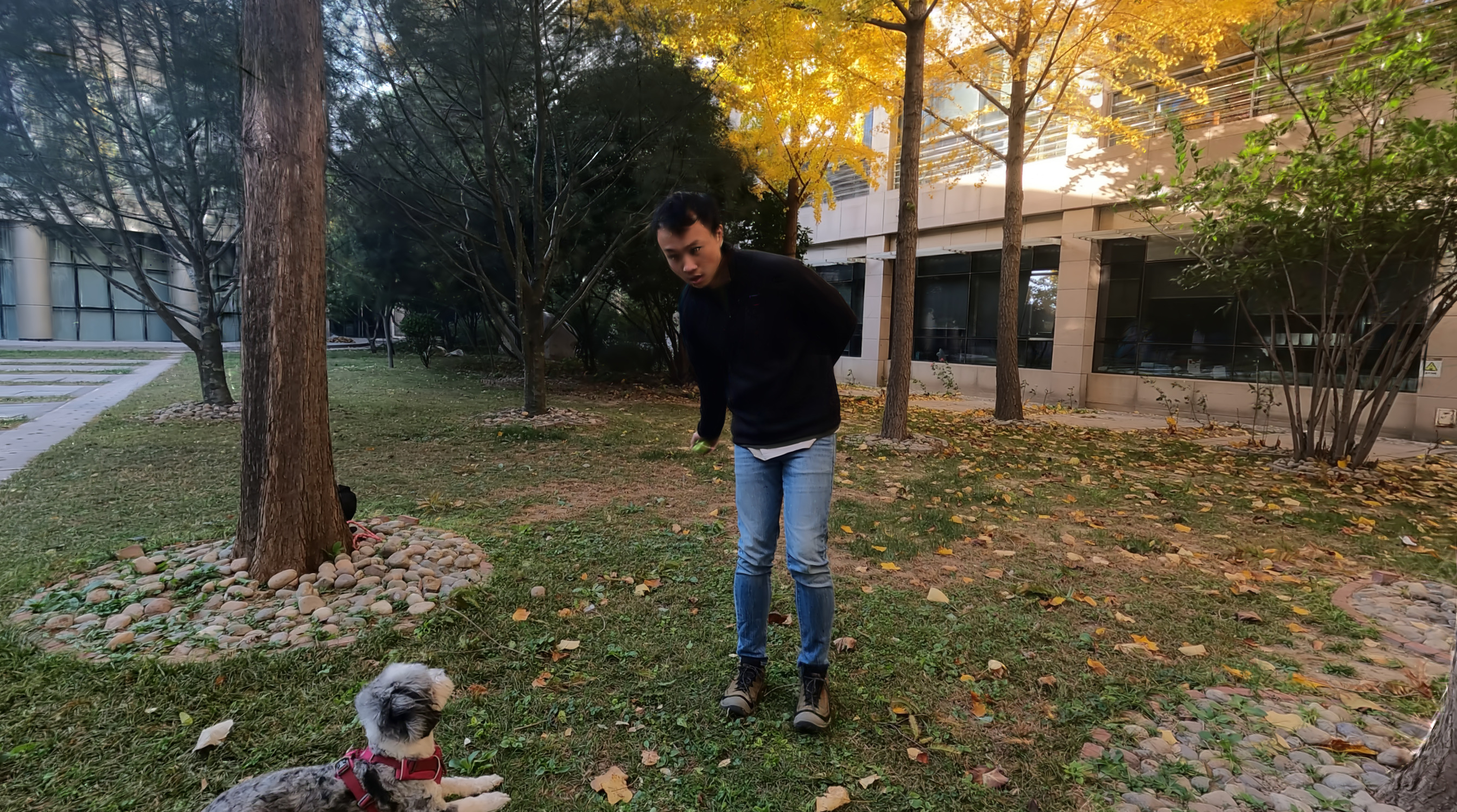} \\\vspace{1.5pt}\fcolorbox[HTML]{00FFFF}{00FFFF}{\includegraphics[width=\dimexpr0.49\linewidth-2\fboxrule\relax]{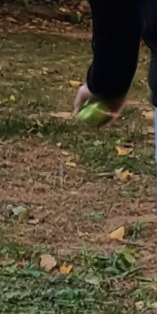}}\hfill\fcolorbox[HTML]{FF69B4}{FF69B4}{\includegraphics[width=\dimexpr0.49\linewidth-2\fboxrule\relax]{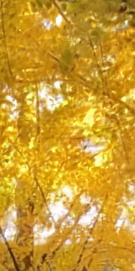}}\end{minipage}} &
        \parbox{0.155\textwidth}{\begin{minipage}{\linewidth}\centering\setlength{\fboxsep}{0pt}\setlength{\fboxrule}{1.2pt}\includegraphics[width=\linewidth]{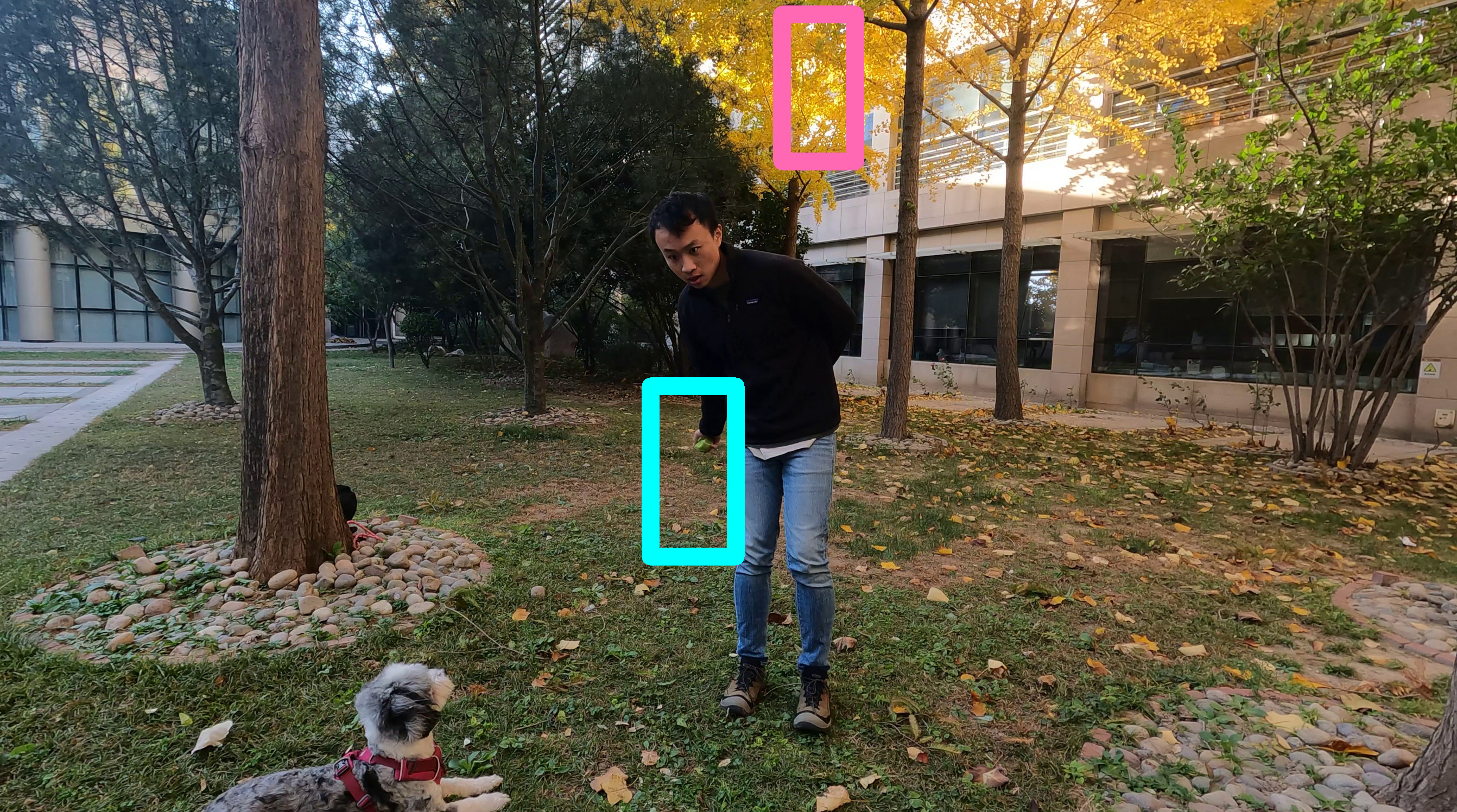} \\\vspace{1.5pt}\fcolorbox[HTML]{00FFFF}{00FFFF}{\includegraphics[width=\dimexpr0.49\linewidth-2\fboxrule\relax]{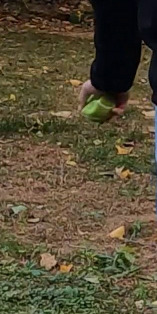}}\hfill\fcolorbox[HTML]{FF69B4}{FF69B4}{\includegraphics[width=\dimexpr0.49\linewidth-2\fboxrule\relax]{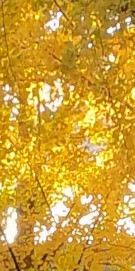}}\end{minipage}} \\
        
    \end{tabular}

    \caption{\textbf{Qualitative Comparison on \textit{ImViD} Dataset.}
    }
    \label{fig:qualitative_imvid}
\end{figure*}

Fig.~\ref{fig:qualitative_imvid} presents qualitative comparisons between our method and representative baselines on our proposed dataset.
Benefiting from effective spatio-temporal consistency supervision and camera-wise temporal alignment, our method better preserves sharp and fast-moving structures. 
Notably, our approach not only robustly models fast foreground motion but also faithfully reconstructs subtle background dynamics, such as the swaying leaves.

Additional qualitative results and visual comparisons across four other benchmark datasets (\textit{N3V}, \textit{MPEG-GSC}, \textit{MeetRoom}, and \textit{Google Immersive}) are provided in the supplementary material, which is available at \url{https://sheng-qi.github.io/IVV/}.
As shown in Supp.~Fig.~1 on \textit{N3V}, the fine details of rapidly moving objects, such as the spoon and flamethrower, are reconstructed more accurately.
In scenes with even faster motion (Supp.~Fig.~2), such as the basketball and the shaking hand in \textit{MPEG-GSC}, our method maintains clear geometry and appearance without noticeable temporal artifacts.
On \textit{MeetRoom} (Supp.~Fig.~3), our method preserves clearer boundaries in fast-moving regions, while maintaining more stable visual details in low-texture indoor conditions.
Across various scenes in Supp.~Fig.~4, particularly for flames, our method achieves clearer and more coherent reconstructions, with Gaussian motion better aligned with the underlying dynamics.

\subsubsection{Ablation Studies}

We conduct ablation studies on the \textit{ImViD} dataset to evaluate the contribution of each core component in our framework.

\paragraph{Effectiveness of Flow-Guided Sparse Initialization}
Compared with initializing sparse point clouds at every frame (as adopted in STG~\cite{li2024spacetime}), our flow-guided strategy substantially reduces the number of initial primitives.
As shown in Table~\ref{tab:ablation_init}, our method achieves higher reconstruction quality while using only about half of the final spatio-temporal Gaussian primitives.
This indicates that the proposed initialization effectively suppresses redundant geometry in repetitive or low-motion regions while enabling efficient primitive growth in dynamically informative areas.

\begin{table}[t]
\centering
\renewcommand{\arraystretch}{0.9}
\caption{Ablation on Flow-Guided Initialization. Our strategy maintains high reconstruction fidelity while significantly reducing initial primitive redundancy.}
\label{tab:ablation_init}
\footnotesize
\begin{tabular*}{\columnwidth}{@{\extracolsep{\fill}}l@{\hspace{4pt}}c@{\hspace{4pt}}c@{\hspace{4pt}}c@{\hspace{4pt}}c@{\hspace{4pt}}c}
\toprule
Setting                 & Init. Pts (Ratio) & Final Pts & PSNR $\uparrow$ & SSIM $\uparrow$ & LPIPS $\downarrow$ \\
\midrule
w/o Flow Init.          & 2.48M (100\%)     & 4.52M          & 33.13          & 0.915          & 0.070 \\
w Flow Init. (Ours)     & \textbf{0.29M} (\textbf{11.6\%}) & \textbf{2.50M} & \textbf{33.51} & \textbf{0.916} & 0.070 \\
\bottomrule
\end{tabular*}
\end{table}

\paragraph{Effectiveness of Joint Camera Temporal Calibration}
Table~\ref{tab:ablation_time} and Fig.~\ref{fig:ablation_cam00} validate the effectiveness of jointly optimizing camera-wise temporal offsets ($\Delta \gamma$).
Although hardware synchronization constrains the temporal error at the millisecond level, residual misalignment still introduces incorrect supervision on fast-moving regions, leading to visible artifacts.
By jointly optimizing temporal offsets and applying an $\ell_2$ regularization term, our method suppresses excessive temporal drifting and significantly improves reconstruction stability.

\begin{figure*}[!t]
    \centering
    \setlength{\tabcolsep}{2pt}
    \renewcommand{\arraystretch}{0.6}
    \begin{tabular}{ccccc}
        \small{w/o Flow \& Depth} & \small{w/o Depth} & \small{w/o Flow} & \small{Full Framework} & \small{Ground Truth} \\
        
        \parbox{0.19\linewidth}{\begin{minipage}{\linewidth}\centering\setlength{\fboxsep}{0pt}\setlength{\fboxrule}{1.2pt}\includegraphics[width=\linewidth]{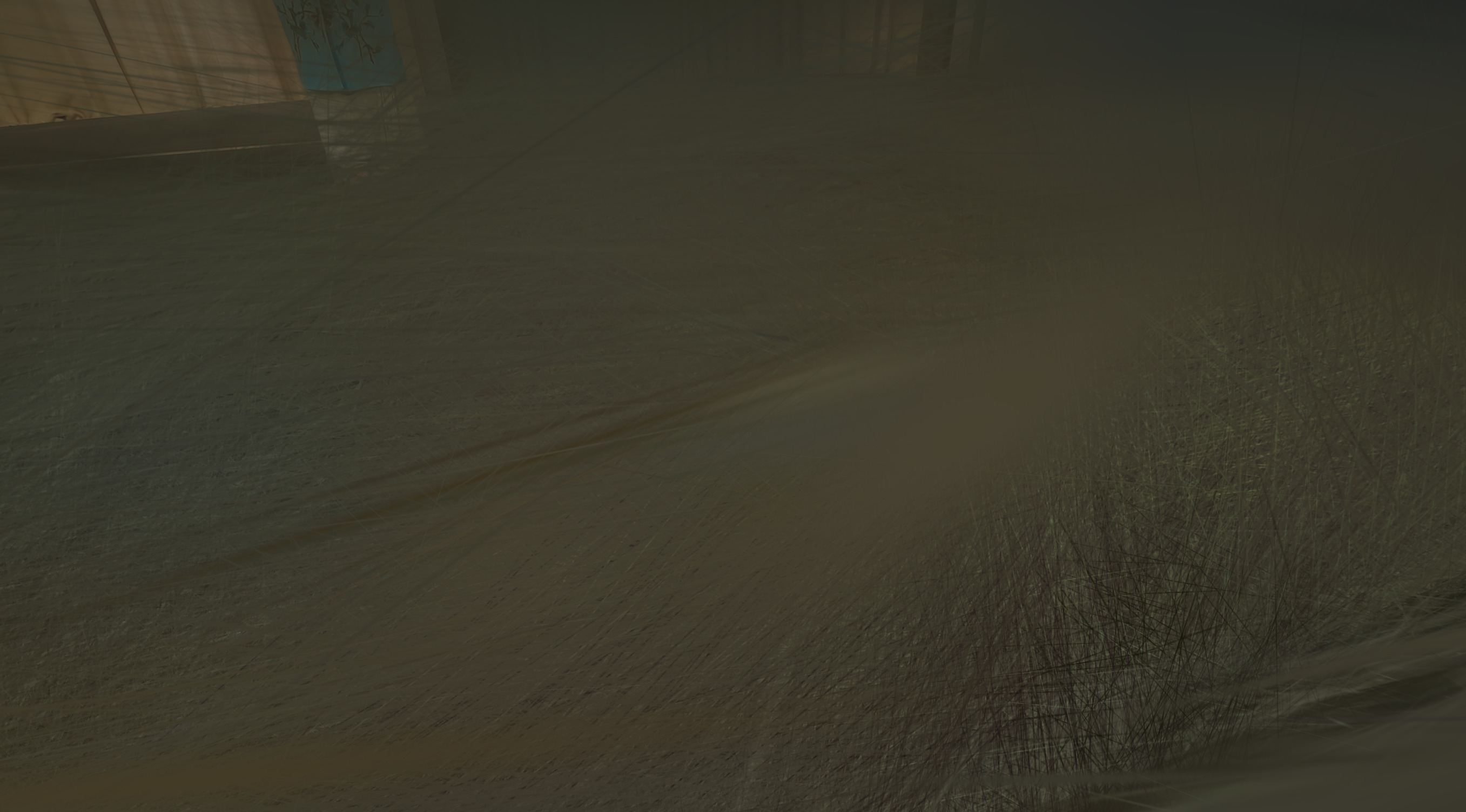} \\\vspace{1.5pt}\fcolorbox[HTML]{00FFFF}{00FFFF}{\includegraphics[width=\dimexpr0.49\linewidth-2\fboxrule\relax]{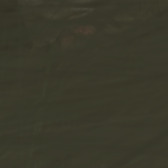}}\hfill\fcolorbox[HTML]{FF69B4}{FF69B4}{\includegraphics[width=\dimexpr0.49\linewidth-2\fboxrule\relax]{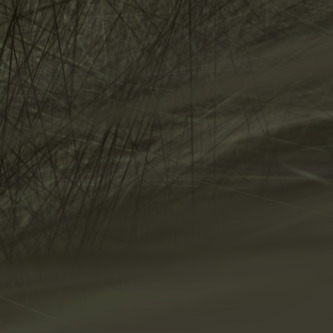}} \\\vspace{1.5pt}\fcolorbox[HTML]{FDB933}{FDB933}{\includegraphics[width=\dimexpr\linewidth-2\fboxrule\relax]{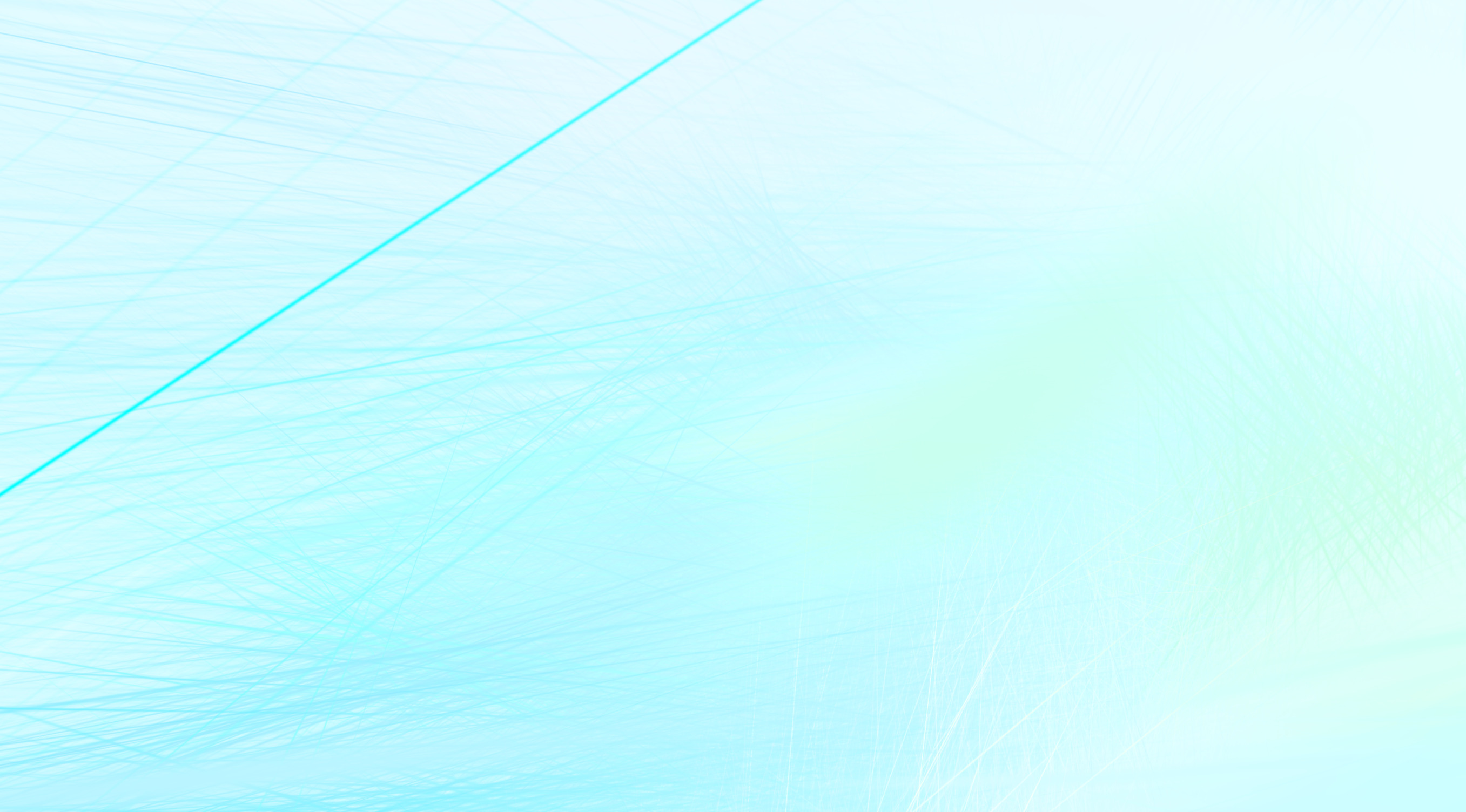}}\end{minipage}} &
        \parbox{0.19\linewidth}{\begin{minipage}{\linewidth}\centering\setlength{\fboxsep}{0pt}\setlength{\fboxrule}{1.2pt}\includegraphics[width=\linewidth]{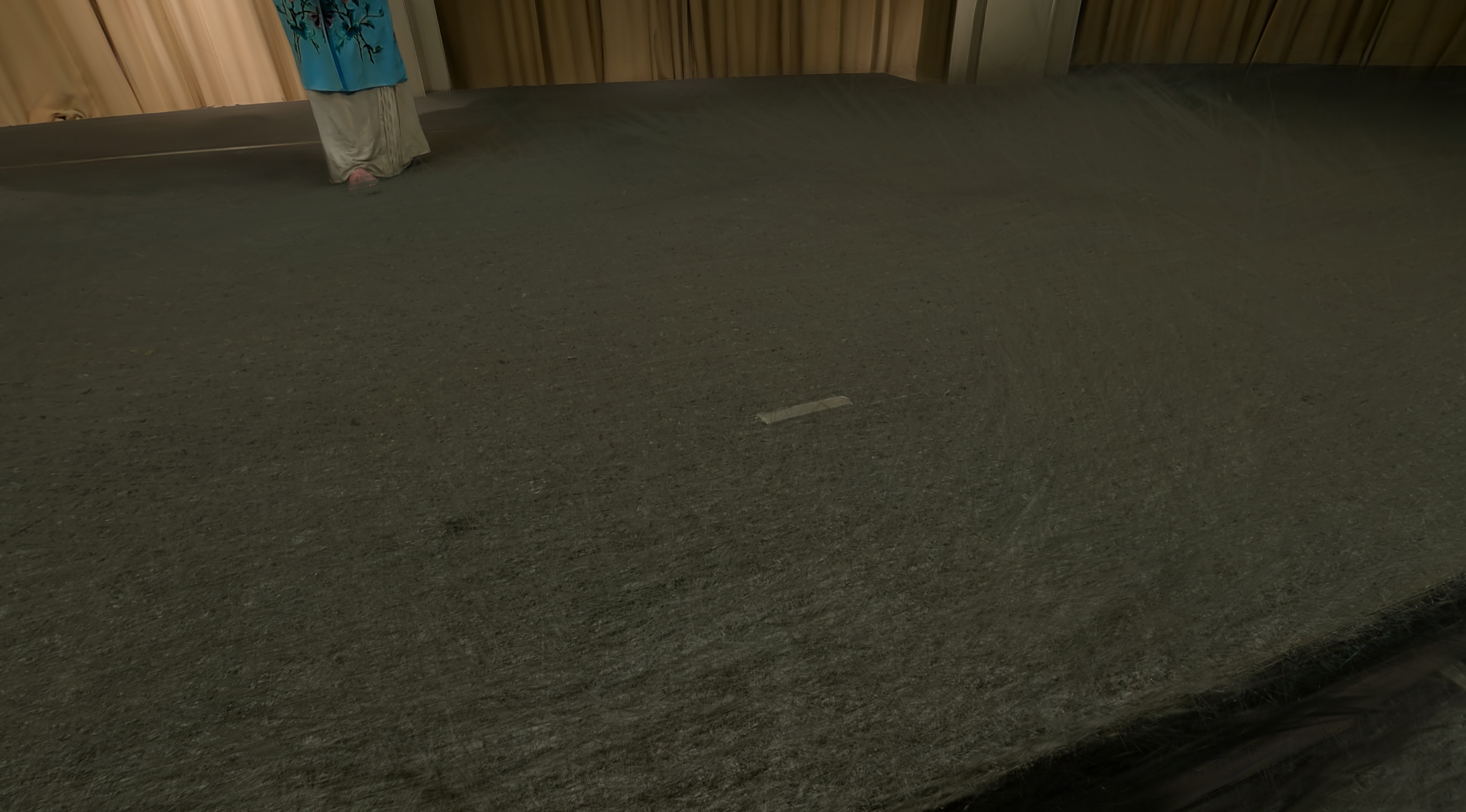} \\\vspace{1.5pt}\fcolorbox[HTML]{00FFFF}{00FFFF}{\includegraphics[width=\dimexpr0.49\linewidth-2\fboxrule\relax]{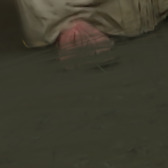}}\hfill\fcolorbox[HTML]{FF69B4}{FF69B4}{\includegraphics[width=\dimexpr0.49\linewidth-2\fboxrule\relax]{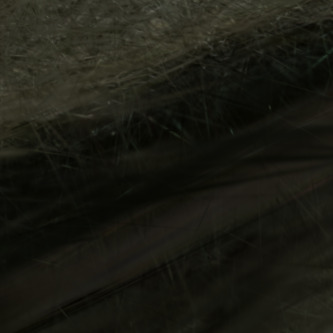}} \\\vspace{1.5pt}\fcolorbox[HTML]{FDB933}{FDB933}{\includegraphics[width=\dimexpr\linewidth-2\fboxrule\relax]{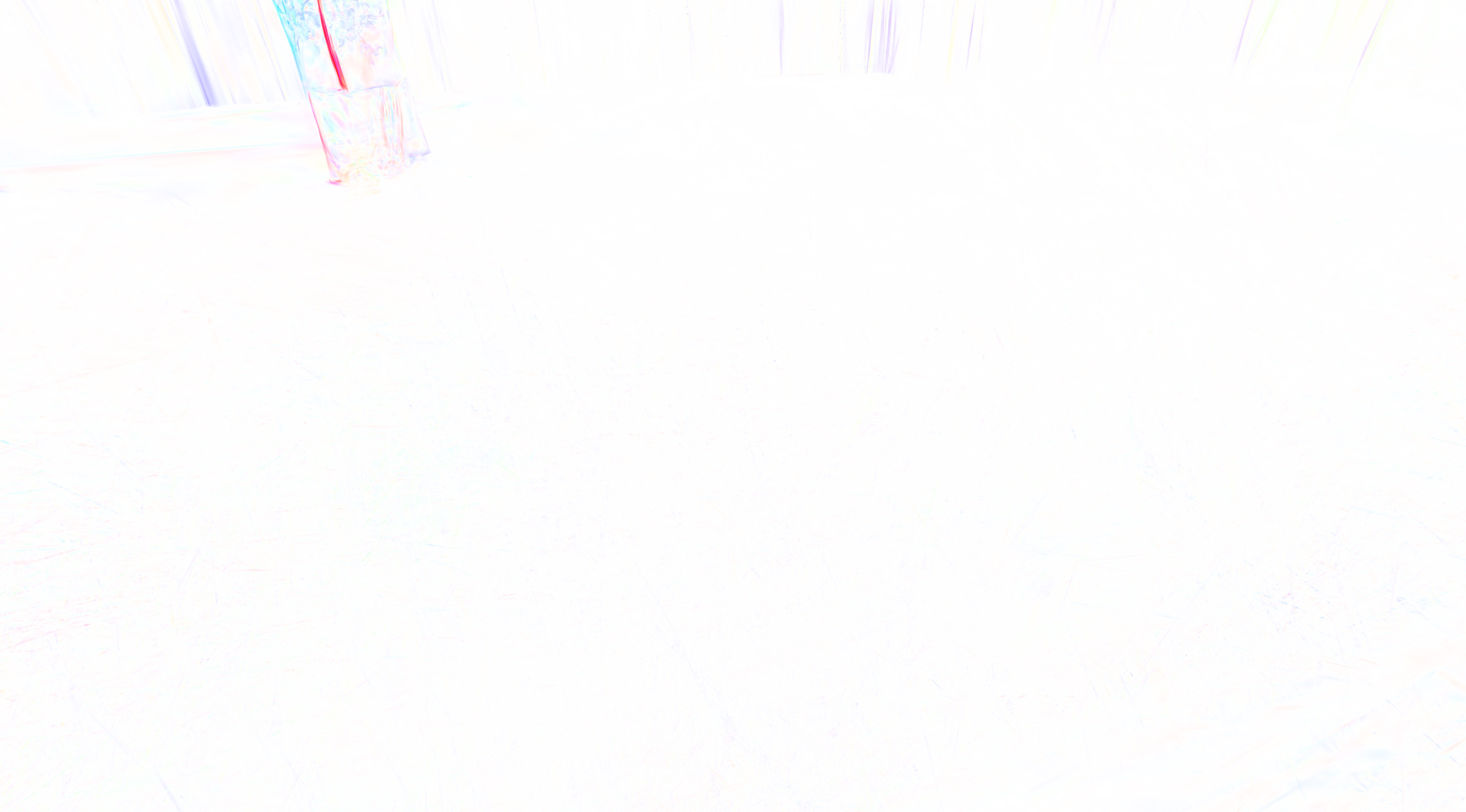}}\end{minipage}} &
        \parbox{0.19\linewidth}{\begin{minipage}{\linewidth}\centering\setlength{\fboxsep}{0pt}\setlength{\fboxrule}{1.2pt}\includegraphics[width=\linewidth]{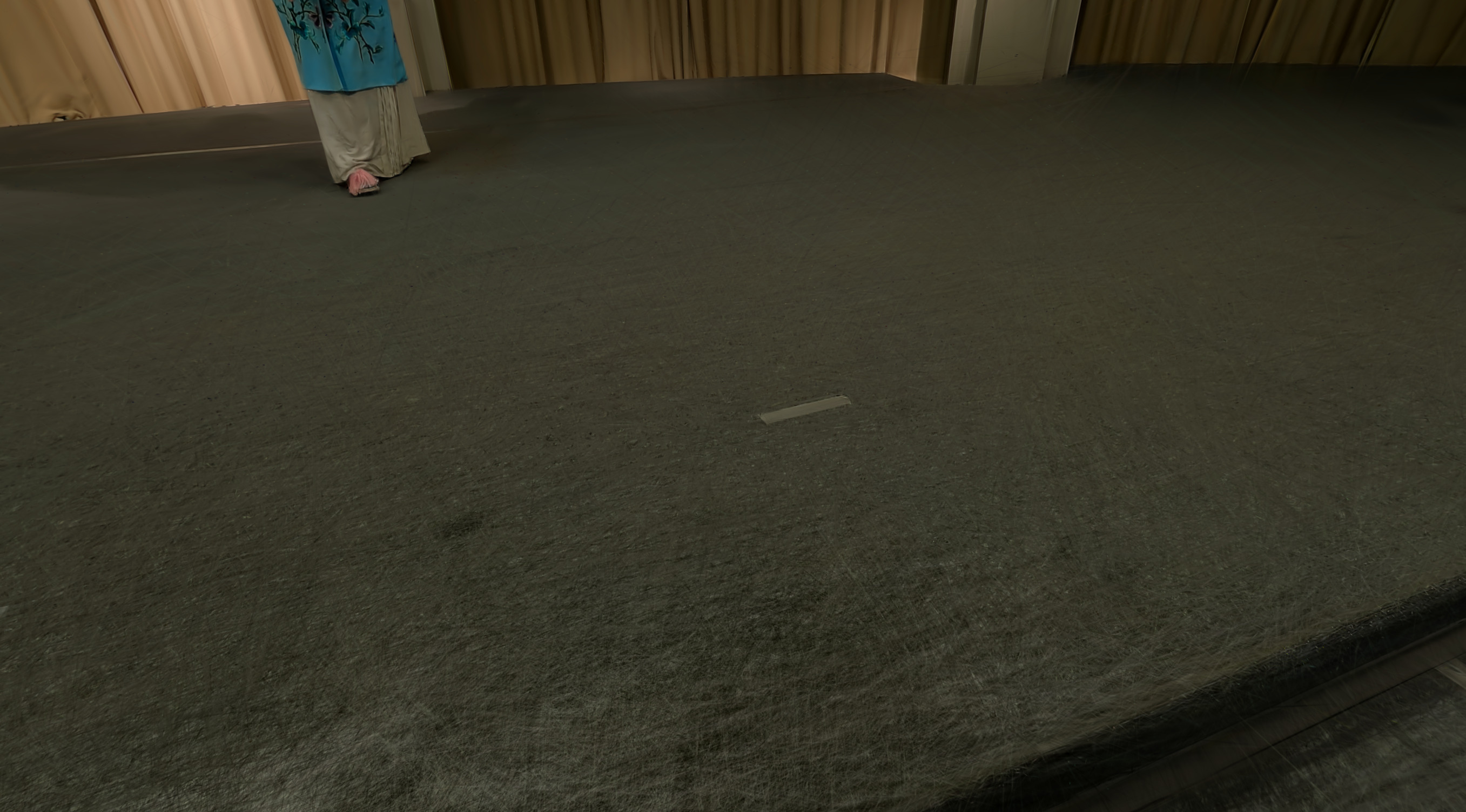} \\\vspace{1.5pt}\fcolorbox[HTML]{00FFFF}{00FFFF}{\includegraphics[width=\dimexpr0.49\linewidth-2\fboxrule\relax]{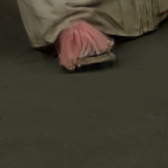}}\hfill\fcolorbox[HTML]{FF69B4}{FF69B4}{\includegraphics[width=\dimexpr0.49\linewidth-2\fboxrule\relax]{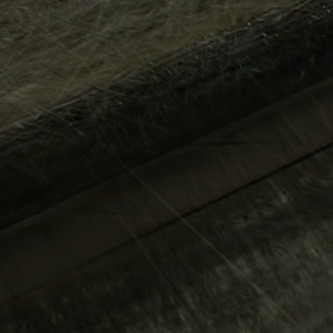}} \\\vspace{1.5pt}\fcolorbox[HTML]{FDB933}{FDB933}{\includegraphics[width=\dimexpr\linewidth-2\fboxrule\relax]{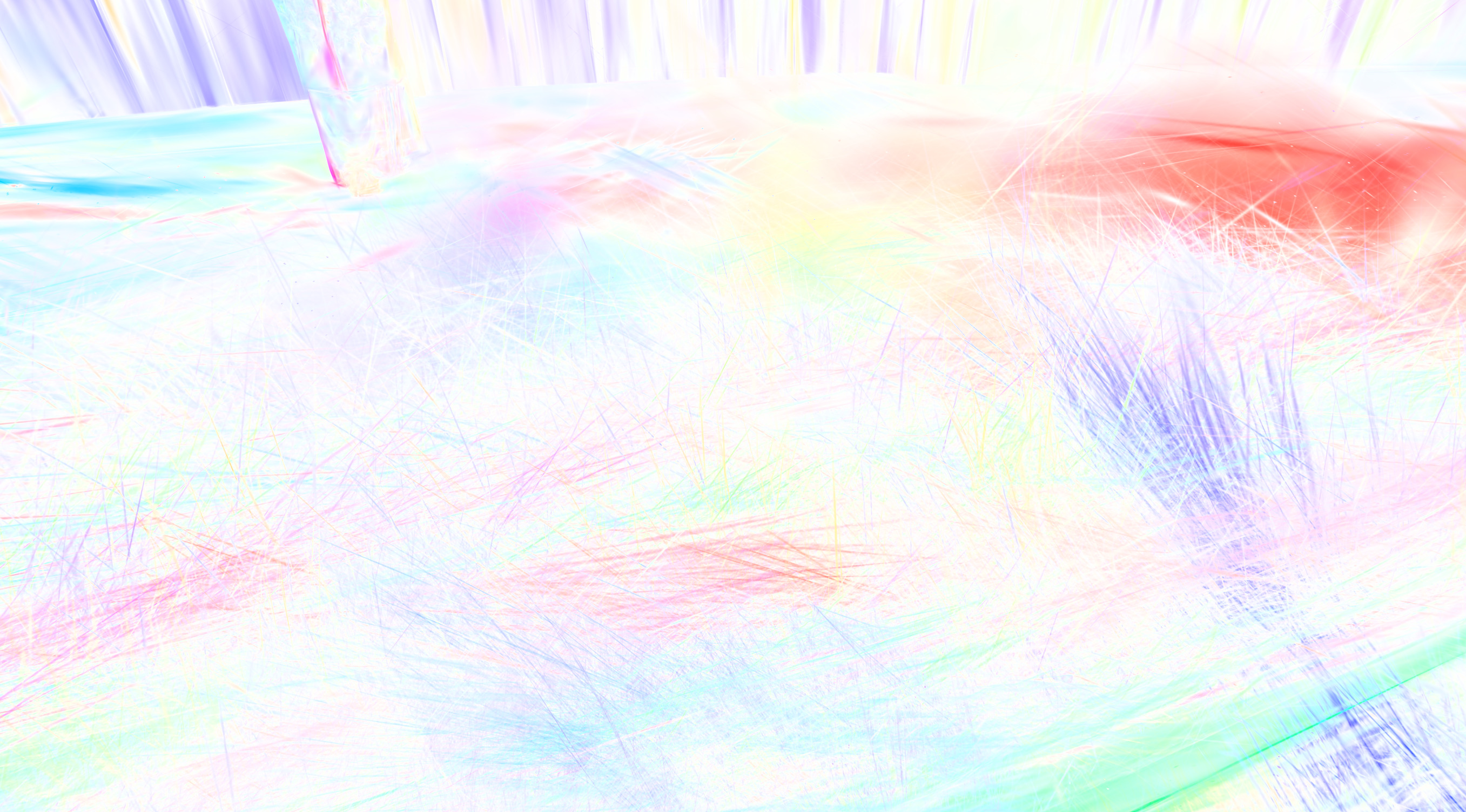}}\end{minipage}} &
        \parbox{0.19\linewidth}{\begin{minipage}{\linewidth}\centering\setlength{\fboxsep}{0pt}\setlength{\fboxrule}{1.2pt}\includegraphics[width=\linewidth]{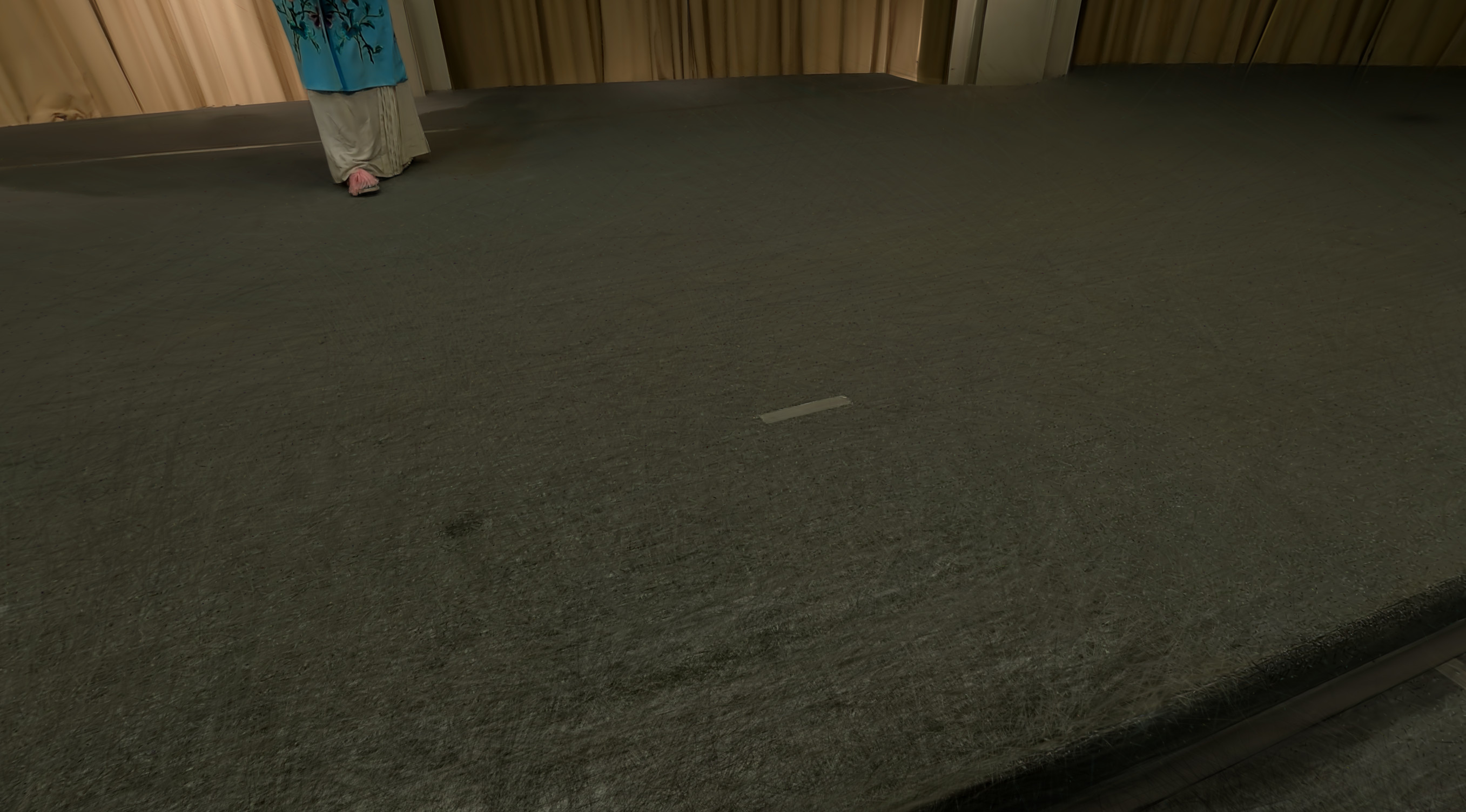} \\\vspace{1.5pt}\fcolorbox[HTML]{00FFFF}{00FFFF}{\includegraphics[width=\dimexpr0.49\linewidth-2\fboxrule\relax]{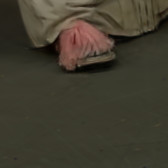}}\hfill\fcolorbox[HTML]{FF69B4}{FF69B4}{\includegraphics[width=\dimexpr0.49\linewidth-2\fboxrule\relax]{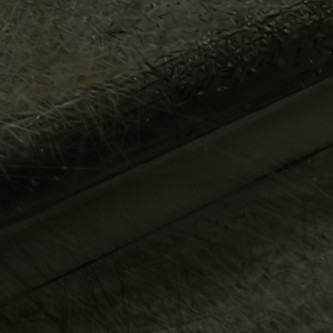}} \\\vspace{1.5pt}\fcolorbox[HTML]{FDB933}{FDB933}{\includegraphics[width=\dimexpr\linewidth-2\fboxrule\relax]{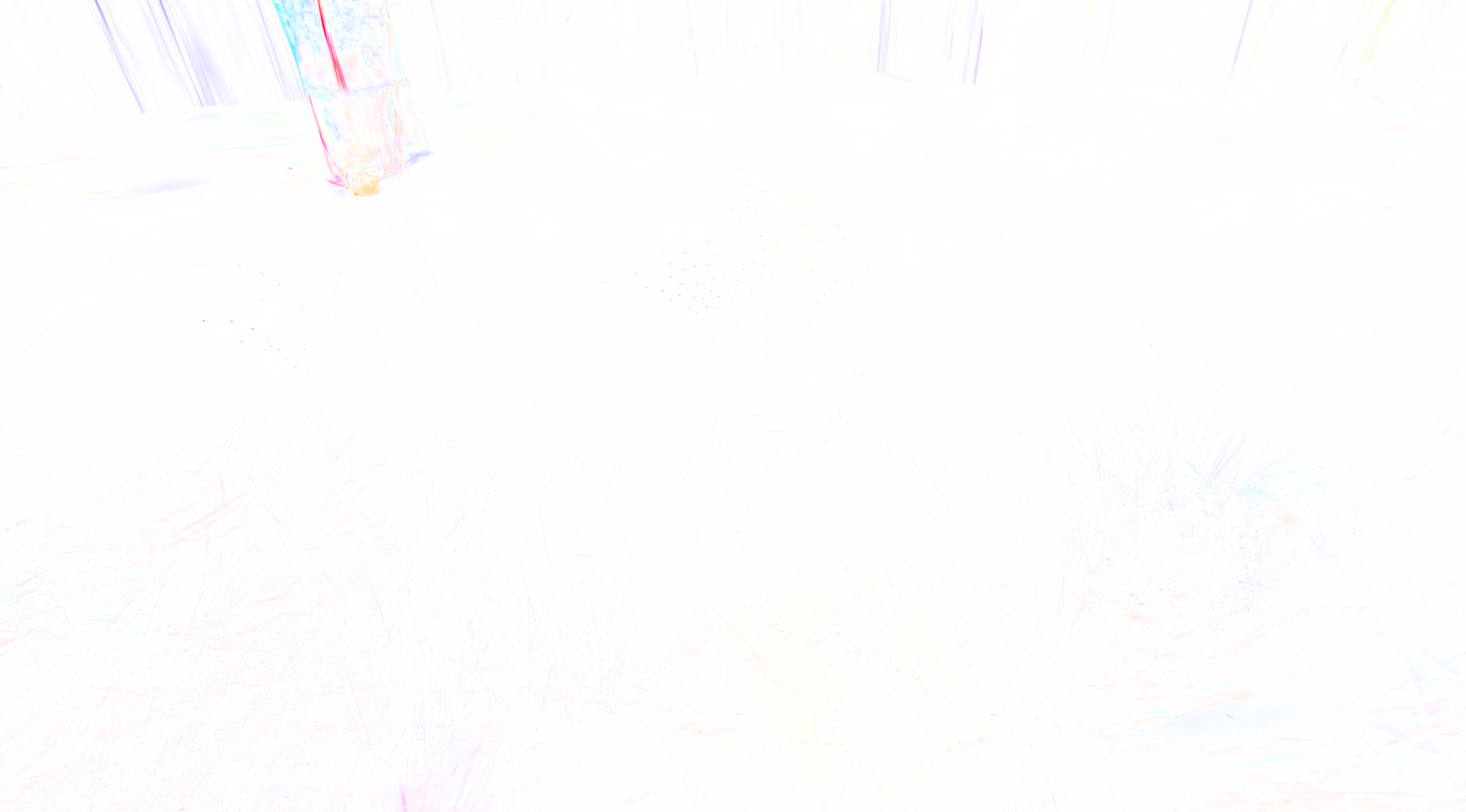}}\end{minipage}} &
        \parbox{0.19\linewidth}{\begin{minipage}{\linewidth}\centering\setlength{\fboxsep}{0pt}\setlength{\fboxrule}{1.2pt}\includegraphics[width=\linewidth]{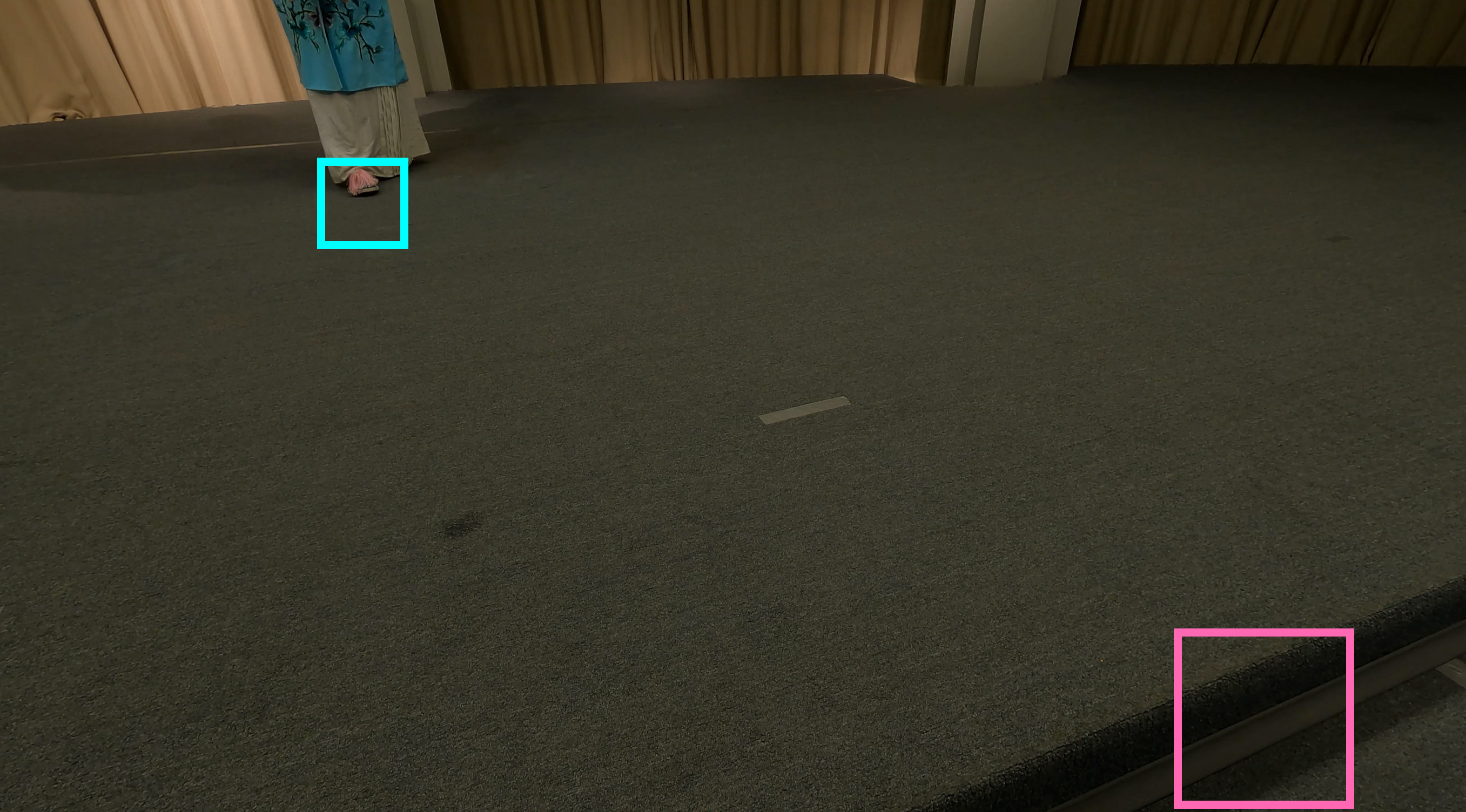} \\\vspace{1.5pt}\fcolorbox[HTML]{00FFFF}{00FFFF}{\includegraphics[width=\dimexpr0.49\linewidth-2\fboxrule\relax]{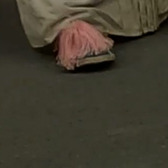}}\hfill\fcolorbox[HTML]{FF69B4}{FF69B4}{\includegraphics[width=\dimexpr0.49\linewidth-2\fboxrule\relax]{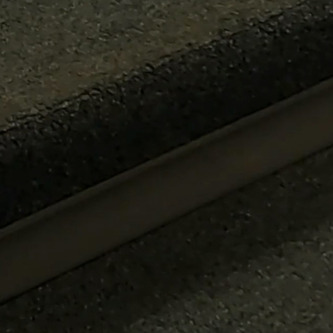}} \\\vspace{1.5pt}\fcolorbox[HTML]{FDB933}{FDB933}{\includegraphics[width=\dimexpr\linewidth-2\fboxrule\relax]{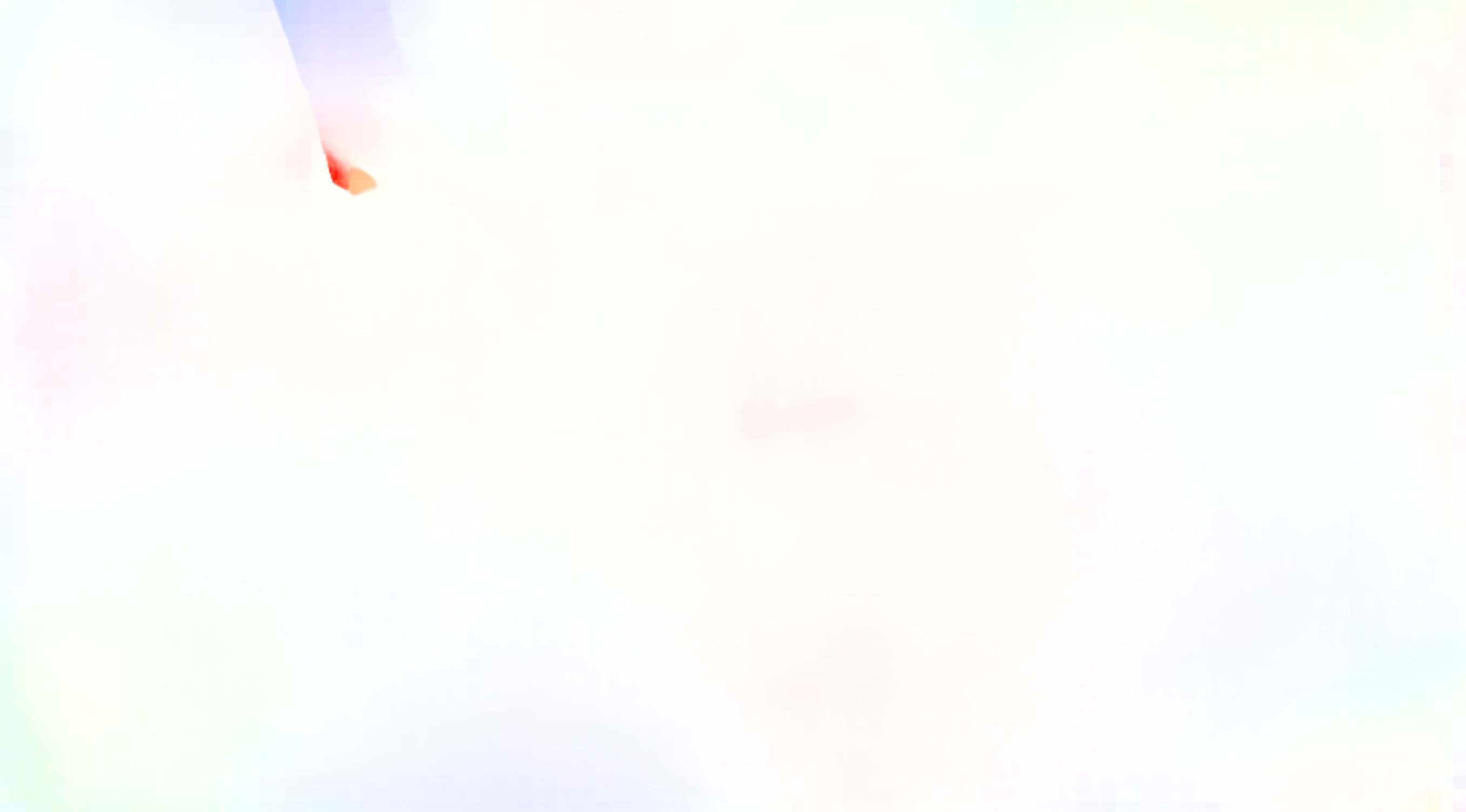}}\end{minipage}} \\
    \end{tabular}
    \caption{\textbf{Ablation on Spatio-temporal Supervision.} Qualitative results on Cam~10 show that removing depth and flow constraints leads to severe artifacts. Rendered flow maps reveal that lacking flow supervision induces chaotic background motion, directly causing severe temporal flickering.}
    \label{fig:ablation_supervision_cam10}
\end{figure*}

\begin{figure}
    \centering
    \setlength{\tabcolsep}{2pt}
    \renewcommand{\arraystretch}{0.6}
    \begin{tabular}{ccc}
       \small{w/o $\Delta \gamma$ Opt.} & \small{Full Framework} & \small{Ground Truth} \\
        \parbox{0.31\linewidth}{\begin{minipage}{\linewidth}\centering\setlength{\fboxsep}{0pt}\setlength{\fboxrule}{1.2pt}\includegraphics[width=\linewidth]{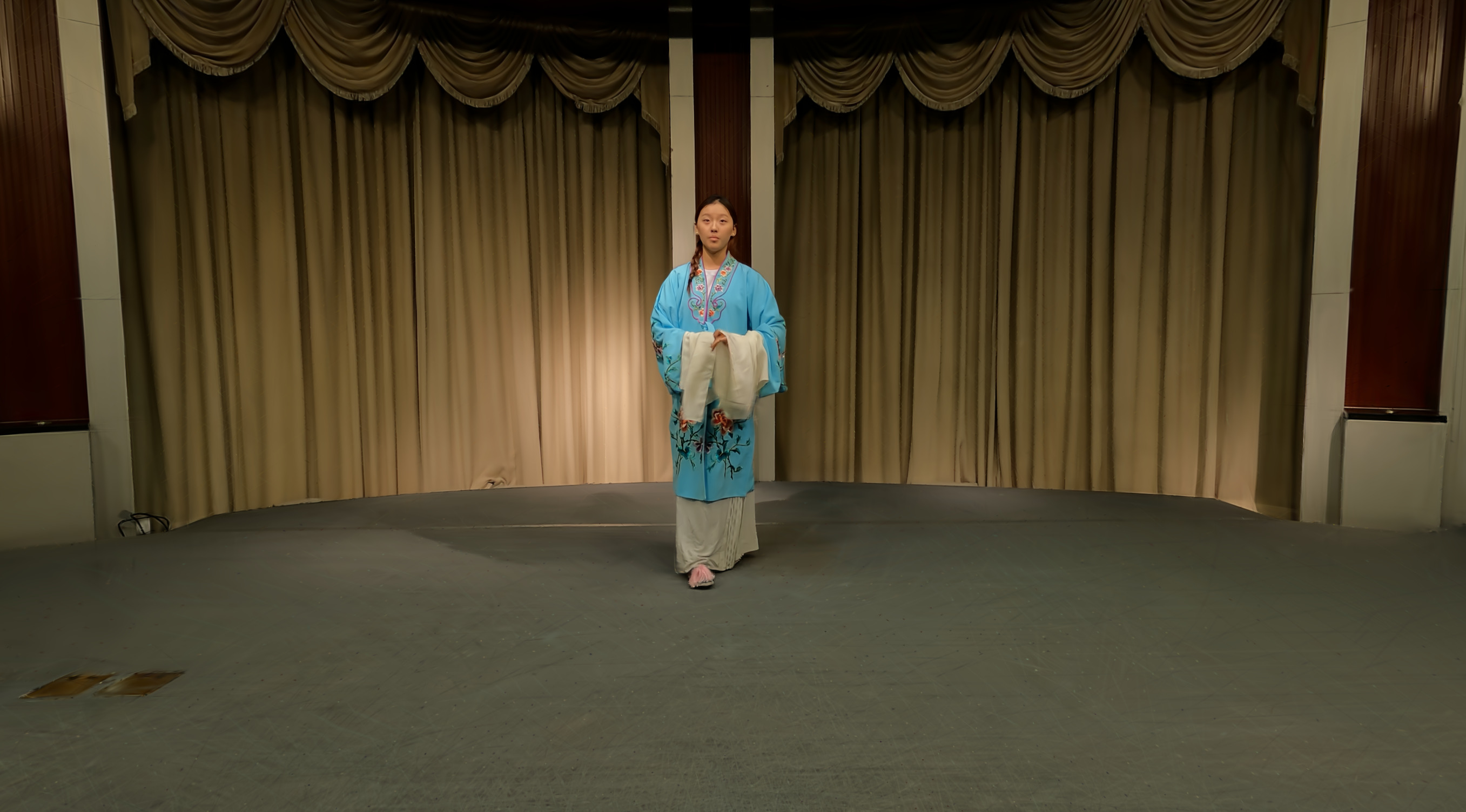} \\\vspace{1.5pt}\fcolorbox[HTML]{00FFFF}{00FFFF}{\includegraphics[width=\dimexpr\linewidth-2\fboxrule\relax]{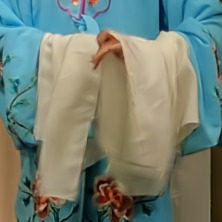}}\end{minipage}} &
        \parbox{0.31\linewidth}{\begin{minipage}{\linewidth}\centering\setlength{\fboxsep}{0pt}\setlength{\fboxrule}{1.2pt}\includegraphics[width=\linewidth]{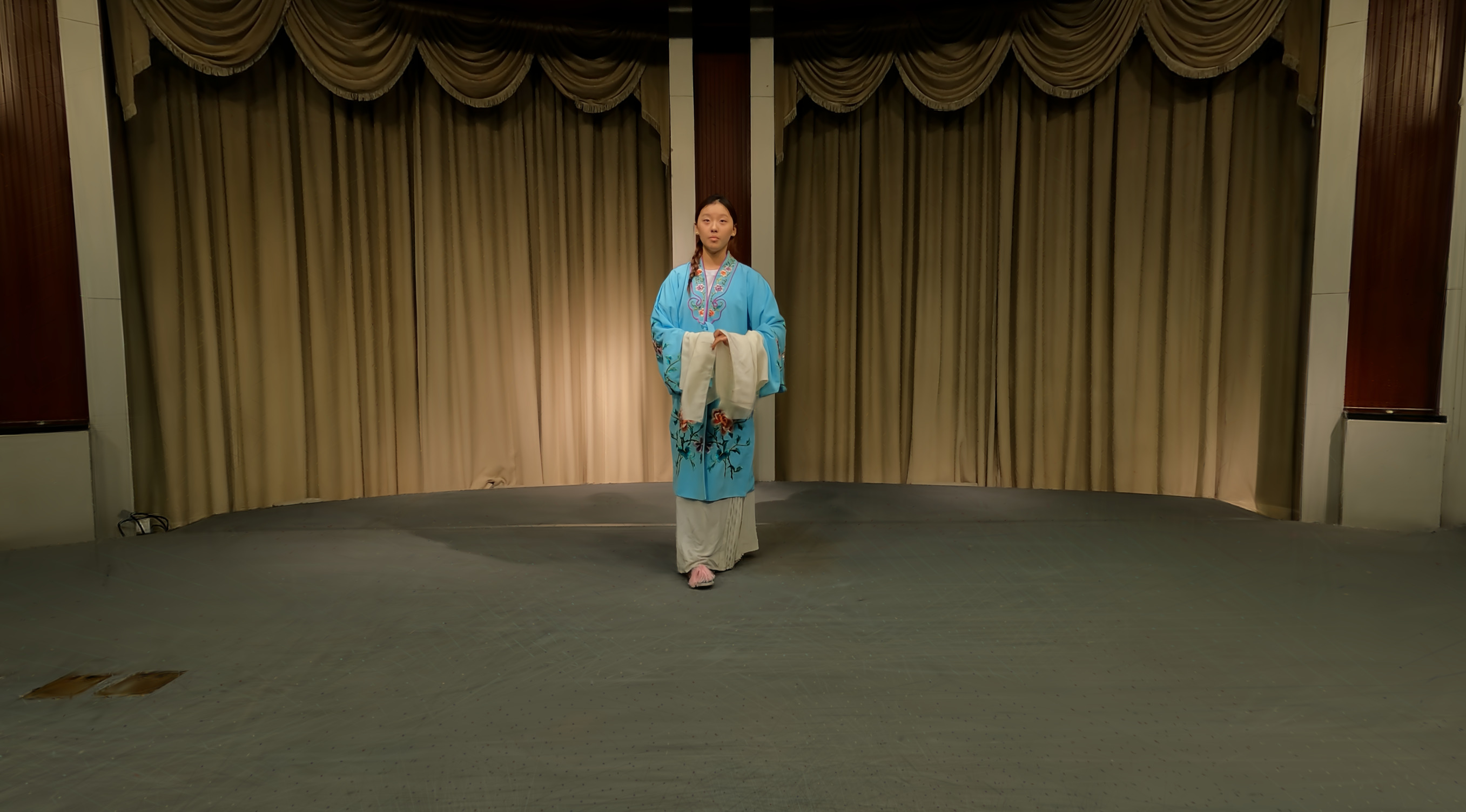} \\\vspace{1.5pt}\fcolorbox[HTML]{00FFFF}{00FFFF}{\includegraphics[width=\dimexpr\linewidth-2\fboxrule\relax]{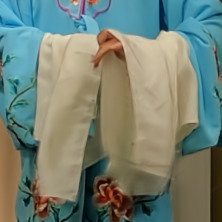}}\end{minipage}} &
        \parbox{0.31\linewidth}{\begin{minipage}{\linewidth}\centering\setlength{\fboxsep}{0pt}\setlength{\fboxrule}{1.2pt}\includegraphics[width=\linewidth]{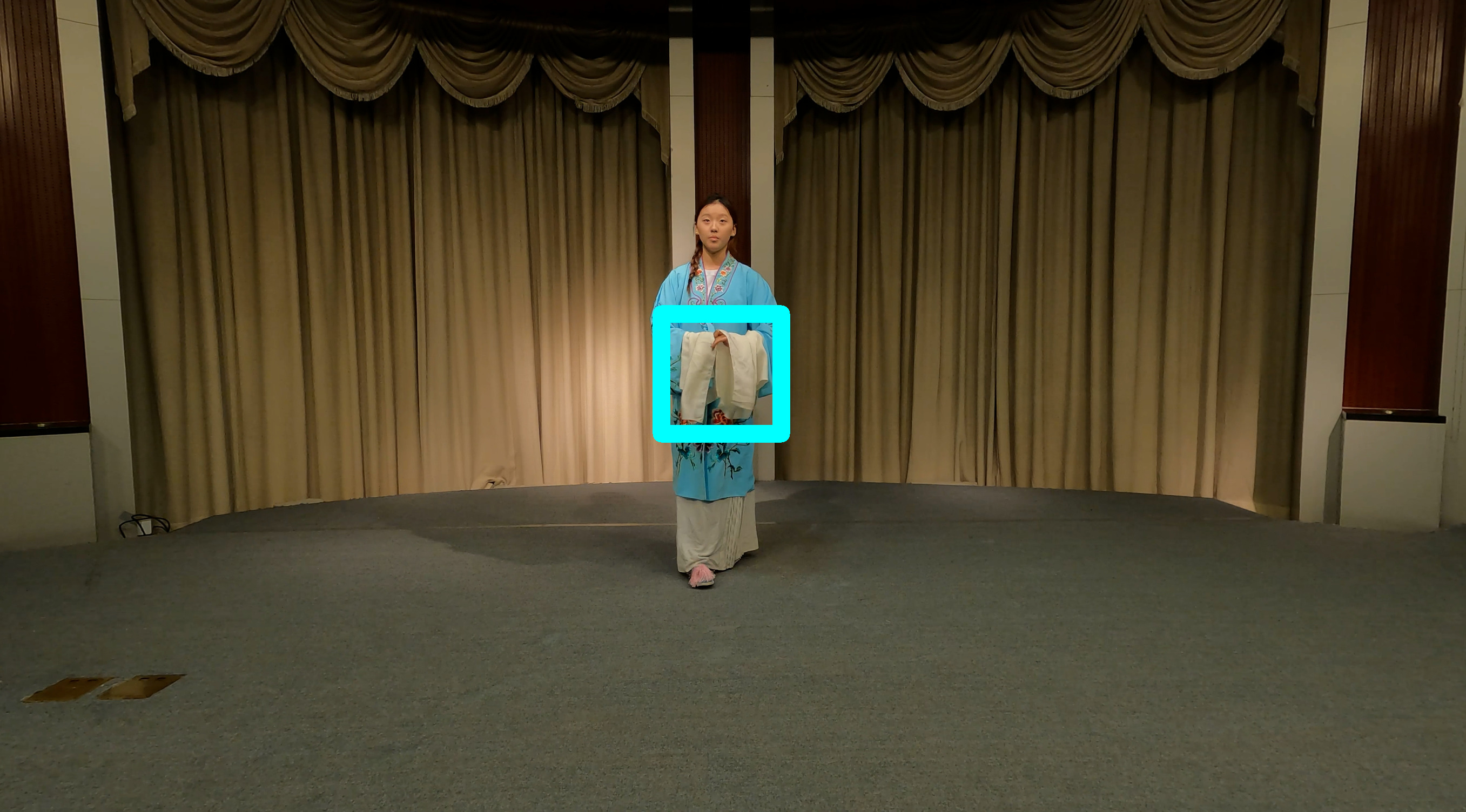} \\\vspace{1.5pt}\fcolorbox[HTML]{00FFFF}{00FFFF}{\includegraphics[width=\dimexpr\linewidth-2\fboxrule\relax]{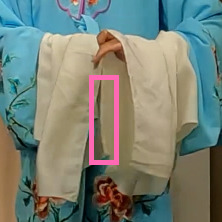}}\end{minipage}} \\
    \end{tabular}
    \caption{\textbf{Ablation on Joint Camera Temporal Calibration.} Qualitative results on the test view (Cam~00) show that jointly optimizing camera temporal offsets effectively improves reconstruction quality in fast-moving regions, yielding a clearer texture and sharper motion boundaries.}
    \label{fig:ablation_cam00}
\end{figure}

\paragraph{Effectiveness of Spatio-temporal Supervision}
We evaluate the contribution of geometric (depth) and motion (flow) constraints to reconstruction quality.
Since the central view is inherently observed by more cameras and thus benefits from stronger multi-view supervision, using it as the test view cannot fully reveal the impact of these constraints, especially in scenes with moderate motion.
Therefore, we adopt a peripheral view (Cam~10) as the test view for evaluation.

As shown in Table~\ref{tab:ablation_supervision}, removing the depth and flow constraints leads to a substantial performance degradation.
In these regions with limited multi-view coverage, the optimization lacks sufficient visual cues and tends to overfit, resulting in disorganized ``floater'' artifacts that severely degrade image quality.
As illustrated in Fig.~\ref{fig:ablation_supervision_cam10}, incorporating spatio-temporal supervision effectively regularizes both geometry and motion in such challenging areas.
These results demonstrate that the proposed constraints are crucial for achieving high-quality reconstruction across the full 6-DoF interaction space.

\begin{table}[t]
\centering
\caption{Ablation on Joint Camera Temporal Calibration. Jointly optimizing camera temporal offsets ($\Delta \gamma$) improves reconstruction quality. Metrics are reported exclusively for the dynamic foreground involving the opera performers.}
\label{tab:ablation_time}
\footnotesize
\begin{tabular*}{\columnwidth}{@{\extracolsep{\fill}}lcccc}
\toprule
Setting & $\lambda_{reg}$ & PSNR $\uparrow$ & SSIM $\uparrow$ & LPIPS $\downarrow$ \\
\midrule
w/o $\Delta \gamma$ Opt.       & -         & 30.59          & 0.910          & 0.035 \\
w/ $\Delta \gamma$ Opt.       & 0         & 29.93          & 0.903          & 0.033 \\
w/ $\Delta \gamma$ Opt.       & $10^{-2}$ & 30.50          & 0.911          & 0.031 \\
w/ $\Delta \gamma$ Opt. (Ours) & $10^{-4}$ & \textbf{30.94} & \textbf{0.917} & \textbf{0.028} \\
\bottomrule
\end{tabular*}
\end{table}

\begin{table}[t]
\centering
\caption{Ablation on Spatio-temporal Supervision. Metrics are evaluated on Cam~10 to reveal the contribution of geometric and motion constraints.}
\label{tab:ablation_supervision}
\footnotesize
\begin{tabular*}{\columnwidth}{@{\extracolsep{\fill}}lcccc}
\toprule
\noalign{\vskip -1pt}
\multicolumn{2}{c}{Loss Component} & \multicolumn{3}{c}{Peripheral View Metrics} \\
\noalign{\vskip -1pt}
\cmidrule(r){1-2} \cmidrule(l){3-5}
\noalign{\vskip -1pt}
$\mathcal{L}_{flow}$ & $\mathcal{L}_{depth}$ & PSNR $\uparrow$ & SSIM $\uparrow$ & LPIPS $\downarrow$ \\
\noalign{\vskip -1pt}
\midrule
-            & -            & 24.43          & 0.729          & 0.287          \\
\checkmark   & -            & 30.03          & 0.748          & 0.153          \\
-            & \checkmark   & 30.45          & 0.760          & 0.134          \\
\checkmark   & \checkmark   & \textbf{30.98} & \textbf{0.772} & \textbf{0.124} \\
\bottomrule
\end{tabular*}
\end{table}

\subsection{Effectiveness of Sound Field Data}

\begin{table}[!t]
\centering
\caption{User study for the sound field construction. 21 participants rated the sound field reconstruction.}
\label{tab:sound_res}
\footnotesize
\begin{tabular*}{\columnwidth}{@{\extracolsep{\fill}}l@{\hspace{2pt}}ccc}
\toprule
Rating & Spatial Perception & Sound Quality & Immersiveness \\
\midrule
Very poor & 0.00\%   & 0.00\%   & 0.00\%  \\
Poor      & 0.00\%   & 0.00\%   & 0.00\%  \\
Fair      & 14.28\%  & 19.04\%  & 9.52\%  \\
Good      & 23.80\%  & 33.33\%  & 42.85\% \\
Excellent & 61.90\%  & 47.61\%  & 47.61\% \\
\bottomrule
\end{tabular*}
\end{table}

\begin{table}[!t]
\centering
\caption{Ablation studies on the Sound Field Reconstruction (SFR) baseline. 58 participants rated sense of direction and distance (Scale 1--5).}
\label{tab:sound_field_ablation}
\footnotesize
\begin{tabular*}{\columnwidth}{@{\extracolsep{\fill}}lcc}
\toprule
Method & Direction Score $\uparrow$ & Distance Score $\uparrow$ \\
\midrule
w/o SFR                     & 1.17 & 1.67 \\
SFR + distance              & 1.69 & 3.02 \\
SFR + direction             & 3.81 & 3.03 \\
SFR + distance + direction  & \textbf{3.91} & \textbf{3.07} \\
\bottomrule
\end{tabular*}
\end{table}

\paragraph{Quantitative Results}
Since the proposed sound field reconstruction algorithm is non-trainable and operates on real-world multi-view audiovisual recordings, the reconstructed spatial audio does not have a corresponding ground-truth reference, making standard quantitative evaluation infeasible. Furthermore, reconstructing sound fields from multi-view audiovisual data is, to the best of our knowledge, largely unexplored in prior work, and no established benchmark or objective evaluation protocol currently exists. Therefore, we conduct a user study with 21 experts (all of whom provided informed consent prior to the study) to assess the perceptual quality and spatial plausibility of the reconstructed sound field.

Each participant rated the sound field reconstruction on three dimensions: auditory spatial perception, sound quality, and immersiveness. Auditory spatial perception refers to the listener's ability to perceive the distribution and localization of sound in space. Sound quality refers to the audio quality of the spatial audio compared to the corresponding microphone-recorded signal. Immersiveness refers to the listener's sense of being in a space where the sound source genuinely exists.

The results of the user study are shown in Table~\ref{tab:sound_res}, with numbers expressed as percentages of participants. 
The majority of the participants (61.90\%) rated the auditory spatial perception as excellent; 80.94\% of participants felt that the generated spatial audio did not exhibit significant degradation in quality, and 90.46\% found the audio immersive, which demonstrates the effectiveness of our data capture and sound field reconstruction methods.

\paragraph{Ablations} 
To further assess the effectiveness of each module in our proposed SFR method, we conduct an ablation study based on user feedback shown in Table~\ref{tab:sound_field_ablation}. The average scores for both metrics indicate that incorporating direction and distance modeling in sound field reconstruction significantly enhances participants' immersive experiences, further demonstrating our SFR algorithm as a practical baseline.

\begin{figure}
    \centering
    \includegraphics[width=\linewidth]{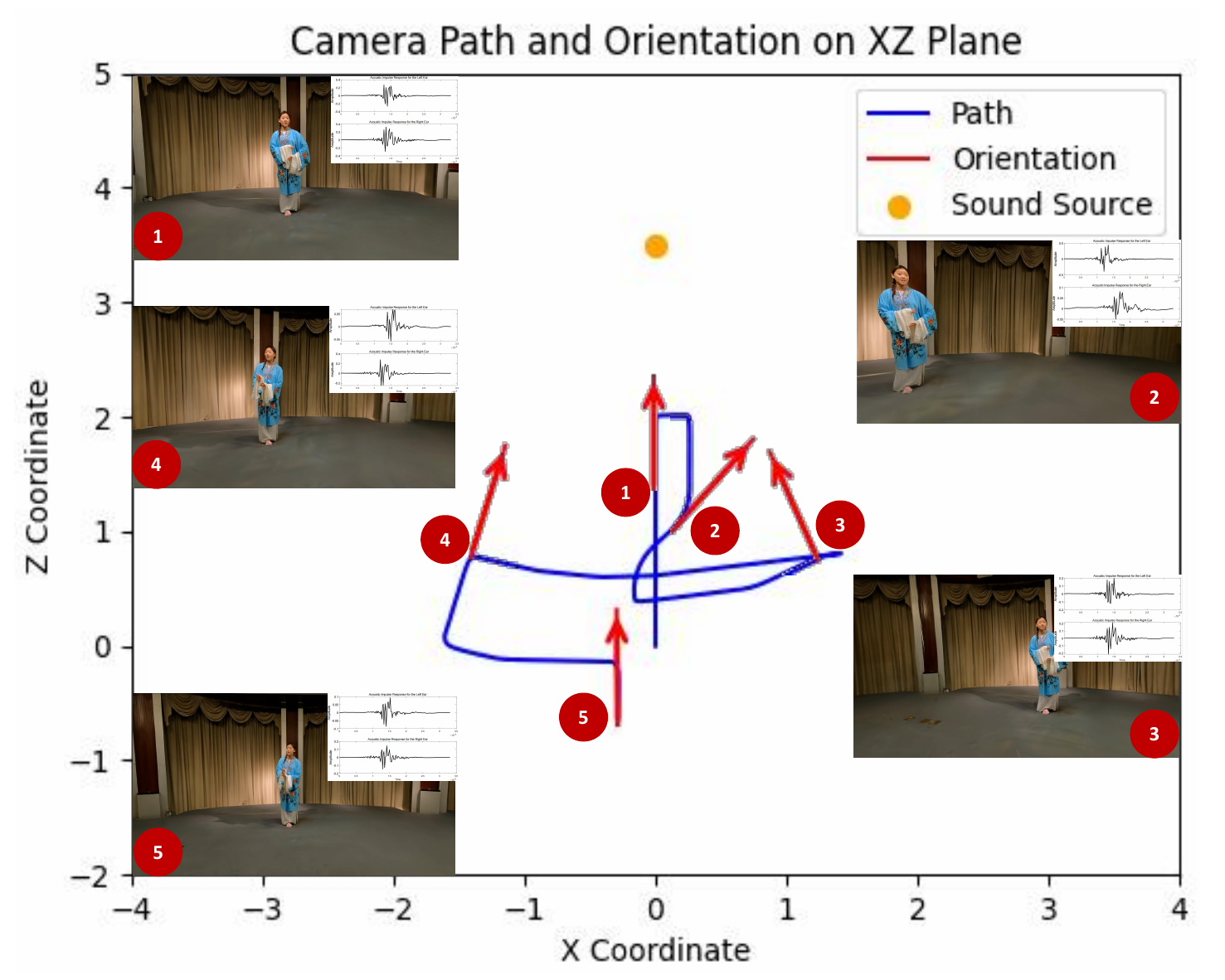}
    \caption{Visualization of the interaction trajectory and corresponding visual \& auditory results. The multi-modal feedback changes consistently with the positions and orientations.}
    \label{fig:application}
\end{figure}

\subsection{Multi-modal VR Experiences}
We finally integrated dynamic light field and sound field reconstruction results to achieve a 6-DoF multi-modal immersive VR experience with a real-time rendering speed of 60 FPS on a single 3090 GPU. As shown in Fig.~\ref{fig:application} , the visual and auditory experiences correspond uniquely to the user's position and viewing direction. The experience can be further explored by listening through headphones with the supplementary video.

\section{Discussion}
\label{sec:discussion}
Despite the promising results, our current framework has certain limitations.
First, the substantial volume of image data requires caching into memory to maintain training efficiency, which restricts the number of frames that can be processed concurrently on a single machine. Future research could investigate more scalable training strategies or advanced data-loading schemes to mitigate this memory bottleneck.
Second, our pipeline does not specifically optimize for preprocessing and training speed, which remains an area for potential improvement. Future work could incorporate learning-based priors, or even explore end-to-end reconstruction paradigms, to further enhance efficiency.

\section{Conclusion}
\label{sec:conclusion}
In this paper, we introduced the concept of \emph{Immersive Volumetric Videos (IVV)} and presented a comprehensive pipeline for its production.
First, we introduced \emph{ImViD}, a large-scale multimodal dataset captured with a custom-designed rig, offering wide spatial coverage and tightly synchronized audiovisual data tailored for 6-DoF VR experiences, addressing the critical scarcity of efficiently captured high-quality data in open-world scenarios. 
Second, to handle complex real-world motion, we proposed a robust dynamic light field reconstruction framework. Built upon a Gaussian-based spatio-temporal representation, our method integrates flow-guided initialization, joint temporal calibration and multi-term supervision, enabling the faithful reconstruction of high-fidelity, temporally coherent 4D light fields. 
Finally, we incorporated a sound field reconstruction module to achieve full multimodal immersion.
Extensive experiments demonstrate that our approach significantly outperforms existing methods in rendering quality, temporal stability, and audiovisual coherence. 
We believe this work establishes a solid foundation for next-generation immersive media and will inspire further research into holistic scene reconstruction, understanding, and generation.

\bibliographystyle{IEEEtran}
\bibliography{main.bib}

\vfill
\end{document}